%% file: main.tex
\documentclass[10pt,twocolumn,letterpaper]{article}

\usepackage{cvpr}              

\input{preamble}

\definecolor{cvprblue}{rgb}{0.21,0.49,0.74}
\usepackage[pagebackref,breaklinks,colorlinks,allcolors=cvprblue]{hyperref}

\def\confName{CVPR}
\def\confYear{2027}

\title{A Gene Informed Benchmark for BioScoring of Pathology Foundation Models }

\author{%
Paul Vilars$^{1,2,3}$ \qquad Vincent Quoc-Huy Trinh$^{4}$ \qquad Maria Vakalopoulou$^{1}$\\
Dimitris Samaras$^{5}$ \qquad Stergios Christodoulidis$^{1}$ \qquad Mahdi S. Hosseini$^{2,3}$\\[4pt]
$^{1}$CentraleSup\'elec \qquad $^{2}$Concordia University \qquad $^{3}$Mila--Quebec AI Institute\\
$^{4}$Universit\'e de Montr\'eal \qquad $^{5}$Stony Brook University\\
{\tt\small mahdi.hosseini@concordia.ca}
}

\begin{document}
\maketitle
\input{sec/0_abstract}    
\input{sec/1_intro}
\input{sec/2_related}
\input{sec/3_bioscore}
\input{sec/4_validity}
\input{sec/5_results}

\input{sec/6_conclusion}
\input{sec/7_llm_statement}
{
    \small
    \bibliographystyle{ieeenat_fullname}
    \bibliography{main}
}

\input{sec/X_suppl}

\end{document}

%% file: preamble.tex
\usepackage{microtype}
\usepackage{placeins}  

\renewcommand{\paragraph}[1]{\vspace{.5em}\noindent\textbf{#1.}}



%% file: sec/0_abstract.tex
\begin{abstract}
Pathology foundation models are usually compared by their performance on downstream tasks such as classification and
retrieval.
While such benchmarks measure how useful a model is, they provide no insight into the biology it represents.
We introduce BioScore, a benchmark that quantifies how well an image encoder captures cell identities, with spatial
transcriptomics of the same tissue as the reference.
A sparse autoencoder splits the encoder's patch embeddings into sparse features, which we cluster into a fixed number
of groups.
BioScore then measures, for each of 52 cell identities defined by marker genes, how specifically one group follows
the identity's expression over the tissue.
Because the autoencoder training is fully unsupervised, a match shows that the model represents the
cell identity from the image alone.
We score 8 pathology foundation models and 3 natural-image models.
Every pathology model scores above every natural-image model (0.299--0.371 \vs 0.214--0.232), and paired tests
separate three tiers among the pathology models.
Unlike a supervised probe, BioScore needs no labels, builds its reference from the data and scores even from few sections, yet it ranks the encoders as a probe does. Its per-identity breakdown 
indicates
which cell identities
each encoder captures 
or misses.
\end{abstract}

%% file: sec/1_intro.tex
\section{Introduction}
\label{sec:intro}

\begin{figure}[t]
  \centering
  \includegraphics[width=\linewidth]{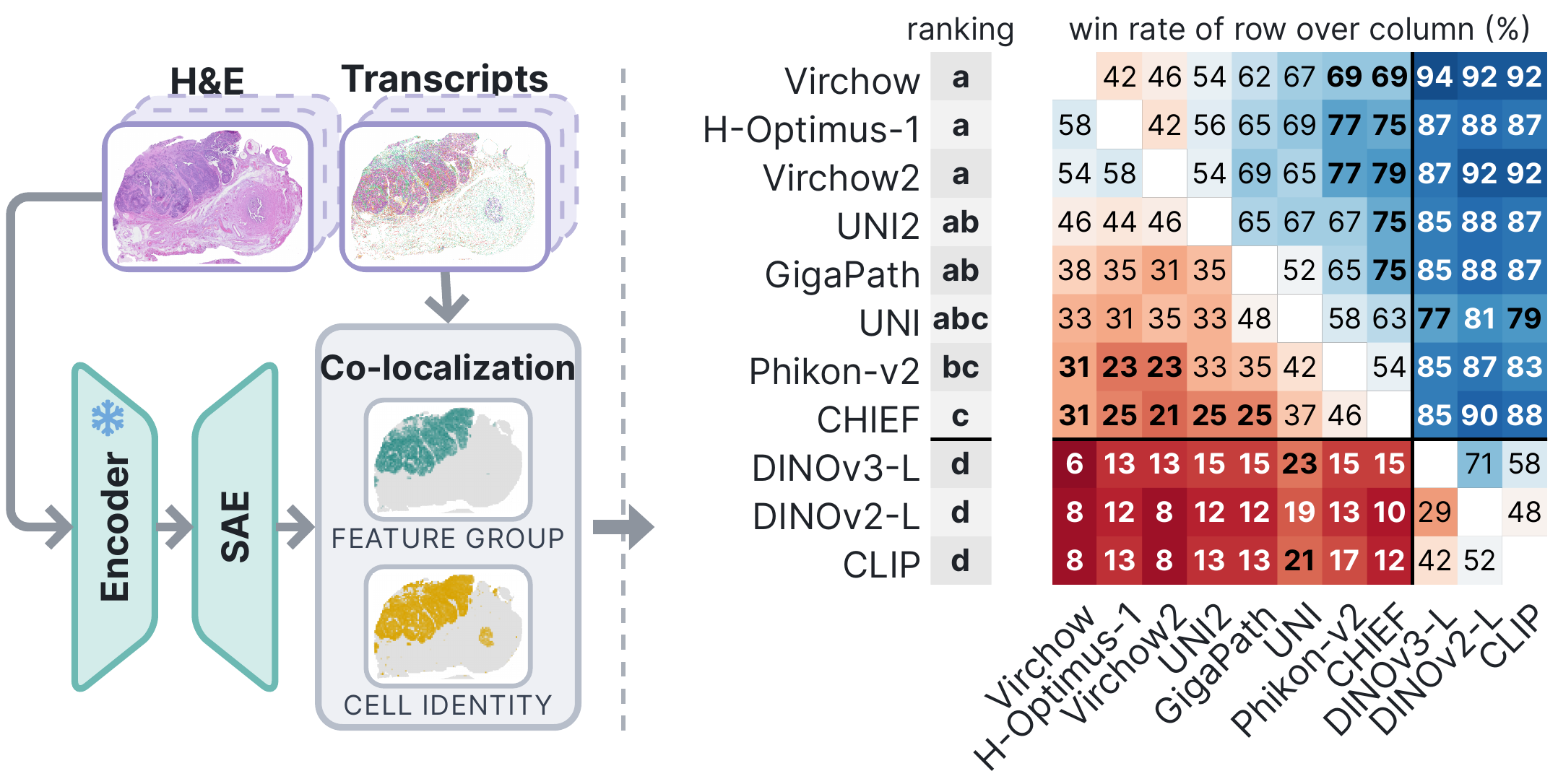}
  \caption{\textbf{BioScore overview and encoder ranking.}
  Left: SAE activation map is compared to cell-identity. Right: win rate (\%) of row over column across the 52 identities; bold: Holm-corrected Wilcoxon $p<0.05$; and compact letter ranking.}
  \label{fig:win}
\end{figure}

Pathology foundation models \cite{chen2024uni, vorontsov2024virchow, xu2024gigapath} are trained in a self-supervised manner on millions of H\&E patches, then compared on downstream tasks such as patch and slide classification, retrieval and, more recently, prediction of gene expression \cite{campanella2025clinical, neidlinger2025benchmarking, marza2025thunder,jaume2024hest}. Each of these benchmarks trains a probe on the embeddings, reports a performance metric and allow high level understanding. However, it does not elaborate on what the model relies on within the tissue. An accuracy on tumor subtyping, for instance, may come from tumor cells, stroma or immune infiltration.

Two kinds of tools come closer to asking whether a model has distinct units for biological concepts, but fall short of an explicit comparison between encoders: first, gene-expression regression \cite{jaume2024hest} fits a model on H\&E embeddings and measures whether expression can be decoded, not whether the encoder has a unit that tracks a kind of cell; a regression can recover a cell's markers from a mixture of many directions (\ie embedding dimensions). Second, sparse autoencoders (SAEs) split an embedding into sparse, interpretable features \cite{bricken2023monosemanticity, gao2024scaling}, and have recently been used to find concepts in pathology models \cite{le2025saeutility, kim2026picasso}. PICASSO trains an SAE on Virchow2 \cite{zimmermann2024virchow2} and has pathologists interpret selected concepts to audit and steer the model. Such expert review is not easily scalable, nor provide encoder ranking. 

Paired histology and in-situ transcriptomics could replace the expert in this loop. Xenium records the position of every transcript on the same section that was stained \cite{janesick2023xenium}, so the expression of known marker genes can be read in every image patch \cite{giraud2026sthelar}. We call a cell identity the cell type defined by a set of marker genes (\eg, plasma cells, by MZB1, JCHAIN and SDC1), and map the expression of these markers over the tissue. Our premise is that an encoder has a representation for a cell identity if one of its SAE features is active where the identity is, and only there.

We build BioScore on this premise (\cref{fig:overview}). We train one SAE per encoder, group its features, and, for each of 52 cell identities curated from CellMarker \cite{hu2026cellmarker3} markers, keep the group whose activation best follows the identity's map and not those of other identities (\cref{sec:bioscore}). The mean over identities ranks the encoders, and the per-identity values show which cell identities each encoder represents and which it misses. Because the SAE is trained without labels, a match shows that the encoder represents the identity from the image alone. Code and panel will be released upon publication.

We ensure that BioScore is a strong metric (\cref{sec:validity}): the ranking is stable over SAE seeds and recipes, and a supervised probe reflects a similar encoder ranking. On 11 encoders (\cref{sec:results}), BioScore clearly separates general vision from pathology models, resolves tiers among pathology models (\cref{fig:win}) and locates the differences between them on specific cell identities.

We make three contributions:
\begin{itemize}
  \item BioScore, a label-free benchmark of the cell identities an image encoder represents, with a panel of 52 cell identities built from CellMarker signatures and checked against Xenium transcriptomics (\cref{sec:bioscore});
  \item Evidence that BioScore measures specific biological signal: the importance of the impurity term, the stability over SAE seeds and recipes, and the agreement with a supervised probe (\cref{sec:validity}).
  \item A ranking of 11 encoders by the cell identities they represent: every pathology model scores above every natural-image model, and a leading group of six pathology models stands above Phikon-v2 and CHIEF; the per-identity scores further show which identities each encoder represents and which it misses (\cref{sec:results}).
\end{itemize}

%% file: sec/2_related.tex
\section{Related work}
\label{sec:related}

\begin{figure*}[h!]
  \centering
  \includegraphics[width=\linewidth]{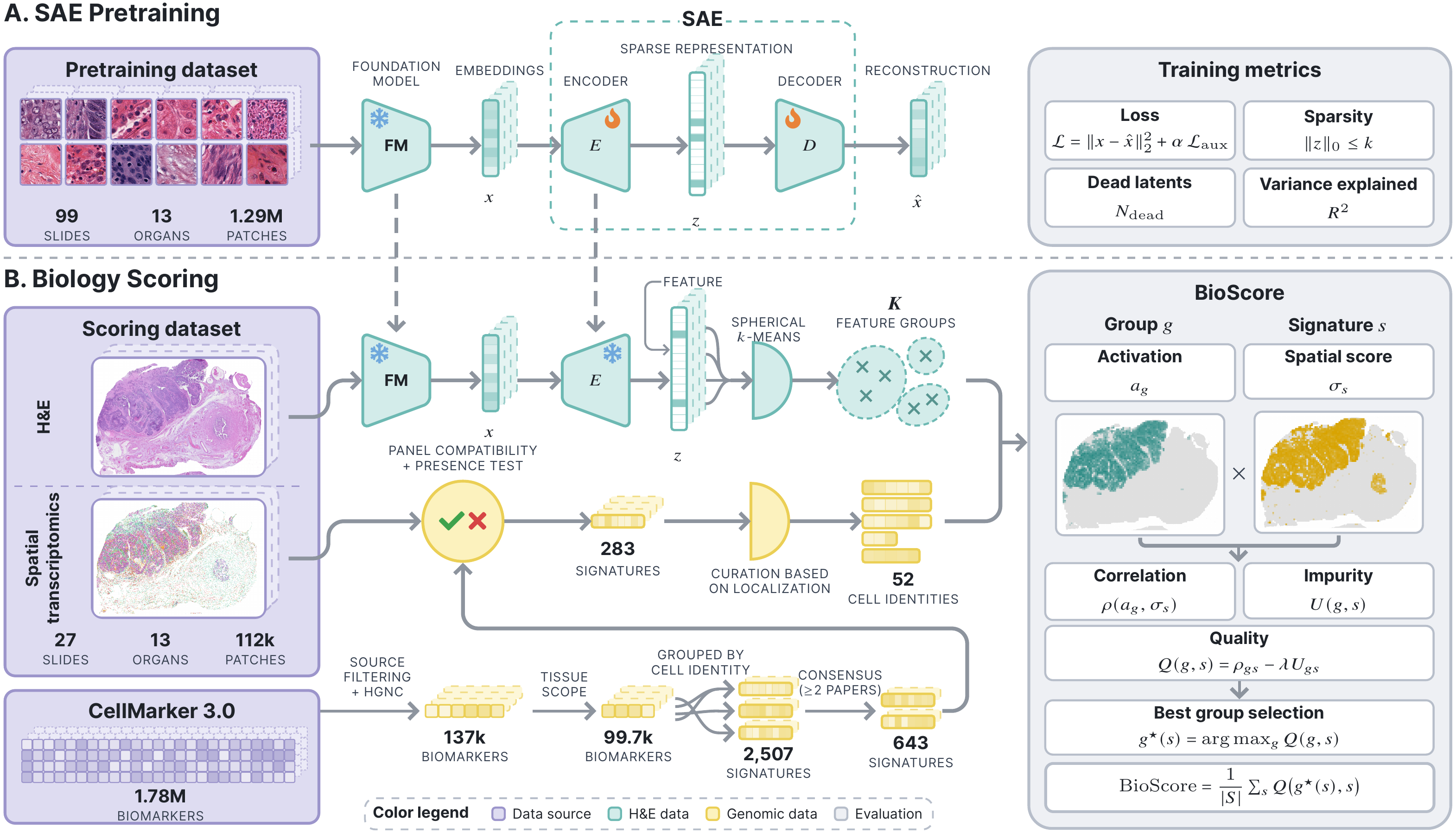}
  \caption{\textbf{Overview of BioScore.}
  (A)~An SAE is trained on the frozen patch embeddings of the encoder under test.
  (B)~On the scoring H\&E sections, its features are clustered into $K$ feature groups (blue); a curated panel of 52 cell identities is built from CellMarker (yellow); and
  each identity is credited with the group of best quality $Q$.
  Example: the Virchow2 group matched to ovarian carcinoma cells ($\rho=0.81$)}
  \label{fig:overview}
\end{figure*}

\paragraph{Evaluating pathology foundation models}
Pathology foundation models (\eg, UNI~\cite{chen2024uni}, Virchow~\cite{vorontsov2024virchow,zimmermann2024virchow2}) are compared on slide-level clinical tasks \cite{campanella2025clinical,neidlinger2025benchmarking,ma2025pathbench} and tile-level tasks \cite{marza2025thunder} by training a probe or a multiple-instance aggregator on the frozen embedding.
CellPath-Bench \cite{zhao2026cellpath} moves this evaluation to cells, with linear probes that decode cell identities from embeddings sampled at the nuclei of 25 H\&E--Xenium sections. These benchmarks measure what can be decoded from an embedding; BioScore uses no probe and no label, and asks whether the encoder has a sparse unit that follows a cell identity.

\paragraph{Histology and spatial transcriptomics}
ST-Net~\cite{he2020stnet} and HEST-1k~\cite{jaume2024hest} predict spatial gene expression from H\&E, the latter from
the frozen patch embeddings of foundation models, while OmiCLIP~\cite{chen2025loki} and
HESCAPE~\cite{gindra2025hescape} align histology and expression by contrastive training.
Xenium-based datasets such as STHELAR~\cite{giraud2026sthelar,janesick2023xenium} give paired H\&E and
transcriptomics at subcellular resolution.
These works predict transcriptomics or train with it; we use it only to evaluate a frozen encoder.

\paragraph{Sparse autoencoders}
SAEs decompose the activations of a network into many sparse features that are often easier to interpret than the
original dimensions~\cite{bricken2023monosemanticity,cunningham2024sae}, with sparsity enforced by an $\ell_1$
penalty~\cite{bricken2023monosemanticity}, a top-$k$ per sample~\cite{gao2024scaling} or per
batch~\cite{bussmann2024batchtopk}, or a learned threshold~\cite{rajamanoharan2024jumprelu}.
In vision, they have been used to align concepts between models~\cite{thasarathan2025usae} and to obtain more
stable dictionaries~\cite{fel2025archetypal}.
We borrow from SAE-TM~\cite{girrbach2025sae} the grouping of features through their decoder atoms.
Unlike SAEBench~\cite{karvonen2025saebench}, which evaluates SAEs themselves, we use the SAE as a fixed instrument:
every encoder goes through the same recipe.

\paragraph{Interpreting pathology foundation models}
Network Dissection~\cite{bau2017netdissect} scores the units of a network against segmentation masks of labeled
concepts.
In pathology, SAE features have been linked to cell counts, absent from a natural-image model~\cite{le2025saeutility},
and interpreted by pathologists to audit and steer Virchow2~\cite{kim2026picasso}.
These works study one model at a time and take the meaning of a unit from human review or cell annotations.
BioScore instead scores units against cell identities measured by gene expression on the same tissue, penalizes units
that are not specific, and ranks encoders.

%% file: sec/3_bioscore.tex
\section{BioScore}
\label{sec:bioscore}

For an image encoder $f$ and each of 52 cell identities, BioScore tests whether $f$ has a group of sparse features
whose activation over the tissue follows the expression of that identity's marker genes, and not the markers of other
identities. \cref{fig:overview} details how this scoring is performed. An SAE is first trained on the encoder's patch embeddings (\cref{sec:groups}). On  paired transcriptomics sections (\cref{sec:data}), its features are grouped into $K$ feature groups, and each group activation map is compared with the maps of a panel of cell identities built from curated gene signatures (\cref{sec:panel,sec:score}). Gene expression enters only at this comparison: neither the encoder nor the SAE is trained with transcripts.

\subsection{Data}
\label{sec:data}

\paragraph{Scoring set}
We compute BioScore on the STHELAR dataset \cite{giraud2026sthelar}: FFPE sections, H\&E whole-slide image
matched to 10x Xenium in situ transcriptomics \cite{janesick2023xenium}. Xenium records the position of every detected transcript, so the expression of an image patch is the transcripts present inside it, counted per gene.
We use 27 sections from 13 organs, cancerous and healthy; the gene panel differs between sections, from about 280 to
5,000 genes.
We leave out heart, bone-marrow and bone sections (4), for which there were no external slides  for the
pretraining set.
We tile the H\&E images into $224\times224$ px patches using AtlasPatch~\cite{alagha2026atlaspatch} automatically performing tissue detection, at 20$\times$
(0.54\,\textmu m/px, patch covers 120\,\textmu m); the magnification most of the encoders were
pretrained at.
A small portion of patches contained no transcripts (scanning artifacts), and could not be used for scoring  \cref{sec:supp_data}.
In total, we use 112,056 patches (\cref{tab:supp_pretrain}).

\paragraph{Pretraining}
We train the SAE  on a tissue-balanced set of 99 whole slides  including  27 scoring sections and 72 FFPE
diagnostic slides from TCGA~\cite{weinstein2013tcga} and CGCI-BLGSP~\cite{grande2019burkitt} for the same 13 organs,
one slide per patient.
For each organ, we  add external slides reaching 100k patches at 20$\times$, and keep every patch of each slide,  1.29M total patches (\cref{tab:supp_pretrain}). This pre-training dataset lets the SAE decompose the scoring sections, while enlarging its training to ensure a proper coverage of the original embedding space.

\subsection{From embeddings to feature groups}
\label{sec:groups}

\paragraph{Sparse autoencoder}
An embedding dimension of a foundation model has been shown to be very polysemantic ~\cite{bricken2023monosemanticity}, so asking whether the model has a unit for a cell
identity needs a basis in which units are sparse and monosemantic.
We train an SAE on the patch embeddings $x_p=f(I_p)\in\mathbb{R}^d$ of the pretraining set, with $8 \times$ expansion factor and the batch-level top-$k$ activation of Bussmann \etal~\cite{bussmann2024batchtopk}, which keeps
the $kB$ largest pre-activations of a batch of $B$ patches ($k=64$ per patch on average).
\begin{equation}
  \begin{aligned}
  z_p &= \mathrm{BatchTopK}\big(W_\mathrm{enc}(x_p-b_\mathrm{dec})+b_\mathrm{enc}\big),\\
  \hat x_p &= W_\mathrm{dec}\,z_p + b_\mathrm{dec},
  \end{aligned}
  \label{eq:sae}
\end{equation}
The loss is the reconstruction error plus the auxiliary loss of Gao \etal~\cite{gao2024scaling}, which also reconstructs the
residual from features that have not fired for 10M samples.
We use an open-source  
library~\cite{marks2024dictionarylearning} without changes,
with the same recipe for every encoder (\cref{sec:setup}).

\paragraph{Feature groups}
BioScore keeps the best match among the candidate units of an encoder (\cref{eq:bioscore}), so an encoder with more
features has mechanically more chances to match.
With an $8\times$ expansion, $n$ ranges from 6,144 (CLIP) to 20,480 (Virchow, Virchow2).
We therefore choose to group the features of every encoder into the same number $K$ of feature groups.
Following SAE-TM~\cite{girrbach2025sae}, which reads SAE features as topics and groups them through their decoder
atoms, we cluster the $\ell_2$-normalized decoder columns of the features that fire on the scoring set using spherical
$k$-means~\cite{dhillon2001concept}, weighting each feature by its mean activation share over the scoring patches
(3 initializations).
We define the activation of group $g$ in patch $p$ to be its share of the patch's total activation,
\begin{equation}
  a_g(p)=\frac{\sum_{f\in g}z_f(p)}{\sum_{f}z_f(p)}.
  \label{eq:group}
\end{equation}
where $z_f(p)$ is the activation of SAE feature $f$ on patch $p$ (\cref{eq:sae}).
We use $K=3000$ for every encoder.
Experiments show that grouping equalizes the number of candidates without removing signal (\cref{sec:stability}).

\subsection{Cell-identity panel}
\label{sec:panel}

\paragraph{Known signatures}
We score against marker lists from literature with specific meaning (\eg evidence of B-cell markers) and independent of both the images and the SAE.
We start from  1.78M   CellMarker 3.0~\cite{hu2026cellmarker3} human biomarkers, each linking one gene to one cell
identity in one paper.
We keep only entries from single-cell sequencing, experiments and reviews, based on independent studies, and normalize gene names to HGNC symbols (137k entries). Entries are further restricted to the 13 organs of the scoring set or general scope (\eg, blood, vasculature), leaving 99.7k entries, which we group by cell identity and scope using  Cell Ontology~\cite{diehl2016cell}. A consensus check keeps  biomarkers reported by at least 2 papers, and signatures with fewer than 3 such genes
are dropped (643 left).

\paragraph{Panel compatibility}
A signature can only be scored on sections whose panel measures its genes, and a significant portion of signatures contain
markers that Xenium panels lack.
For each signature we  keep the largest set of its genes, ranked by the number of supporting papers and at
least 3 long, that is fully measured on at least one section (the whole signature when possible). This leaves 404
signatures.
A signature $j$ is scored on the sections $\mathcal{R}_j$ that measure all its genes $G_j$.

\begin{table}[t]
  \centering
  \caption{\textbf{Symbols used} (\cref{sec:panel}).}
  \label{tab:sigscore}
  \small
  \setlength{\tabcolsep}{4pt}
  \begin{tabular}{@{}cl@{}}
    \toprule
    Symbol & Meaning \\
    \midrule
    $\sigma_j(p)$ & score of signature $j$ in patch $p$ \\
    $G_j$ & genes of signature $j$ \\
    $\mathcal{R}_j$ & sections whose panel measures all of $G_j$ \\
    $c_{pi}$ & number of transcripts of gene $i$ in patch $p$ (\cref{sec:supp_data}) \\
    $T_p$ & depth of a patch $p$: $\sum_i c_{pi}$  \\
    $N_r$ & number of genes in the panel of section $r$ \\
    $T'_p$ & rescaled depth, comparable across panels \\
    $m_i$ & base frequency of gene $i$: mean of $c_{qi}/T'_q$ over the \\
          & patches of all sections in $\mathcal{R}_j$ \\
    $s$ & cell identity, a set of signatures \\
    $\sigma_s(p)$ & map of identity $s$ in patch $p$ (\cref{eq:identity}) \\
    \bottomrule
  \end{tabular}
\end{table}

\paragraph{Presence}
A curated marker list is only useful here if the cell identity is present in the data and its markers are captured
well enough to show it.
We keep the signatures whose 1\% highest-scoring patches on their best section reach a mean score (\cref{sec:sigscore}) of at least 2 (a
four-fold excess of their markers), which leaves 283 signatures.

\paragraph{Curation based on localization}
CellMarker often describes one cell identity several times, in different tissues or studies, and on this data such
versions give the same map.
We cluster the signatures by the correlation of their maps on the sections they share (average linkage) and from a set whose mean correlation is at least 0.6, we retrieve the main identity that most of its signatures describe.
This step is data-dependent: on our sections, for instance, a mesothelial signature (WT1, UPK3B, MSLN) follows the
ovarian carcinoma and falls into the same set, but it is dropped in favor of the ovarian carcinoma signatures.
This process led to the creation of a final panel of 52 cell identities from 104 signatures: immune 31, cancer 9, stromal 4, parenchyma 3, endothelial 2,
epithelial 2 and other 1 (\cref{tab:supp_panel}). Immune related genes are  most often included in Xenium panels, and therefore make up a large part of our cell panel. Inversely, some identities have their main markers absent from every panel (\eg, fibroblasts and hepatocytes, which rely on COL1A2 and
ALB) and could not be scored on this cohort.

\paragraph{Signature score}
\label{sec:sigscore} The score of signature $j$ in patch $p$ of section $r$ compares the count of each of its genes with an expected count. We take the over-expected count ratio for each of its genes, and apply log geometric mean to compute a signature score (\cref{eq:sigscore}).
\begin{equation}
  \begin{aligned}
  \sigma_j(p)&=\max\Big(0,\ \frac{1}{|G_j|}\sum_{i\in G_j}\log_2\frac{c_{pi}+\frac12}{m_i\,T'_p+\frac12}\Big),\\
  T'_p&=T_p\,\frac{|G_j|}{N_r},
  \end{aligned}
  \label{eq:sigscore}
\end{equation}

\cref{tab:sigscore} lists symbols used in this section.
The score handles multiple properties of Xenium data.
Panels range from about 280 to 5,000 genes, so the rescaled depth $T'_p$ accounts for the actual signature fraction of a panel.
Markers differ in expression level, so each gene is compared with its own base frequency. Additionally, if a gene makes up a fixed share of the transcripts, a patch twice as deep should hold twice as many copies. We account for this, and the expected count is the base frequency $m_i$ of gene $i$, times the depth of the patch $T'_p$.
The use of the geometric mean ensures activation only when all markers are up together, and the pseudo-count
$\frac12$ is the standard correction for log-ratios of counts.
Finally, the cut-off at 0 shapes the score like a SAE activation, non-negative with 0 for absent.

\paragraph{Identity map}
The map of an identity $s$ on a section is the per-patch maximum of its measured signatures scores.
\begin{equation}
  \sigma_s(p)=\max_{j\in s,\ r\in\mathcal{R}_j}\sigma_j(p),
  \label{eq:identity}
\end{equation}
and the identity is scored on all the sections of its signatures (from 1 to 26, median 7; \cref{tab:supp_panel}). To ensure the quality of our panel, a board-certified pathologist reviewed the cell identity activation maps against the H\&E sections to check that their localization is coherent with the identity they describe.


\subsection{Matching Score}
\label{sec:score}

\paragraph{Correlation}
The correlation $\rho(g,s)$ is the Pearson correlation between $a_g$ and $\sigma_s$, pooled over the sections of $s$ where at least one of the two is not null. It rewards groups whose activation follows patches where the identity is clearly present, in all sections, penalizing groups that follow a general identity (\eg, T cells) in some tissues but not others.

\paragraph{Impurity}
A high $\rho(g,s)$ does not yet make $g$ a unit for $s$: the group may follow other cell identities that share some localization 
with $s$ (\eg, all immune cells, or a whole tumor region).
If $g$ were $s$ plus independent noise, its correlation with another identity $s'$ would equal
$\rho(s,s')\,\rho(g,s)$, the value at which $g$ and $s'$ have zero partial correlation given $s$.
Correlation above this floor is evidence that $g$ also encodes $s'$, and we measure it as
\begin{equation}
  U(g,s) = \Big(\frac{1}{|S'_s|}\sum_{s'\in S'_s}\big[\rho(g,s')-\rho(s,s')\,\rho(g,s)\big]_+^2\Big)^{1/2},
  \label{eq:impurity}
\end{equation}
where $S'_s$ holds  identities that share at least one section with $s$, with correlations computed on the patches of the shared sections.
The penalty is one-sided: a group that follows $s'$ less than the floor is more specific to $s$ than expected.

\paragraph{BioScore}
We define the quality of group $g$ for identity $s$, the best group $g^\star$, and the BioScore of an encoder as follows: 
\begin{equation}
  \begin{aligned}
  Q(g,s) &= \rho(g,s)-\lambda\,U(g,s),\\
  g^\star(s) &= \arg\max_g Q(g,s),\\
  \mathrm{BioScore} &= \frac{1}{|C|}\sum_{c\in C}\frac{1}{|S_c|}\sum_{s\in S_c}Q\big(g^\star(s),s\big),
  \end{aligned}
  \label{eq:bioscore}
\end{equation}
with $\lambda=3$ (\cref{sec:impurity}). We equalize the mean across lineage using classes $C$, one for each of immune (31), cancer (9), stromal (4), parenchyma (3) and one
class for the remaining five (endothelial, epithelial and the neurovascular unit). This prevents the 31 immune identities from dominating the score.
The values $Q(g^\star(s),s)$ also give a breakdown by identity.
By those means, BioScore provides a comparison metric between encoders: we train five SAEs per encoder (seeds 40--44), report the mean and standard deviation of BioScore over
them, and compare encoders with paired tests over identities (\cref{sec:ranking}).

%% file: sec/4_validity.tex
\section{Does BioScore measure biology?}
\label{sec:validity}

We first check that BioScore properly measures biology.
It should credit groups that follow one cell identity, not broad activity, and  be independent of the SAE
training run.

\subsection{Encoders and protocol}
\label{sec:setup}

\paragraph{Frozen encoders}
We compare 8 pathology foundation models and 3 natural-image models (\cref{tab:encoders}), all given the same
$224\times224$\, patches.
Most pathology models were pretrained on 20$\times$ patches; Virchow2 was pretrained on 5$\times$, 10$\times$,
20$\times$, 40$\times$ patches, and CHIEF was trained on 10$\times$ tiles.
DINOv2-L and Prov-GigaPath resize their input to 256\,px and crop its center back to 224\,px by default; we give them
the whole patch instead, so that each encoder sees the same tissue as the transcripts counted for the patch
(\cref{sec:supp_encoders}).

\paragraph{Protocol}
For each encoder, an SAE with expansion rate of $8 \times$ and $k=64$ is trained for 100 epochs on the 1.29M patches of the
pretraining set (\cref{sec:groups}) with the default settings of the
library~\cite{marks2024dictionarylearning} (batch 2,048, automatic learning rate with warm-up and decay;
\cref{sec:supp_encoders}), with five seeds (40--44).
We monitor training metrics to ensure SAEs reconstruct the embeddings well: high reconstruction is reported, with very low dead latents proportion (\cref{tab:supp_health}).
Dictionary size and sparsity of the SAE are also varied to ensure that the ranking is not recipe-dependent (\cref{sec:stability}).

\subsection{Correlation and impurity}
\label{sec:impurity}

\begin{figure}[t]
  \centering
  \includegraphics[width=\linewidth]{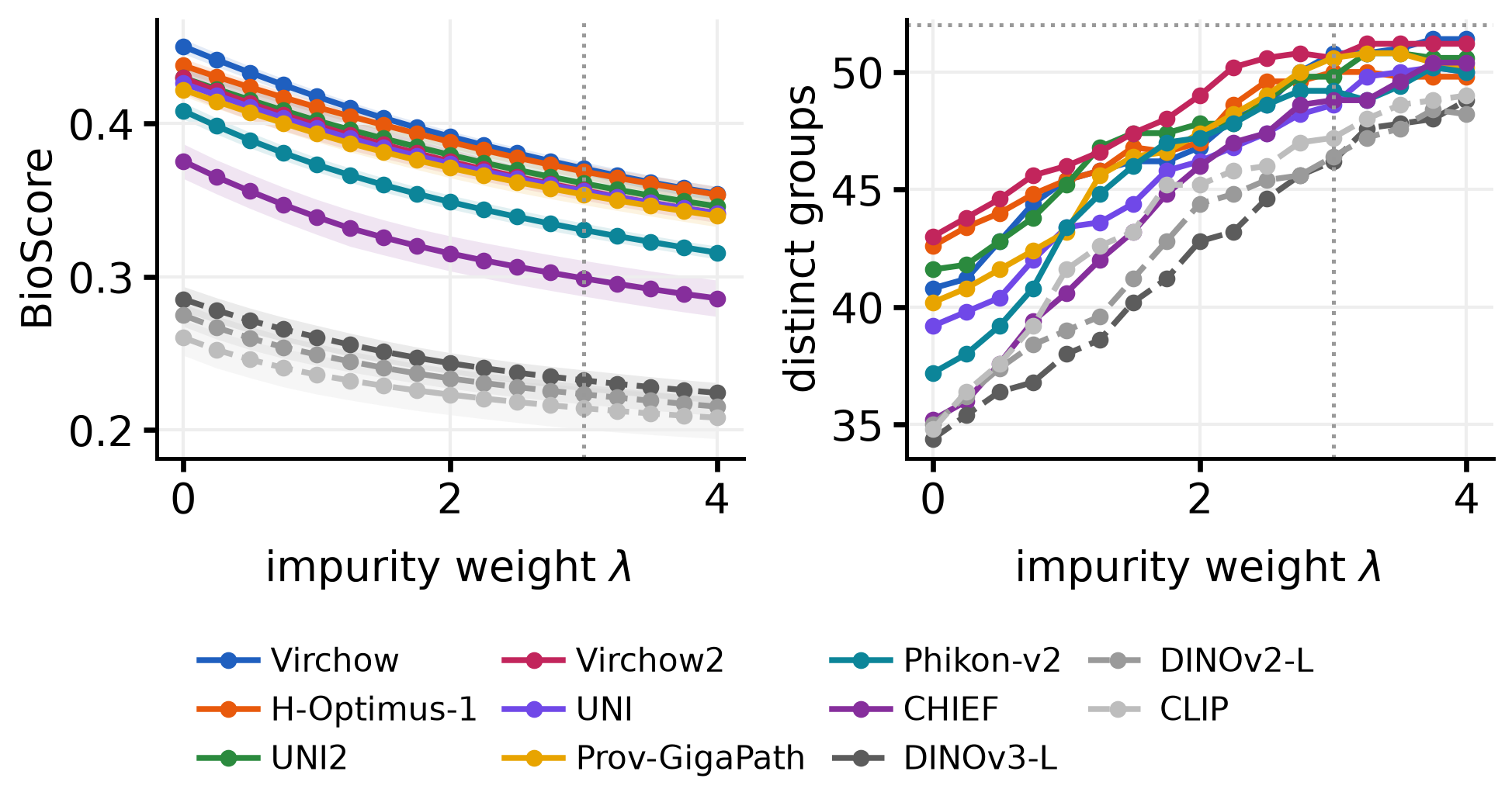}
  \caption{\textbf{Effect of impurity weight $\lambda$.}
  BioScore (left) and distinct matched groups (right); dotted:  $\lambda=3$. Full results in \cref{tab:lambda}.}
  \label{fig:lambda}
\end{figure}

\paragraph{Correlation alone orders the encoders}
BioScore without the impurity term  ($\lambda=0$) ranks the
encoders almost as the default one (Kendall $\tau=0.96$), and the ranking is identical for $0.25\leq\lambda\leq3.75$
(\cref{tab:lambda}).
Raising $\lambda$ lowers every encoder in parallel (\cref{fig:lambda}, left).

\paragraph{Impurity separates the matches}
Increasing $\lambda$ refines which group matches each identity.
Without impurity, the best group for an identity can actually follow several co-localized identities, and the
52 identities are matched to only 34 to 43 distinct groups.
The number of distinct groups rises with $\lambda$ until $\lambda\approx3$, where the 52 identities are matched to
46 to 51 distinct groups, and then levels off (\cref{fig:lambda}, right; \cref{tab:lambda}).
We therefore use $\lambda=3$, where the number of distinct group plateaus.

\paragraph{Choice of correlation}
Using a rank correlation (Spearman) instead of Pearson favored groups that are mildly active over much of the tissue; whereas Pearson correlation weighs hotspots where an identity is clearly present, which motivated this selection of correlation metric (\cref{sec:supp_corr}).

\subsection{Comparison with a supervised probe}
\label{sec:downstream}

We test whether the encoders’ embeddings carry our identities with a supervised probe, the approach commonly used to measure how well pathology encoders capture spatial expression~\cite{jaume2024hest}.
For each identity, we train a probe to predict its map $\sigma_s$ from the patch embeddings of the scoring set.
We use HEST implementation: the frozen embeddings are standardized and reduced to 256 principal components fitted on the
training patches, and a ridge regression (no intercept) predicts $\sigma_s$.
Each section of the identity is held out once, and the probe is trained on its other sections.
We compute the Pearson correlation between the predictions and $\sigma_s$ on all patches over its sections as for $\rho$ (\cref{sec:score}).
This protocol needs the identity present on at least two sections; otherwise the fold that holds out its only hotspot section
trains without a positive example.
Only 27 of the 52 identities meet this criterion and constitute our comparison basis.
The probe gives a supervised reference for the ranking of encoders, which we can compare to the ranking of
BioScore.

\begin{table}[t]
  \centering
  \caption{\textbf{Supervised probe vs.\ BioScore.}
  Mean over the 27 identities seen on at least two sections; rank in parentheses.
  The two rankings agree at Kendall $\tau=0.82$.}
  \label{tab:probe}
  \scriptsize
  \setlength{\tabcolsep}{4pt}
  \begin{tabular}{@{}lcc@{}}
    \toprule
    Encoder & Probe & BioScore \\
    \midrule
    Virchow       & .537 (1) & .295 (1) \\
    H-Optimus-1   & .513 (2) & .289 (2) \\
    Virchow2      & .512 (3) & .286 (4) \\
    UNI2          & .509 (4) & .287 (3) \\
    Phikon-v2     & .487 (5) & .256 (7) \\
    UNI           & .484 (6) & .267 (6) \\
    Prov-GigaPath & .482 (7) & .274 (5) \\
    CHIEF         & .436 (8) & .241 (8) \\
    \midrule
    DINOv3-L      & .346 (9) & .171 (9) \\
    DINOv2-L      & .344 (10) & .156 (11) \\
    CLIP          & .333 (11) & .165 (10) \\
    \bottomrule
  \end{tabular}
\end{table}

\begin{figure}[t]
  \centering
  \includegraphics[width=\linewidth]{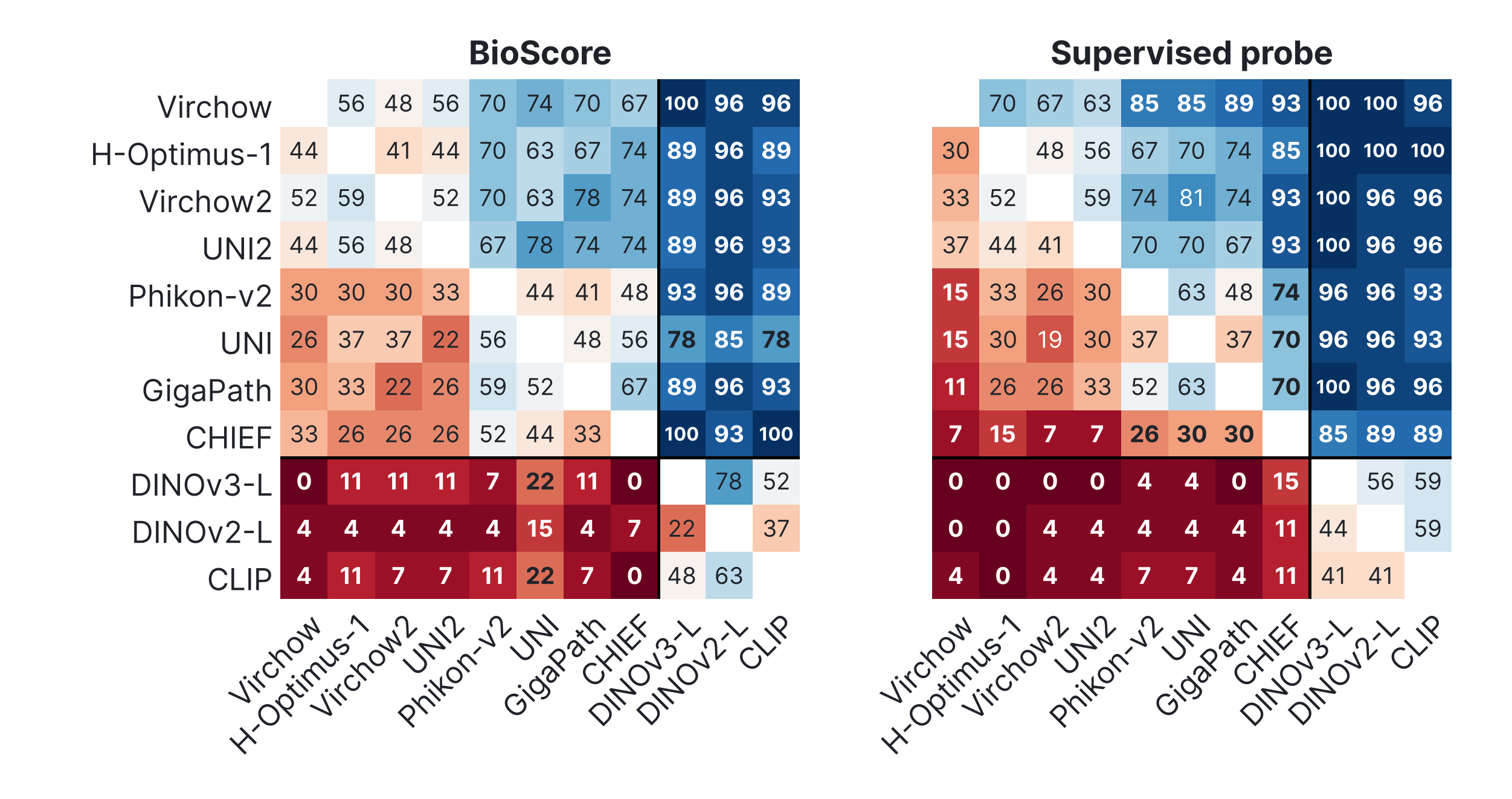}
  \caption{
  Win rate (\%) of row over column across the 27 comparison identities of \cref{tab:probe}; bold: Holm-corrected Wilcoxon
  $p<0.05$.}
  \label{fig:win_probe}
  \vspace{-0.2in}
\end{figure}

\paragraph{The probe and BioScore agree on encoder ranking}
Averaged over 27 identities (\cref{tab:probe}), the two orders agree at Kendall $\tau=0.82$ over  11 encoders, and both place all pathology models above all natural-image models (probe scores
0.44--0.54 for pathology, 0.33--0.35 for natural-image models).
An encoder that decodes the identities well also tends to have feature groups that follow them.
The same paired tests as for BioScore (\cref{sec:ranking}), run on the probe correlations (\cref{fig:win_probe}), confirm the superiority of pathology models above natural-image models
(win rates 85--100\%, all 24 pairs significant) and provide a similar ordering of pathology models to BioScore ($\tau=0.71$ over the 8 pathology models alone).
Both methods rank Virchow, H-Optimus-1, Virchow2 and UNI2 as the strongest models.
Probing with a few units is an established test of interpretability~\cite{gurnee2023finding,karvonen2025saebench},
and in language models SAE features rarely match a supervised probe~\cite{kantamneni2025sparse}.
Similarly here, the group chosen through BioScore follows the identity less precisely than the probe does ($\rho$
0.21--0.37 against 0.33--0.54 on average \cref{tab:supp_probe_rho}), but it orders the encoders in the same way, which confirms that BioScore comparisons do translate to prediction performances.

\paragraph{Ranking on specific identities}
On the other hand, depending on the identity evaluated, the order of the 8 pathology models may vary a lot compared to BioScore (median Spearman 0.33 between probe and
BioScore; \cref{tab:supp_probe}), implying that the two methods separate the leading models on different
identities.
On pancreatic islet endocrine cells, BioScore separates the models more clearly than the probe: the groups of
Virchow, UNI and H-Optimus-1 follow the islets closely ($\rho$ 0.66--0.71), that of CHIEF much less (0.30), and those
of the natural-image models hardly at all (0.06--0.12).
The probe scores are all low and overlap between the two families (0.09--0.34 for pathology models against 0.02--0.13
for natural-image models; \cref{tab:supp_probe_examples}).
On T cells and B cells, the reverse holds: the probe separates pathology from natural-image models as clearly as on
most identities, but no pathology model has a group that follows these identities markedly better than the
natural-image models (gap between the two families divided by the spread over encoders: 2.0 and 1.9 for the probe,
0.5 and $-$0.03 for BioScore) (\cref{tab:supp_probe_examples}).

\paragraph{Beyond the probe}
Both methods rank the encoders alike and differ on single identities.
The probe, however, needs labels and held-out sections, so it cannot evaluate the 25 identities seen on a single section
(\eg, ovarian carcinoma cells, cDC1).
BioScore selects among feature groups learned without labels: it scores every identity and returns a group whose map can
be inspected (\cref{fig:maps_main}).

\subsection{Independence from the SAE}
\label{sec:seeds}
\label{sec:stability}

BioScore reads an encoder through an SAE, but its result must be independent from SAE training.
We check for seed, dictionary size, sparsity and number of feature groups.

\begin{table}[t]
  \centering
  \caption{\textbf{SAE recipe ablation.}
  Mean BioScore over the 11 encoders without grouping ($K=n$) and Kendall $\tau$ against the main recipe; per encoder
  in \cref{tab:supp_recipe}.}
  \label{tab:recipe}
  \small
  \begin{tabular}{@{}lcc@{}}
    \toprule
    Recipe ($K=n$) & BioScore & $\tau$ \\
    \midrule
    $\times8$, $k=64$ (main) & .319 & -- \\
    $\times4$, $k=64$ & .309 & 0.96 \\
    $\times16$, $k=64$ & .322 & 0.96 \\
    $\times8$, $k=32$ & .315 & 0.89 \\
    $\times8$, $k=128$ & .320 & 1.00 \\
    \bottomrule
  \end{tabular}
\end{table}

\begin{figure}[t]
  \centering
  \includegraphics[width=\linewidth]{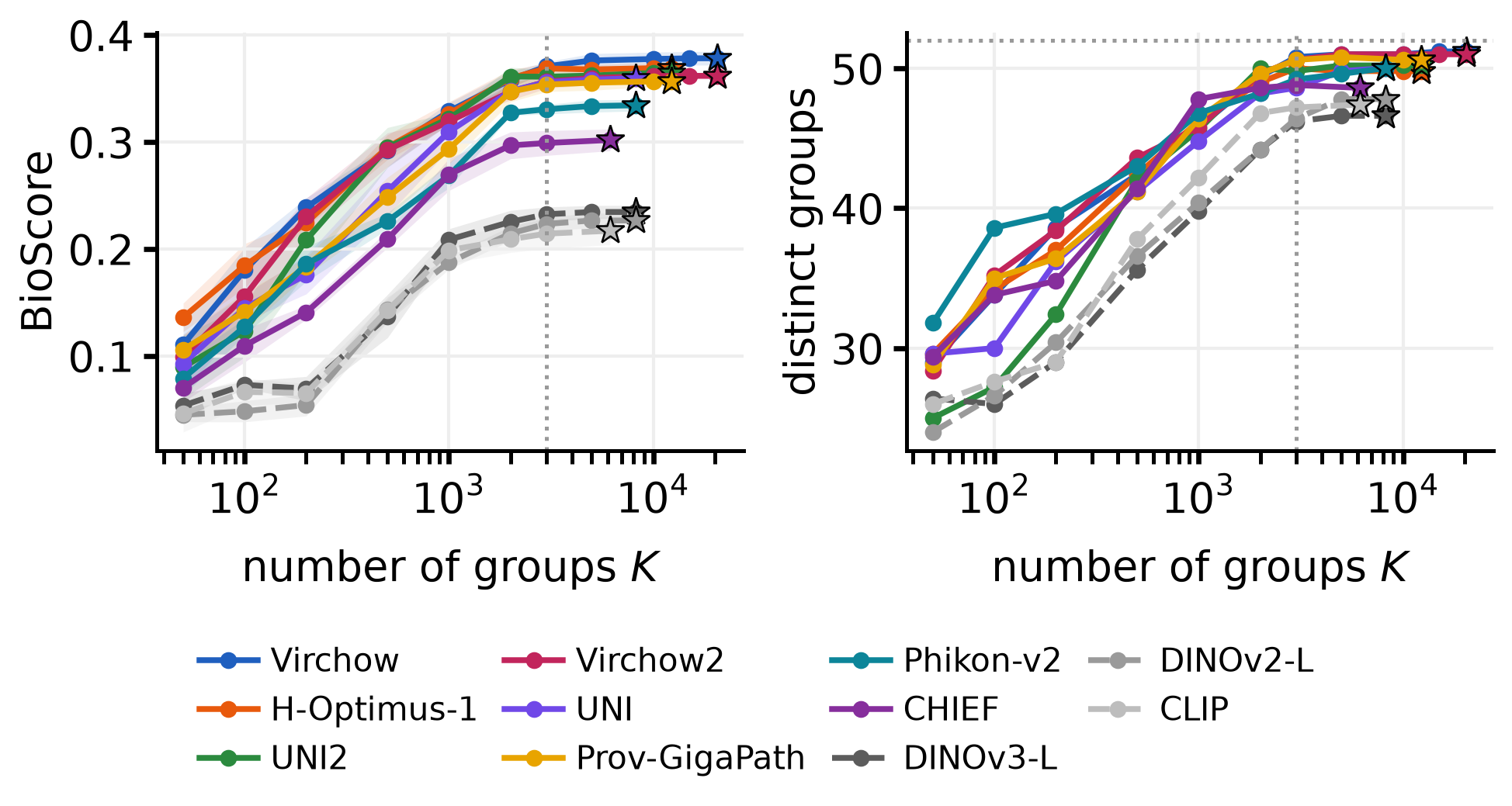}
  \caption{\textbf{K ablation.} BioScore (left) and distinct matched groups (right), plotted against K up to one group per feature (stars); dotted: default $K=3000$. Full results in \cref{tab:K}.}
  \label{fig:K}
  \vspace{-0.2in}
\end{figure}

\paragraph{Ranking hardly depends on training seed}
Over five seeds, the standard deviation of BioScore is 0.003--0.014 per encoder (\cref{fig:ranking,tab:main}), an order of magnitude below the gap of 0.126 between pathology and natural-image models.
The ranking itself is largely fixed.
Rankings obtained with two different seeds agree at Kendall $\tau$ 0.83 on average (\cref{tab:supp_seeds}), and in every seed Phikon-v2 is 7th, CHIEF 8th and the 3 natural-image models take
the last 3 places.
Only the order within the 6 strongest pathology models changes from seed to seed, Virchow and H-Optimus-1
always among the first three.
Each pathology model also beats each natural-image model in all 25 pairings of their 5 seeds.

\paragraph{Training recipe does not changes the results}
We vary the dictionary size ($\times4$ and $\times16$ instead of $\times8$) and the sparsity ($k=32$ and $128$
instead of 64), with five seeds each.
A $\times4$ dictionary of the smaller encoders has fewer than 3,000 active features, so we compare recipes without
grouping ($K=n$): over the 11 encoders, the mean BioScore is 0.309--0.322 for the four variants against 0.319 for the
main recipe, and the rankings agree with the main one at Kendall $\tau$ 0.89--1.00 (\cref{tab:recipe}; per
encoder in \cref{tab:supp_recipe}).

\paragraph{The number of groups is not critical above 2,000}
BioScore rises with $K$ up to about 2,000--3,000 groups and then levels off.
Without grouping, the score of each encoder changes by at most 0.008, and the ranking is the same as at $K=3000$;
between $K=500$ and 2,000 it agrees with $K=3000$ at Kendall $\tau$ 0.82--0.93, and it departs further for
$K\leq200$ ($\tau$ 0.75--0.78; \cref{tab:K,fig:K}).
The number of distinct matched groups follows the same curve, from 24--32 at $K=50$ to 46--51 at $K=3000$.

\subsection{Choice of magnification}
\label{sec:choice_mag}

We build the scoring and pretraining sets at 20$\times$ (120\,\textmu m patches) and  40$\times$ (60\,\textmu m),
and train the SAE at 20$\times$, 40$\times$ or on a mix of both, which gives 4 settings (training / scoring):
20$\times$/20$\times$, 20$\times$/40$\times$, 40$\times$/40$\times$ and mixed/40$\times$.
We report 20$\times$/20$\times$, the magnification  most encoders were pretrained with (\cref{tab:encoders}).
All pathology models score above  natural-image models in all 4 settings; scoring at 40$\times$ reorders the
leading models (Kendall $\tau$ 0.71--0.78 against 20$\times$), whereas the training magnification matters less
(\cref{sec:magnification}).

%% file: sec/5_results.tex
\section{What biology do foundation models capture?}
\label{sec:results}

\begin{figure*}[h!]
  \centering
  \includegraphics[width=\linewidth]{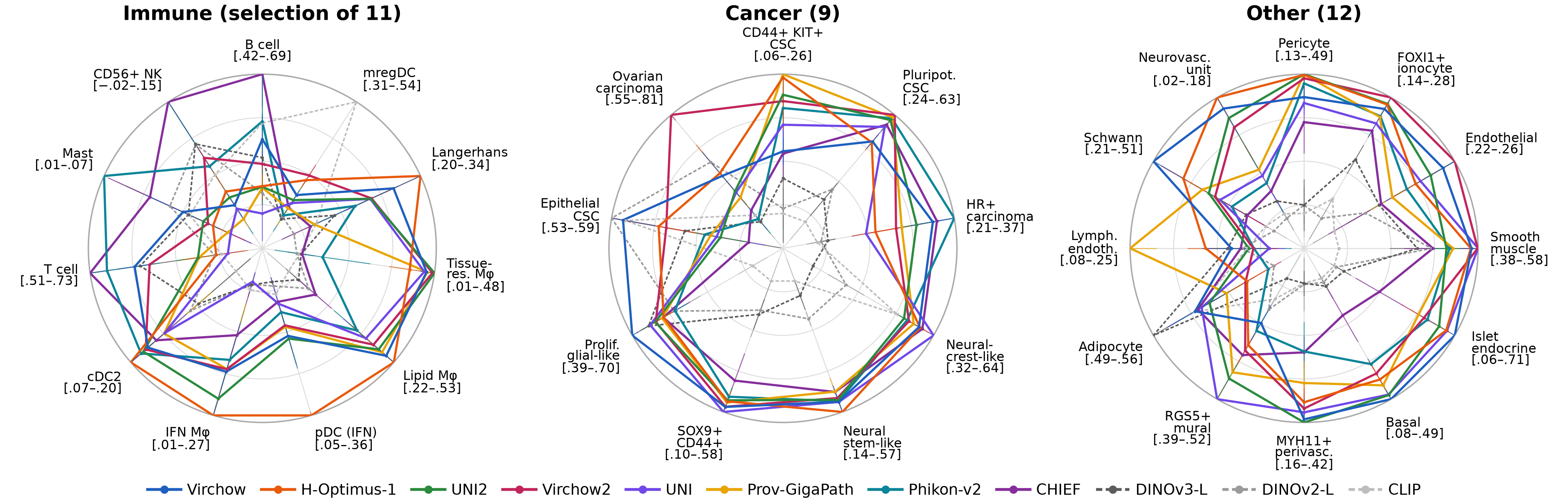}
  \caption{\textbf{Correlation per identity.}
  Correlation $\rho$ of the matched group, one scaled axis per identity (range
  in brackets); immune: selection of 11 identities, all 52 in \cref{fig:supp_radar_all}. Full results in \cref{tab:supp_identity_rho}.}
  \label{fig:radar}
  \vspace{-0.1in}
\end{figure*}

\begin{figure*}[h!]
  \centering
  \includegraphics[width=\linewidth]{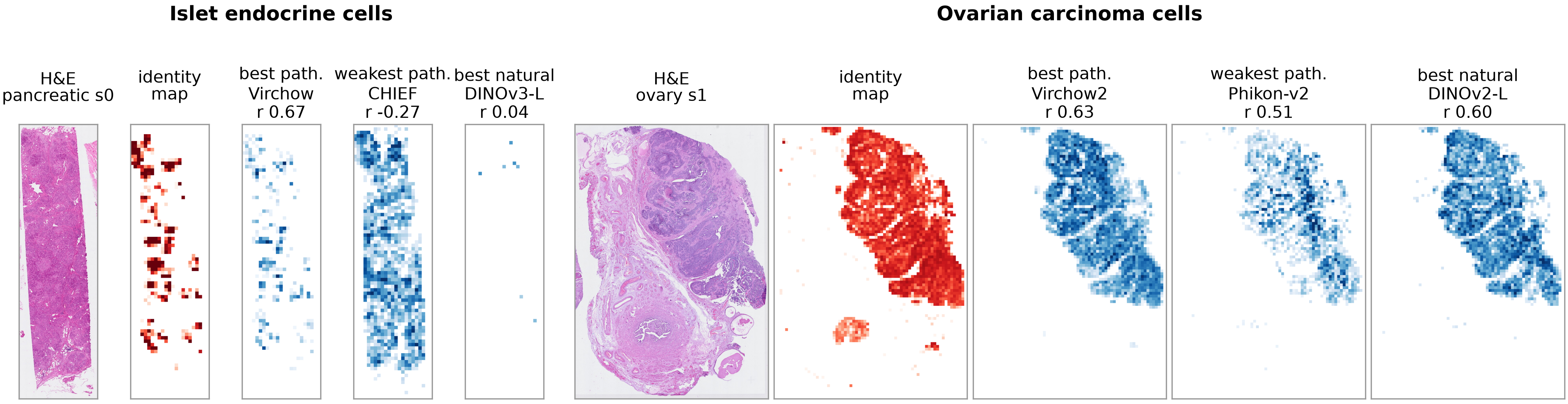}
  \caption{\textbf{Sample of two matched groups with cell identity.}
  H\&E, identity map (red) and matched groups (blue) of the best and weakest pathology models and the best
  natural-image model for that identity; $r$: correlation on the section.}
  \label{fig:maps_main}
  \vspace{-0.2in}
\end{figure*}

We score the 11 encoders of \cref{sec:setup} with SAEs trained and scored at 20$\times$ ($K=3000$, $\lambda=3$, five
seeds, \cref{tab:main}).

\subsection{Ranking}
\label{sec:ranking}

\begin{figure}[t]
  \centering
  \includegraphics[width=\linewidth]{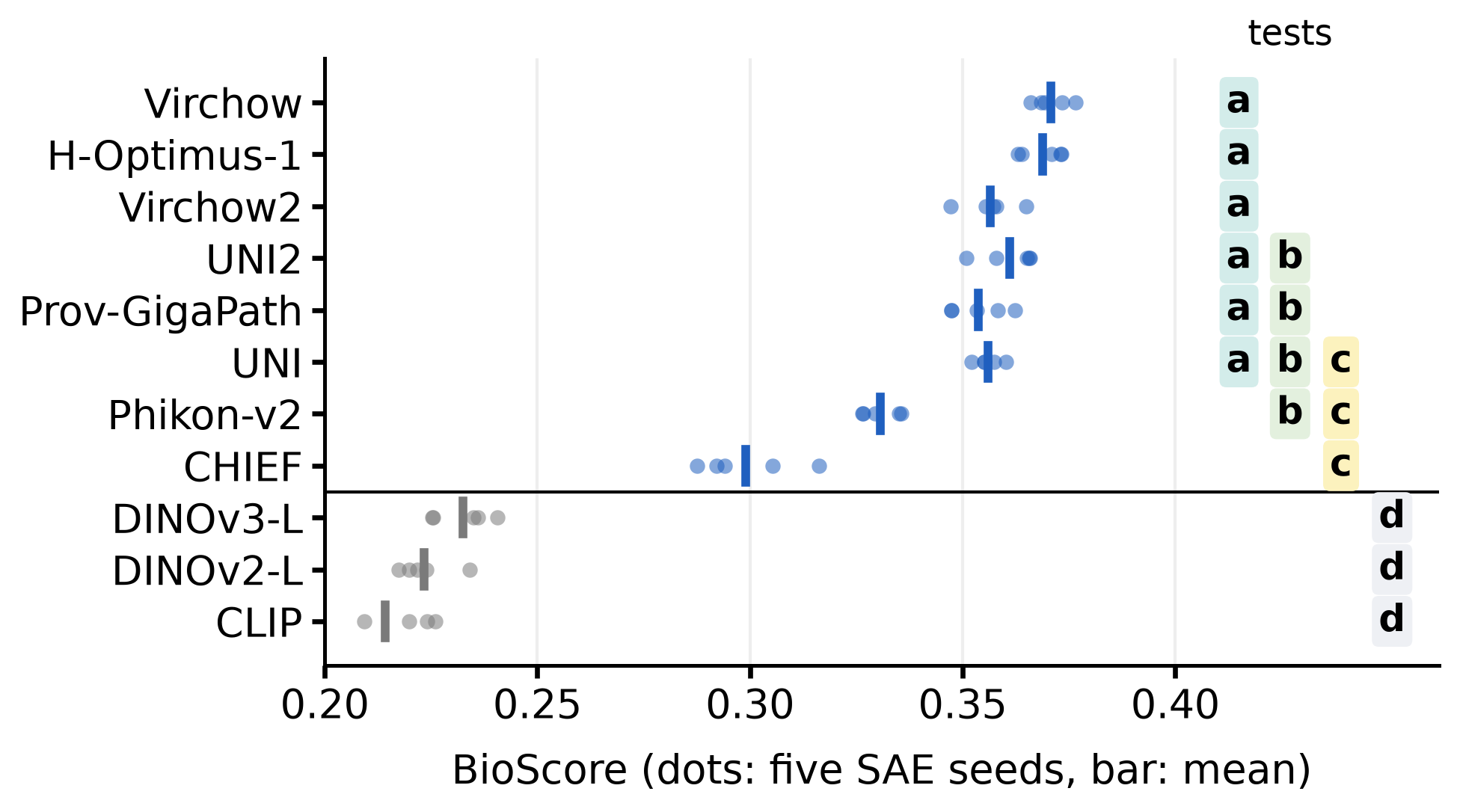}
  \caption{\textbf{BioScore per seed and test tiers.}
  BioScore per SAE seed (dots) and mean (bar), with the compact letter display of \cref{fig:win}. Full results in \cref{tab:main}.}
  \label{fig:ranking}
  \vspace{-0.3in}
\end{figure}

\paragraph{Pathology model versus natural-image model ranking}
The eight pathology foundation models score significantly above natural-image models (0.299--0.371 \vs 
0.214--0.232; \cref{fig:ranking}); the smallest margin, between CHIEF and DINOv3-L (0.067), is several times
the seed standard deviation.

\paragraph{The strongest pathology models are close}
Virchow, H-Optimus-1, UNI2, Virchow2, UNI and Prov-GigaPath lie within 0.017 of each other (\cref{fig:ranking}).
To tell which differences hold, we compare encoders identity by identity on the quality averaged over seeds
(\cref{fig:win}): the win rate of A over B is the share of the 52 identities on which A has the higher quality, and a
Wilcoxon signed-rank test over identities, with a Holm correction for the 55 pairs, decides significance.
These tests weigh every identity equally, whereas raw BioScore balances lineages.
All 24 pairs of a pathology and a natural-image model differ significantly (win rates 0.77--0.94). Additionally, the compact letter display of \cref{fig:win,fig:ranking} (encoders that share a letter do not differ significantly)
gives three levels among pathology models: Virchow, H-Optimus-1 and Virchow2 (a), UNI2 and Prov-GigaPath (ab) and UNI
(abc) form the leading group; Phikon-v2 (bc, 0.331) differs from the top 3; and CHIEF (c,
0.299) differs from every leading model except UNI.

\subsection{Breakdown by lineage and cell identity}
\label{sec:breakdown}

\paragraph{Pathology models gain most outside the immune lineage}
Averaged over identities, the quality $Q$ of pathology models exceeds that of natural-image models by 0.20 for
parenchymal identities, 0.18 for epithelial, 0.15 for stromal and 0.11 for cancer identities, against 0.08 for immune
identities and 0.02 for the two endothelial ones (\cref{tab:supp_lineage}).
Immune identities are harder on average for every encoder (\eg, 0.23 for Virchow against 0.40 on  other identities). We hypothesize that many immune cells are too small and scattered to dominate the embedding at 20x.

\paragraph{At the level of identities, gap between models varies}
\cref{fig:radar} shows  correlation of the matched group for a selection of single identities (all 52 identities in \cref{fig:supp_radar_all}).
We can isolate model differences on single identities: on a pancreatic section, the  Virchow group follows the scattered islets ($r=0.67$), whereas the CHIEF group is simply broadly active ($r=-0.27$) and DINOv3-L is silent.
On an ovarian section, Virchow2 and DINOv2-L correctly follow the carcinoma, while Phikon-v2 activation is too sparse \cref{fig:maps_main}.
On another note, the gap between pathology and natural-image models is  not uniform. The mean correlation of pathology models exceeds that of natural-image models by more than 0.1 for 24 of the 52 identities, most significantly on pancreatic islet endocrine cells (0.57 \vs 0.09), and stem-like cancer cells (\eg, SOX9+ CD44+, 0.54 \vs 0.18).
Inversely, the difference is small (max. 0.05) for 14 identities (\cref{tab:supp_identity_gap}), for distinct reasons.
On 7 of them, both families follow the identity well ($\rho\geq0.35$ ; \eg T cells, adipocytes, ovarian carcinoma cells).
On 4 others, neither model family does: iNOS+ macrophages, mast cells, cDC1 and CD56+ NK cells reach mean
correlations of only 0.02--0.04 for pathology models.

%% file: sec/6_conclusion.tex
\section{Conclusion}
\label{sec:conclusion}

Downstream benchmarks and supervised probes show what can be decoded from an embedding; we asked instead if an
encoder has a unit that follows a cell identity over the tissue.
With paired H\&E and Xenium transcriptomics as the reference, BioScore finds such units without labels, and its ranking
of encoders is stable over SAE seeds and recipes and agrees with a supervised probe (Kendall $\tau=0.82$).
Each pathology model represents identities in our panel better than every natural-image model, with largest
gap on parenchymal and epithelial identities (0.20 and 0.18 in quality, against 0.08 for immune identities).
Among pathology models, the overall score separates tiers rather than a single best model, and the differences lie at
the level of single identities.
Some identities, such as B cells, T cells and ovarian carcinoma cells, are followed as well by natural-image models;
others, such as mast cells and CD56+ NK cells, by no encoder.
A score per cell identity thus tells not only which encoder is stronger, but also which biology each encoder misses.

%% file: sec/7_llm_statement.tex
\section*{LLM Usage Statement}

We used Anthropic Claude Opus 5.5 to assist with manuscript polishing, grammar checks, code implementations and debugging. For manuscript polishing and grammar checks, text was reviewed and modified by the LLM, with human detailed proofreading afterwards. For code implementation and debugging, the LLM assisted in implementing BioScore, with human review afterward. We are responsible for all experiments, citations, and interpretation of visual results.

%% file: sec/X_suppl.tex
\clearpage
\setcounter{page}{1}
\onecolumn
\begin{center}
  \Large\textbf{\thetitle}\\
  \vspace{0.5em}Supplementary Material
\end{center}
\vspace{1.0em}
\setcounter{section}{0}
\renewcommand{\thesection}{\Alph{section}}
\renewcommand{\theHsection}{supp.\Alph{section}}

\FloatBarrier
\section{Sets construction}
\label{sec:supp_data}

\paragraph{Gene data construction}
For patch $p$ and gene $i$, the count $c_{pi}$ is the number of Xenium transcripts of gene $i$ whose position,
projected onto the H\&E image, falls inside $p$; we keep transcripts with reported quality value $\geq20$, assigned to a cell or
not, negative-control and blank codewords excluded.
The patch depth $T_p=\sum_i c_{pi}$ measures how much RNA was
captured in the patch, which varies with cell density and tissue quality.
We use in the score only patches with at least 20 genes ($T_p\geq20$), since patches with few transcripts give unstable ratios, leading to the exclusion of 3.6\% of the 20$\times$
patches.

\paragraph{Pretraining set construction}
\Cref{tab:supp_pretrain} details the content of the pretraining set (\cref{sec:data}).

\FloatBarrier
\section{Cell-identity panel construction}
\label{sec:supp_panel}

\paragraph{Cell-identity panel breakdown}
\Cref{tab:supp_panel} lists the 52 cell identities of BioScore (\cref{sec:panel}), with, for each, its number of CellMarker
signatures, the number of sections on which it is scored, and a sample of its strongest biomarkers. We additionally provide a detailed table of the curation steps of this panel from the original CellMarker 3.0 data (\cref{tab:supp_funnel}), described in \cref{sec:panel}.
\newcounter{savedtable}
\begin{table}[!htbp]
\begin{minipage}[t]{0.57\linewidth}
  \centering
  \caption{\textbf{Pretraining set.}
  Scoring sections (STHELAR), TCGA / CGCI slides and their projects (BLGSP: CGCI), and patches at 20$\times$ per organ.}
  \label{tab:supp_pretrain}
  \footnotesize
  \setlength{\tabcolsep}{3pt}
    \begin{tabular}{@{}lccll@{}}
    \toprule
    Organ & STHELAR & TCGA/CGCI & TCGA/CGCI source & Patches \\
    \midrule
    Brain      & 1 & 9 & GBM, LGG          & 99,976 \\
    Breast     & 4 & 3 & BRCA              & 101,630 \\
    Cervix     & 1 & 7 & CESC              & 102,122 \\
    Colon      & 2 & 7 & COAD              & 100,000 \\
    Kidney     & 2 & 4 & KICH, KIRC, KIRP  & 94,732 \\
    Liver      & 2 & 5 & LIHC              & 99,045 \\
    Lung       & 2 & 5 & LUAD, LUSC        & 99,526 \\
    Lymph node & 1 & 8 & BLGSP, DLBC, SKCM & 100,385 \\
    Ovary      & 2 & 4 & OV                & 98,033 \\
    Pancreas   & 3 & 5 & PAAD              & 99,368 \\
    Prostate   & 1 & 5 & PRAD              & 99,125 \\
    Skin       & 4 & 4 & SKCM              & 99,401 \\
    Tonsil     & 2 & 6 & HNSC              & 99,778 \\
    \midrule
    Total      & 27 & 72 &                 & 1,293,121 \\
    \bottomrule
  \end{tabular}
\end{minipage}\hfill
\begin{minipage}[t]{0.41\linewidth}
  \centering
  \setcounter{savedtable}{\value{table}}\setcounter{table}{21}
  \caption{\textbf{Curation funnel of the panel.}
  Curation steps are detailed in \cref{sec:panel}.}
  \label{tab:supp_funnel}
  \footnotesize
    \begin{tabular}{@{}lr@{}}
    \toprule
    Step & Remaining \\
    \midrule
    CellMarker 3.0 biomarkers & 1.78M entries \\
    Source filtering & 137k entries \\
    Organ restriction & 99.7k entries \\
    Consensus & 643 signatures \\
    Panel compatibility & 404 signatures \\
    Presence & 283 signatures \\
    Localization curation & 104 signatures, 52 identities \\
    \bottomrule
  \end{tabular}
  \setcounter{table}{\value{savedtable}}
\end{minipage}
\end{table}

\setcounter{savedtable}{\value{table}}\setcounter{table}{26}
\begin{table}[!htbp]
  \centering
  \caption{\textbf{Signature sets lineage breakdown per localization threshold.}
  Lineage breakdown of the sets formed at each co-localization threshold: number of
  signatures left, number of
  single-signature sets, size of the largest set, and number of sets per majority lineage.}
  \label{tab:supp_threshold_sets}
  \small
  \setlength{\tabcolsep}{3.5pt}
  \resizebox{0.95\linewidth}{!}{%
  \input{tables/supp_threshold_sets.tex}}
  \vspace{0.1in}
\end{table}
\setcounter{table}{\value{savedtable}}

\paragraph{Co-localization threshold ablation}
To ensure having only separable cell identities in our panel, the last step of our curation protocol groups signatures into sets whose maps correlate at $r\geq0.6$ (\cref{sec:panel}).
This threshold sets how finely the panel splits spatially similar identities.
At 0.2, 0.3, 0.4, 0.5, 0.6, 0.7 and 0.8, the 283 present signatures form 24, 30, 35, 44, 55, 68 and 93 sets
(\cref{tab:supp_threshold_sets}), and the immune lineage undergoes the most grouping (from 11 at 0.2 up to 64 at 0.8). We argue that 0.6 allows for a proper compromise between grouping identities too coarsely and keeping many almost-duplicate identities.
Between 0.6 and 0.4, related identities whose maps peak in different places merge, which we find undesirable.
Below 0.4, merges start to cross lineages.
Above 0.6, identities that act in practice as duplicates appear, and they become frequent at 0.8.
\Cref{fig:supp_threshold_maps} gives an example of each case.

\begin{table}[h!]
  \centering
  \caption{\textbf{Cell-identity panel.}
  Lineage, number of CellMarker signatures merged into the identity, number of sections on which it is scored, inclusion
  in the supervised-probe comparison of \cref{sec:downstream} ($\ast$: hotspot on at least two sections, 27
  identities), and the genes of its strongest signature (highest hotspot score). Rows grouped as in \cref{fig:supp_radar_all}.}
  \label{tab:supp_panel}
  \scriptsize
  \setlength{\tabcolsep}{4pt}
  \resizebox{0.79\linewidth}{!}{%
  \begin{tabular}{@{}llrrcl@{}}
    \toprule
    Cell identity & Lineage & Sign. & Sect. & Probe & Genes of the strongest signature \\
    \midrule
    \multicolumn{6}{@{}l}{\textit{Lymphoid} (13)} \\
    Antibody-secreting plasma cell & immune & 2 & 5 & $\ast$ & MZB1, IGKC, JCHAIN \\
    B cell & immune & 5 & 24 & $\ast$ & MS4A1, CD79A, CD19 \\
    CD138+ BCMA+ plasma cell & immune & 1 & 5 & $\ast$ & SDC1, CD38, MZB1, TNFRSF17 \\
    CD56+ NK cell & immune & 1 & 8 &  & NCAM1, NKG7, GNLY \\
    Checkpoint-high exhausted T cell & immune & 1 & 26 & $\ast$ & PDCD1, CTLA4, HAVCR2 \\
    Cytotoxic T cell & immune & 3 & 9 & $\ast$ & CD8A, GZMB, PRF1, IFNG, GZMA \\
    Regulatory T cell & immune & 4 & 25 & $\ast$ & FOXP3, IL2RA, CTLA4, TIGIT \\
    T cell & immune & 19 & 26 & $\ast$ & CCR7, SELL, TCF7 \\
    T follicular helper cell & immune & 2 & 13 & $\ast$ & CXCR5, PDCD1, CXCL13 \\
    T helper17 cell & immune & 1 & 8 &  & RORC, CCR6, IL7R, KLRB1 \\
    Tissue-resident memory T cell & immune & 1 & 1 &  & CD3E, CD3D, IL7R, CD4, PTPRC, CD8A, \dots \\
    TOX-driven exhausted T cell & immune & 1 & 13 & $\ast$ & TOX, LAG3, PDCD1 \\
    Tumor-reactive exhausted T cell & immune & 2 & 14 & $\ast$ & PDCD1, TIGIT, CTLA4, LAG3, TOX, CXCL13, \dots \\
    \midrule
    \multicolumn{6}{@{}l}{\textit{Myeloid} (12)} \\
    CD16+ monocyte & immune & 1 & 7 & $\ast$ & FCGR3B, FCGR3A, CD14 \\
    Inflammatory IL-1B macrophage & immune & 1 & 12 & $\ast$ & CD80, CD86, IL1B \\
    iNOS+ classically activated macrophage & immune & 1 & 7 &  & NOS2, CD86, CXCL10 \\
    Interferon-activated macrophage & immune & 1 & 12 & $\ast$ & CD86, CD68, CXCL10, CD80 \\
    Lipid-associated macrophage & immune & 3 & 5 &  & CD14, CD74, FCGR3A, CD163, CD68 \\
    M2-like macrophage & immune & 8 & 21 & $\ast$ & CD163, CD68, MRC1 \\
    Mast cell & immune & 1 & 3 & $\ast$ & TPSAB1, KIT, CPA3 \\
    Mature neutrophil & immune & 1 & 7 &  & CEACAM8, FCGR3B, MME \\
    Monocyte-derived macrophage & immune & 1 & 5 &  & CD14, CD68, CD74, PTPRC, ITGAM, AIF1 \\
    Perivascular C1Q+ macrophage & immune & 1 & 3 &  & C1QC, MRC1, F13A1, FMN1 \\
    S100A8/A9+ inflammatory neutrophil & immune & 1 & 1 &  & CEACAM8, S100A12, S100A9 \\
    Tissue-resident macrophage & immune & 2 & 15 & $\ast$ & CD68, CD163, TIMD4, CD14, MARCO \\
    \midrule
    \multicolumn{6}{@{}l}{\textit{Dendritic cells} (6)} \\
    cDC1 & immune & 2 & 7 &  & CD1C, LAMP3, CLEC9A, XCR1 \\
    cDC2 & immune & 1 & 13 & $\ast$ & CD1C, ITGAX, FCER1A, CLEC10A \\
    Langerhans cell & immune & 1 & 1 &  & CD207, CD1A, S100B \\
    Mature regulatory DC (mregDC) & immune & 1 & 21 & $\ast$ & LAMP3, CD1C, CCL19 \\
    pDC, interferon-producing & immune & 2 & 6 & $\ast$ & CLEC4C, IL3RA, LILRA4 \\
    pDC, lineage program & immune & 1 & 7 & $\ast$ & IRF8, BCL11A, CLEC4C \\
    \midrule
    \multicolumn{6}{@{}l}{\textit{Cancer} (9)} \\
    CD44+ KIT+ cancer stem cell & cancer & 1 & 7 &  & CD44, PROM1, KIT \\
    Epithelial cancer stem cell & cancer & 1 & 7 &  & PROM1, CD44, ALCAM, EPCAM \\
    Hormone-receptor+ carcinoma cell & cancer & 1 & 24 &  & ERBB2, EPCAM, ESR1 \\
    Neural stem-like cancer cell & cancer & 1 & 7 &  & NES, PROM1, SOX2 \\
    Neural-crest-like tumor cell & cancer & 1 & 7 &  & CSPG4, MCAM, NGFR \\
    Ovarian carcinoma cell & cancer & 1 & 7 &  & MUC16, PAX8, EPCAM \\
    Pluripotency-like cancer stem cell & cancer & 2 & 7 &  & PROM1, CTNNB1, SOX2 \\
    Proliferating glial-like tumor cell & cancer & 1 & 6 &  & EGFR, SOX2, TOP2A, CDK4, PTPRZ1 \\
    SOX9+ CD44+ stem-like cancer cell & cancer & 1 & 7 &  & SOX2, PROM1, SOX9, CD44 \\
    \midrule
    \multicolumn{6}{@{}l}{\textit{Other} (12)} \\
    Adipocyte & parenchyma & 2 & 9 & $\ast$ & PLIN1, PPARG, ADIPOQ \\
    Basal cell & epithelial & 2 & 5 & $\ast$ & KRT14, KRT5, ACTA2, MYLK \\
    Capillary pericyte & stromal & 1 & 7 &  & PDGFRB, RGS5, CSPG4, MCAM, TRPC6 \\
    Endothelial cell & endothelial & 4 & 17 & $\ast$ & PECAM1, VWF, CD34 \\
    FOXI1+ ionocyte & epithelial & 1 & 9 & $\ast$ & CFTR, FOXI1, ASCL3 \\
    Lymphatic endothelial cell & endothelial & 1 & 3 &  & PROX1, CCL21, LYVE1 \\
    MYH11+ perivascular cell & stromal & 1 & 3 & $\ast$ & MYH11, MCAM, PDGFRB, RGS5 \\
    Neurovascular unit (endothelium + pericyte) & other & 1 & 3 &  & FLT1, CLDN5, PDGFRB \\
    Pancreatic islet endocrine cell & parenchyma & 1 & 10 & $\ast$ & CHGA, INS, SST \\
    RGS5+ mural cell & stromal & 1 & 5 &  & PDGFRB, RGS5, ACTA2 \\
    Schwann cell & parenchyma & 1 & 1 &  & S100B, MPZ, PLP1 \\
    Smooth muscle cell & stromal & 5 & 14 & $\ast$ & ACTA2, TAGLN, MYH11 \\
    \bottomrule
  \end{tabular}}
\end{table}

\FloatBarrier  
\newcounter{savedfigure}\setcounter{savedfigure}{\value{figure}}\setcounter{figure}{14}
\begin{figure}[p]
  \centering
  \includegraphics[width=\linewidth,height=0.83\textheight,keepaspectratio]{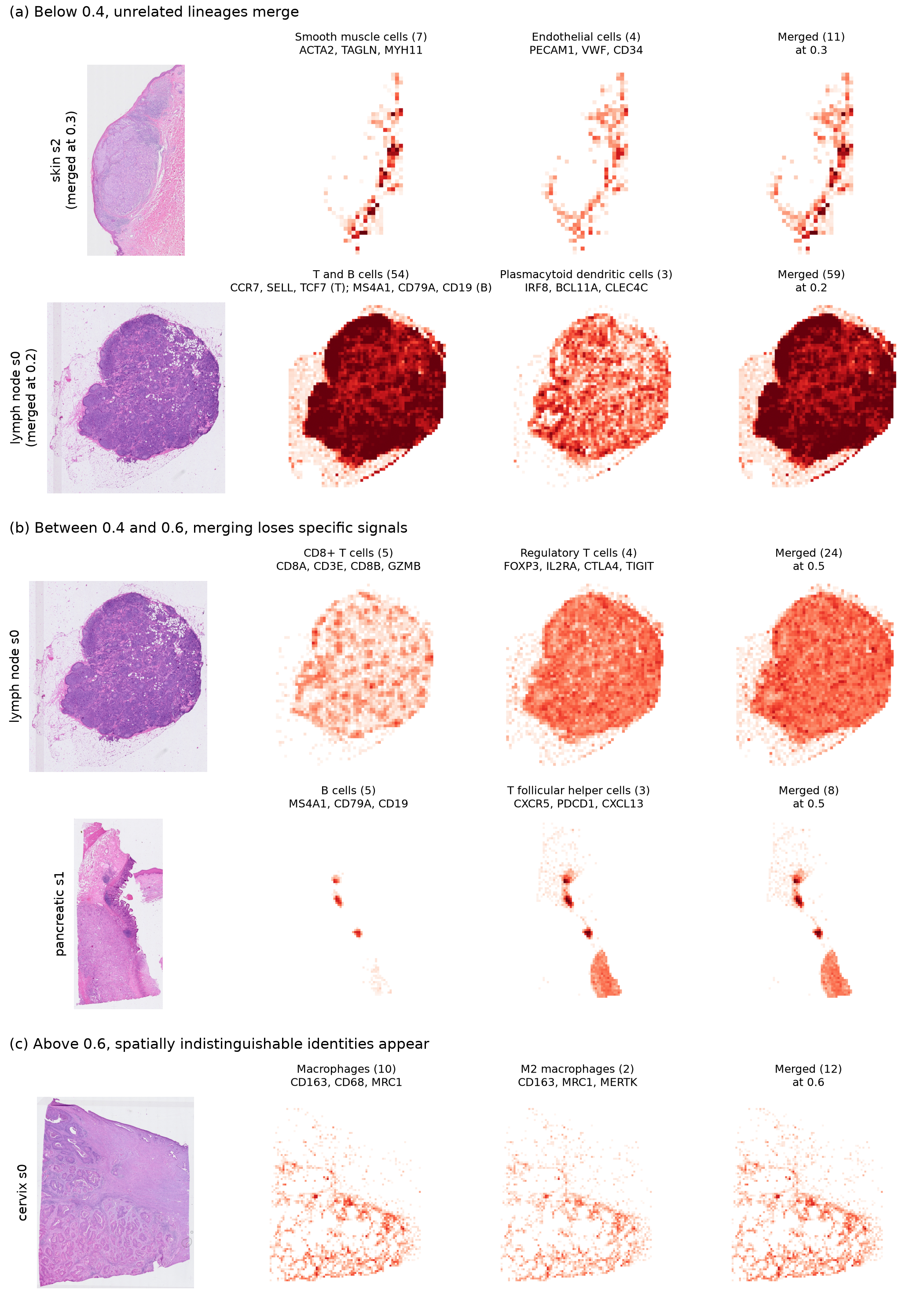}
  \caption{\textbf{Examples of grouping behaviors across co-localization thresholds.}
  Maps of signature sets, with the number of signatures in parentheses and genes of the strongest signature. (a) Below
  0.4, unrelated lineages start to merge despite clear spatial differences: smooth muscle and endothelial cells (merged
  at 0.3), and the T and B cell set with plasmacytoid dendritic cells (merged at 0.2). (b) Between 0.6 and 0.4, related
  identities with specific localizations start to merge. Although they share a similar spatial structure, their
  high-activation spots remain markedly distinct. CD8+ T cells, with clear hotspots over the lymph node, merge at 0.5
  with the broadly active regulatory T cells. Likewise, B cells and T follicular helper cells merge at 0.5, although
  only the latter are active at the bottom of the section. The merged identities lose the spatial specificity of their
  parents. (c) Conversely, above 0.6, spatially near-duplicate identities start to appear. Macrophages and M2
  macrophages are split despite sharing the same spatial structure and high-activation spots. Their merged version
  loses no signal and removes a duplicate from the panel.}
  \label{fig:supp_threshold_maps}
\end{figure}
\setcounter{figure}{\value{savedfigure}}

\FloatBarrier
\section{Encoders and SAE training}
\label{sec:supp_encoders}

\paragraph{Encoders}
All encoders receive the same $224\times224$\,px patches and their embeddings are stored as 16-bit floating-point numbers (float16).
DINOv2-L and Prov-GigaPath originally resize their input to 256\,px and crop its center back to 224\,px, which would keep only the central $\sim$196\,px of our patches. Since gene count is based on the whole patch, to ensure equal treatment between models we choose to remove this preprocessing step. Pathology encoders are trained at varying magnifications, reported in \cref{tab:encoders}. We compute BioScore at 20$\times$, the magnification at which most models are trained. The magnification study is in \cref{sec:magnification}.

\paragraph{SAE training and well-posedness}
We use the BatchTopK SAE~\cite{bussmann2024batchtopk} architecture as implemented in the \texttt{dictionary\_learning} library~\cite{marks2024dictionarylearning}, with $\times 8$ expansion factor, $k=64$, batch 2,048, 100 epochs, learning rate $2\cdot10^{-4}/\sqrt{n/2^{14}}$ with 1,000 warm-up steps and a decay over the last 20\% of training, decoder bias initialized at the geometric median, and the auxiliary loss of Gao \etal~\cite{gao2024scaling} ($\alpha=1/32$, $k_\mathrm{aux}=d/2$) on features that have not fired for 10M samples.
\Cref{tab:supp_health} reports, for the SAEs trained at 20$\times$, the share of dead features (never active over the 1.29M training patches), of dense features (active on more than 10\% of the patches) and of rare features (active on fewer than $10^{-5}$ of them), with the fraction of variance explained $R^2$ and the mean number of active features $L_0$.
The clustering procedure, based on decoder weights (\cref{sec:groups}), follows SAE-TM~\cite{girrbach2025sae}, which computes the selection of features that fire and their weights in the clustering from the corpus being analyzed.

\begin{table}[!htbp]
\begin{minipage}[t]{0.49\linewidth}
  \centering
  \caption{\textbf{Encoders.}
  Dimension of embedding space $d$, dimension of SAE latent space $n$ ($\times 8$ expansion factor), pretraining data and magnification of pretraining as reported by the authors (n.s.: not stated).}
  \label{tab:encoders}
  \small
  \setlength{\tabcolsep}{3pt}
  \begin{tabular}{@{}lrrll@{}}
    \toprule
    Encoder & $d$ & $n$ & Pretraining & Magn. \\
    \midrule
    UNI~\cite{chen2024uni}               & 1024 & 8192  & pathology & 20$\times$ \\
    UNI2-h~\cite{chen2024uni}            & 1536 & 12288 & pathology & n.s. \\
    Virchow~\cite{vorontsov2024virchow}  & 2560 & 20480 & pathology & 20$\times$ \\
    Virchow2~\cite{zimmermann2024virchow2} & 2560 & 20480 & pathology & 5--40$\times$ \\
    Prov-GigaPath~\cite{xu2024gigapath}  & 1536 & 12288 & pathology & 20$\times$ \\
    Phikon-v2~\cite{filiot2024phikonv2}  & 1024 & 8192  & pathology & 20$\times$ \\
    H-Optimus-1~\cite{scalbert2026hoptimus1} & 1536 & 12288 & pathology & 20$\times$ \\
    CHIEF~\cite{wang2024chief,wang2022ctranspath} & 768 & 6144 & pathology & 10$\times$ \\
    \midrule
    CLIP ViT-L/14~\cite{radford2021clip}   & 768  & 6144 & natural images & -- \\
    DINOv2 ViT-L/14~\cite{oquab2024dinov2} & 1024 & 8192 & natural images & -- \\
    DINOv3 ViT-L/16~\cite{simeoni2025dinov3} & 1024 & 8192 & natural images & -- \\
    \bottomrule
  \end{tabular}
\end{minipage}\hfill
\begin{minipage}[t]{0.47\linewidth}
  \centering
  \caption{\textbf{Well-posedness of SAE training.}
  Dead, dense ($>$10\% of patches) and rare ($<$$10^{-5}$) features, variance explained $R^2$ and mean number of
  active features $L_0$.}
  \label{tab:supp_health}
  \small
  \setlength{\tabcolsep}{4pt}
  \begin{tabular}{@{}lrrrcc@{}}
    \toprule
    Encoder & Dead\% & Dense\% & Rare\% & $R^2$ & $L_0$ \\
    \midrule
    Virchow       & 1.96 & 0.22 & 2.75 & 0.94 & 64.0 \\
    H-Optimus-1   & 0.00 & 0.45 & 0.00 & 0.86 & 64.0 \\
    UNI2          & 0.00 & 0.98 & 0.00 & 0.83 & 64.0 \\
    Virchow2      & 8.36 & 0.17 & 2.55 & 0.95 & 64.0 \\
    UNI           & 0.00 & 0.63 & 0.00 & 0.80 & 64.0 \\
    Prov-GigaPath & 0.00 & 0.09 & 0.00 & 0.84 & 64.0 \\
    Phikon-v2     & 0.00 & 0.64 & 0.00 & 0.89 & 64.0 \\
    CHIEF         & 4.66 & 2.94 & 1.46 & 0.94 & 64.0 \\
    \midrule
    DINOv3-L      & 2.79 & 1.87 & 23.09 & 0.92 & 64.0 \\
    DINOv2-L      & 8.46 & 1.49 & 23.92 & 0.93 & 64.0 \\
    CLIP          & 1.08 & 2.65 & 8.33 & 0.92 & 64.0 \\
    \bottomrule
  \end{tabular}
\end{minipage}
\end{table}

\FloatBarrier
\section{Sensitivity to $K$ and $\lambda$}
\label{sec:supp_sens}

BioScore has two hyperparameters: the number $K$ of feature groups into which the SAE features are clustered
(\cref{sec:groups}), and the weight $\lambda$ of the impurity penalty in the quality $Q$ (\cref{sec:score}).
The main setting uses $K=3000$ and $\lambda=3$, and \cref{sec:impurity,sec:stability} study how BioScore and the
ranking of the encoders change with each.
We provide the tabular results of the $\lambda$ and $K$ ablations of \cref{fig:lambda,fig:K} in
\cref{tab:lambda,tab:K}.

\begin{table}[!htbp]
\begin{minipage}[t]{0.48\linewidth}
  \centering
  \caption{\textbf{Effect of $\lambda$.}
  Mean BioScore, Kendall rank correlation $\tau$ against $\lambda=3$, and range of the distinct matched groups over encoders.}
  \label{tab:lambda}
  \small
  \setlength{\tabcolsep}{4pt}
  \begin{tabular}{@{}cccc@{}}
    \toprule
    $\lambda$ & BioScore & $\tau$ & Distinct groups \\
    \midrule
    0 & 0.382 & 0.96 & 34--43 \\
    0.25 & 0.374 & 1.00 & 35--44 \\
    0.5 & 0.366 & 1.00 & 36--45 \\
    1 & 0.353 & 1.00 & 38--46 \\
    1.5 & 0.341 & 1.00 & 40--47 \\
    2 & 0.331 & 1.00 & 43--49 \\
    2.5 & 0.323 & 1.00 & 45--51 \\
    3 (default) & 0.315 & 1.00 & 46--51 \\
    3.5 & 0.308 & 1.00 & 48--51 \\
    3.75 & 0.305 & 1.00 & 48--51 \\
    4 & 0.302 & 0.96 & 48--51 \\
    \bottomrule
  \end{tabular}
\end{minipage}\hfill
\begin{minipage}[t]{0.48\linewidth}
  \centering
  \caption{\textbf{Effect of $K$.}
  Mean BioScore, Kendall rank correlation $\tau$ against $K=3000$, and range of the distinct matched groups. $n$ means no clustering, one group per original SAE feature.}
  \label{tab:K}
  \small
  \setlength{\tabcolsep}{4pt}
  \begin{tabular}{@{}lccc@{}}
    \toprule
    $K$ & BioScore & $\tau$ & Distinct groups \\
    \midrule
    50 & 0.084 & 0.75 & 24--32 \\
    100 & 0.123 & 0.78 & 26--39 \\
    200 & 0.162 & 0.78 & 29--40 \\
    500 & 0.230 & 0.82 & 36--44 \\
    1000 & 0.276 & 0.93 & 40--48 \\
    2000 & 0.309 & 0.82 & 44--50 \\
    3000 (main) & 0.315 & -- & 46--51 \\
    $n$ (no grouping) & 0.319 & 1.00 & 47--51 \\
    \bottomrule
  \end{tabular}
\end{minipage}
\end{table}

\FloatBarrier
\section{Choice of correlation}
\label{sec:supp_corr}

\Cref{tab:supp_corr} compares BioScore computed with the Pearson correlation (ours) and with a Spearman rank correlation over all patches (for both the correlation $\rho$ and the impurity $U$).
With the rank correlation, the matched groups are more broadly active (\cref{fig:supp_corr_examples}; median active on 5.1\% of an identity's
patches, against 1.7\% for Pearson). Additionally, the gap between pathology and natural-image models drops from 0.126 to 0.072.
The actual ranking of the encoders changes little (Kendall $\tau=0.82$ over the 11 encoders, 0.71 among pathology models). We also provide examples of matched group activations in \cref{fig:supp_corr_examples}. 

\begin{table}[!htbp]
  \centering
  \caption{\textbf{Pearson vs.\ Spearman correlation.}
  BioScore, mean (standard deviation) over seeds, rank, and compact letter display of the paired tests, with the Pearson
  correlation on active patches (ours) and the Spearman correlation over all patches.}
  \label{tab:supp_corr}
  \footnotesize
  \setlength{\tabcolsep}{2.5pt}
  \begin{tabular}{@{}lcccccc@{}}
    \toprule
    & \multicolumn{3}{c}{Pearson, active} & \multicolumn{3}{c}{Spearman, all} \\
    \cmidrule(lr){2-4}\cmidrule(lr){5-7}
    Encoder & BioScore & Rank & Tests & BioScore & Rank & Tests \\
    \midrule
    Virchow & .371 (.004) & 1 & a & .301 (.004) & 1 & ab \\
    H-Optimus-1 & .369 (.005) & 2 & a & .289 (.008) & 3 & abc \\
    UNI2 & .361 (.007) & 3 & ab & .282 (.007) & 4 & ac \\
    Virchow2 & .356 (.006) & 4 & a & .293 (.005) & 2 & b \\
    UNI & .356 (.003) & 5 & abc & .269 (.008) & 6 & cd \\
    Prov-GigaPath & .354 (.007) & 6 & ab & .264 (.004) & 7 & c \\
    Phikon-v2 & .331 (.004) & 7 & bc & .276 (.006) & 5 & abc \\
    CHIEF & .299 (.012) & 8 & c & .251 (.002) & 8 & d \\
    \midrule
    DINOv3-L & .232 (.007) & 9 & d & .209 (.005) & 9 & e \\
    DINOv2-L & .223 (.006) & 10 & d & .205 (.005) & 11 & e \\
    CLIP & .214 (.014) & 11 & d & .205 (.004) & 10 & e \\
    \bottomrule
  \end{tabular}
\end{table}

\setcounter{savedfigure}{\value{figure}}\setcounter{figure}{12}
\begin{figure}[!htbp]
  \centering
  \includegraphics[width=\linewidth]{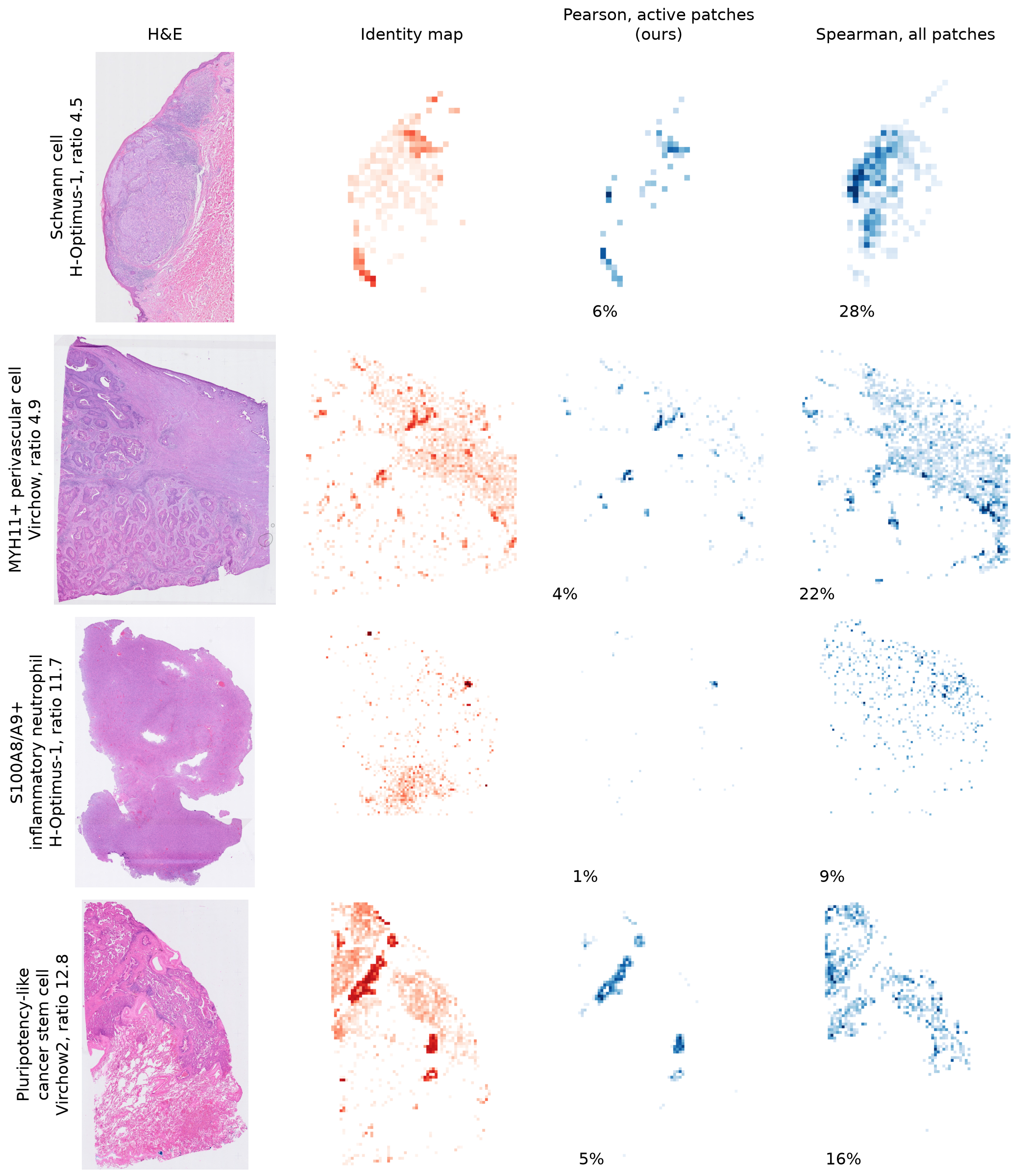}
  \caption{\textbf{Groups matched with Pearson vs.\ Spearman.}
  Sample of four identities: H\&E, identity spatial score map (red), and the matched group of SAE features (blue) with Pearson and with Spearman correlation. Percentage in each panel is the share of the section's patches where the group is active, in addition to the coverage ratio (row caption): share of the Spearman-matched group active patches, divided by the same share for the Pearson-matched group.
  The Pearson-based $Q$ rewards groups following hotspots of the identity map, whereas the Spearman-matched groups tend to be broader. We find the first behavior to be more desirable.}
  \label{fig:supp_corr_examples}
\end{figure}
\setcounter{figure}{\value{savedfigure}}

\FloatBarrier
\section{Supervised probe per identity}
\label{sec:supp_probe}

\Cref{tab:supp_probe_rho} compares, on the 27 identities of \cref{tab:probe}, the probe correlation with the
correlation $\rho$ of the BioScore matched group (0.21--0.37 against 0.33--0.54). The two rankings agree at Kendall $\tau=0.82$ (0.79 among pathology models).
\Cref{tab:supp_probe_examples} details the correlations $\rho$ behind the examples of \cref{sec:downstream}.

\begin{table}[!htbp]
\begin{minipage}[t]{0.40\linewidth}
  \centering
  \caption{\textbf{Probe and matched-group correlation on the 27 comparison identities.}
  Mean over the 27 comparison identities of the probe correlation, BioScore matched-group correlation $\rho$, and of its quality $Q$ (as in \cref{tab:probe}); rank in parentheses.}
  \label{tab:supp_probe_rho}
  \scriptsize
  \setlength{\tabcolsep}{3pt}
  \begin{tabular}{@{}lccc@{}}
    \toprule
    Encoder & Probe & $\rho$ & $Q$ \\
    \midrule
    Virchow & .537 (1) & .373 (.008) (1) & .295 (1) \\
    H-Optimus-1 & .513 (2) & .364 (.011) (2) & .289 (2) \\
    Virchow2 & .512 (3) & .359 (.011) (4) & .286 (4) \\
    UNI2 & .509 (4) & .360 (.007) (3) & .287 (3) \\
    Phikon-v2 & .487 (5) & .341 (.010) (7) & .256 (7) \\
    UNI & .484 (6) & .344 (.004) (5) & .267 (6) \\
    Prov-GigaPath & .482 (7) & .343 (.008) (6) & .274 (5) \\
    CHIEF & .436 (8) & .310 (.008) (8) & .241 (8) \\
    \midrule
    DINOv3-L & .346 (9) & .222 (.009) (10) & .171 (9) \\
    DINOv2-L & .344 (10) & .206 (.009) (11) & .156 (11) \\
    CLIP & .333 (11) & .227 (.008) (9) & .165 (10) \\
    \bottomrule
  \end{tabular}
\end{minipage}\hfill
\begin{minipage}[t]{0.58\linewidth}
  \centering
  \caption{\textbf{Probe vs BioScore on three example identities.}
  Probe: Pearson correlation of the out-of-fold predictions; BioScore: correlation $\rho$ of the matched group (mean over
  seeds). Gap: mean of the pathology minus mean of the natural-image models; effect size: gap over the standard deviation
  across the 11 encoders.}
  \label{tab:supp_probe_examples}
  \scriptsize
  \setlength{\tabcolsep}{3pt}
    \begin{tabular}{@{}lcccccc@{}}
    \toprule
    & \multicolumn{2}{c}{Islet endocrine} & \multicolumn{2}{c}{T cell} & \multicolumn{2}{c}{B cell} \\
    \cmidrule(lr){2-3}\cmidrule(lr){4-5}\cmidrule(lr){6-7}
    Encoder & Probe & BioScore & Probe & BioScore & Probe & BioScore \\
    \midrule
    Virchow & .29 & .71 & .90 & .66 & .87 & .56 \\
    H-Optimus-1 & .22 & .66 & .89 & .53 & .88 & .47 \\
    UNI2 & .34 & .62 & .90 & .56 & .90 & .47 \\
    Virchow2 & .09 & .54 & .90 & .64 & .88 & .51 \\
    UNI & .33 & .67 & .87 & .51 & .84 & .42 \\
    Prov-GigaPath & .10 & .51 & .88 & .56 & .88 & .47 \\
    Phikon-v2 & .24 & .56 & .88 & .71 & .86 & .59 \\
    CHIEF & .16 & .30 & .83 & .73 & .81 & .69 \\
    \midrule
    DINOv3-L & .02 & .12 & .76 & .65 & .77 & .52 \\
    DINOv2-L & .13 & .08 & .74 & .53 & .75 & .46 \\
    CLIP & .08 & .06 & .76 & .52 & .77 & .59 \\
    \midrule
    Gap & .15 & .49 & .13 & .04 & .10 & .00 \\
    Effect size & 1.33 & 1.94 & 2.04 & 0.54 & 1.90 & $-$0.03 \\
    \bottomrule
  \end{tabular}
\end{minipage}
\end{table}

\Cref{tab:supp_probe} provides the complete probe results for each of the 27 comparison identities and 11 encoders.

\begin{table}[!htbp]
  \centering
  \caption{\textbf{Probe and BioScore per identity.}
  For each of the 27 identities and each encoder, correlation of the probe's out-of-fold predictions with the identity map. Additionally, Spearman correlation between the orders of the 8 pathology models by the probe and by BioScore, i.e. the quality $Q$ of the matched group (median 0.26). In bold, the examples of \cref{sec:downstream}. Rows
  grouped as in \cref{fig:supp_radar_all}.}
  \label{tab:supp_probe}
  \scriptsize
  \setlength{\tabcolsep}{3pt}
  \input{tables/supp_probe_identity.tex}
\end{table}

\FloatBarrier
\section{SAE recipes}
\label{sec:supp_recipe}

\Cref{tab:supp_recipe} gives the BioScore of each encoder for each SAE recipe of \cref{tab:recipe}, without grouping ($K=n$, since the $\times4$ expansion factor is not compatible with $K=3000$ for all encoders). Changing training hyperparameters moves individual encoders by at most a few hundredths and does not change which family leads. Per identity (\cref{tab:supp_identity_recipe_sd,tab:supp_identity_seed_sd}), the recipe moves BioScore no more than the seed does (standard deviation 0.020 over recipes, against 0.025 over seeds) and the encoder ranking on a single identity often agrees (mean Kendall $\tau$ 0.66).

\setcounter{savedtable}{\value{table}}\setcounter{table}{23}

\begin{table}[!htbp]
  \centering
  \caption{\textbf{Variation over SAE recipes per identity.}
  Standard deviation over the recipes of the BioScore of each identity (without grouping, $K=n$, mean over the five seeds), per encoder, with its mean and maximum over the 11 encoders, and the mean Kendall $\tau$ between the rankings of the 11 encoders on this identity in the 10 pairs of recipes, range of the identity's BioScore (mean over recipes) over the 11 encoders for comparison with std; rows sorted by the mean standard deviation within each group.}
  \label{tab:supp_identity_recipe_sd}
  \scriptsize
  \setlength{\tabcolsep}{1.4pt}
  \resizebox{0.95\linewidth}{!}{%
  \input{tables/supp_identity_recipe_sd.tex}}
\end{table}
\setcounter{table}{\value{savedtable}}

\begin{table}[!htbp]
  \centering
  \caption{\textbf{SAE recipes per encoder.}
  BioScore without grouping ($K=n$) per dictionary size and sparsity $k$.}
  \label{tab:supp_recipe}
  \normalsize
  \setlength{\tabcolsep}{3pt}
  \begin{tabular}{@{}lccccc@{}}
    \toprule
    & $\times8$, 64 & $\times4$, 64 & $\times16$, 64 & $\times8$, 32 & $\times8$, 128 \\
    \midrule
    Virchow & .378 & .378 & .378 & .366 & .391 \\
    H-Optimus-1 & .369 & .377 & .374 & .365 & .383 \\
    UNI2 & .365 & .372 & .361 & .352 & .374 \\
    Virchow2 & .362 & .366 & .365 & .362 & .372 \\
    UNI & .360 & .357 & .355 & .351 & .366 \\
    Prov-GigaPath & .357 & .349 & .351 & .353 & .363 \\
    Phikon-v2 & .334 & .330 & .331 & .337 & .336 \\
    CHIEF & .302 & .277 & .315 & .301 & .287 \\
    \midrule
    DINOv3-L & .235 & .208 & .249 & .245 & .225 \\
    DINOv2-L & .227 & .188 & .238 & .223 & .216 \\
    CLIP & .217 & .194 & .226 & .212 & .207 \\
    \bottomrule
  \end{tabular}
\end{table}

\FloatBarrier
\section{Main results and SAE seeds}
\label{sec:supp_seeds}

\Cref{tab:main} gives BioScore in the main setting, mean and standard deviation over the five SAE seeds, and
\cref{tab:supp_seeds} the rank of each encoder in each seed; the spread over seeds is shown in \cref{fig:ranking}.

Per identity, the seed has more influence than on BioScore (\cref{tab:supp_identity_seed_sd}).
The standard deviation over seeds is 0.025 on average over identities and encoders, against 0.003--0.014 for BioScore (\cref{tab:main}), and reaches 0.11--0.12 for a few rare identities and single encoders: CHIEF on pancreatic islet cells ranges from 0.14 to 0.43 over seeds, seemingly because a dedicated feature emerges in some seeds only.
The ranking of the encoders on a single identity is accordingly less stable across seeds (mean Kendall $\tau$ 0.52) than their ranking by BioScore (0.83, \cref{tab:supp_seeds}). Per-identity results should therefore be read as means over seeds, as a single seed value is unreliable for one identity.
We also provide a wider selection of activation maps (\cref{fig:supp_maps_more_1,fig:supp_maps_more_2}), extending \cref{fig:maps_main} to 8 more maps.

\setcounter{savedfigure}{\value{figure}}\setcounter{figure}{15}
\begin{figure}[p]
  \centering
  \includegraphics[width=\linewidth,height=0.85\textheight,keepaspectratio]{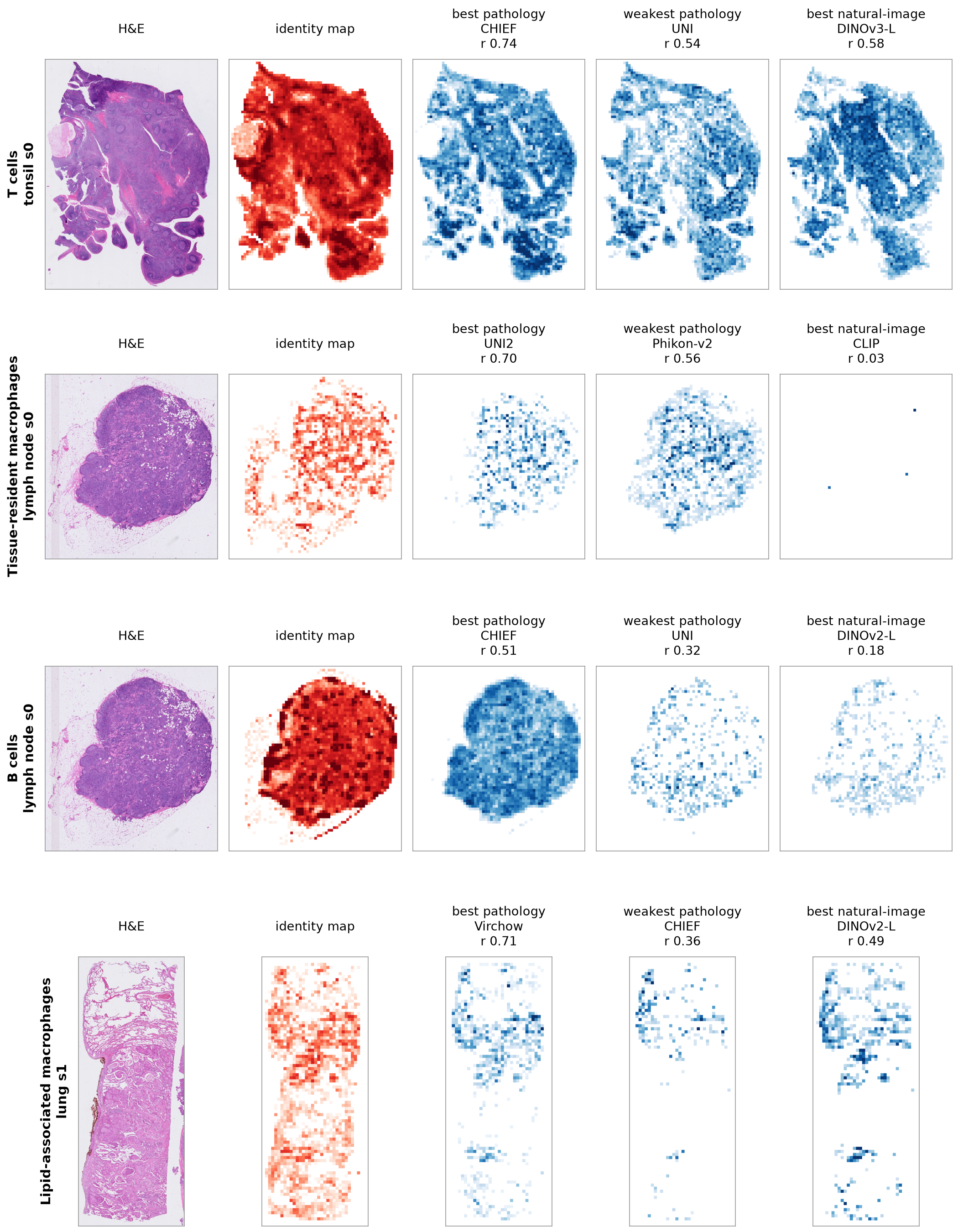}
  \caption{\textbf{More activation maps (1/2).}
  As \cref{fig:maps_main}: H\&E, identity map (red), and the matched groups (blue) of the best and the weakest
  pathology models and of the best natural-image model for the identity. Their
  correlation $r$ on the section is indicated; inactive meaning the group does not fire on the section.}
  \label{fig:supp_maps_more_1}
\end{figure}
\setcounter{figure}{\value{savedfigure}}
\setcounter{savedfigure}{\value{figure}}\setcounter{figure}{16}
\begin{figure}[p]
  \centering
  \includegraphics[width=\linewidth,height=0.85\textheight,keepaspectratio]{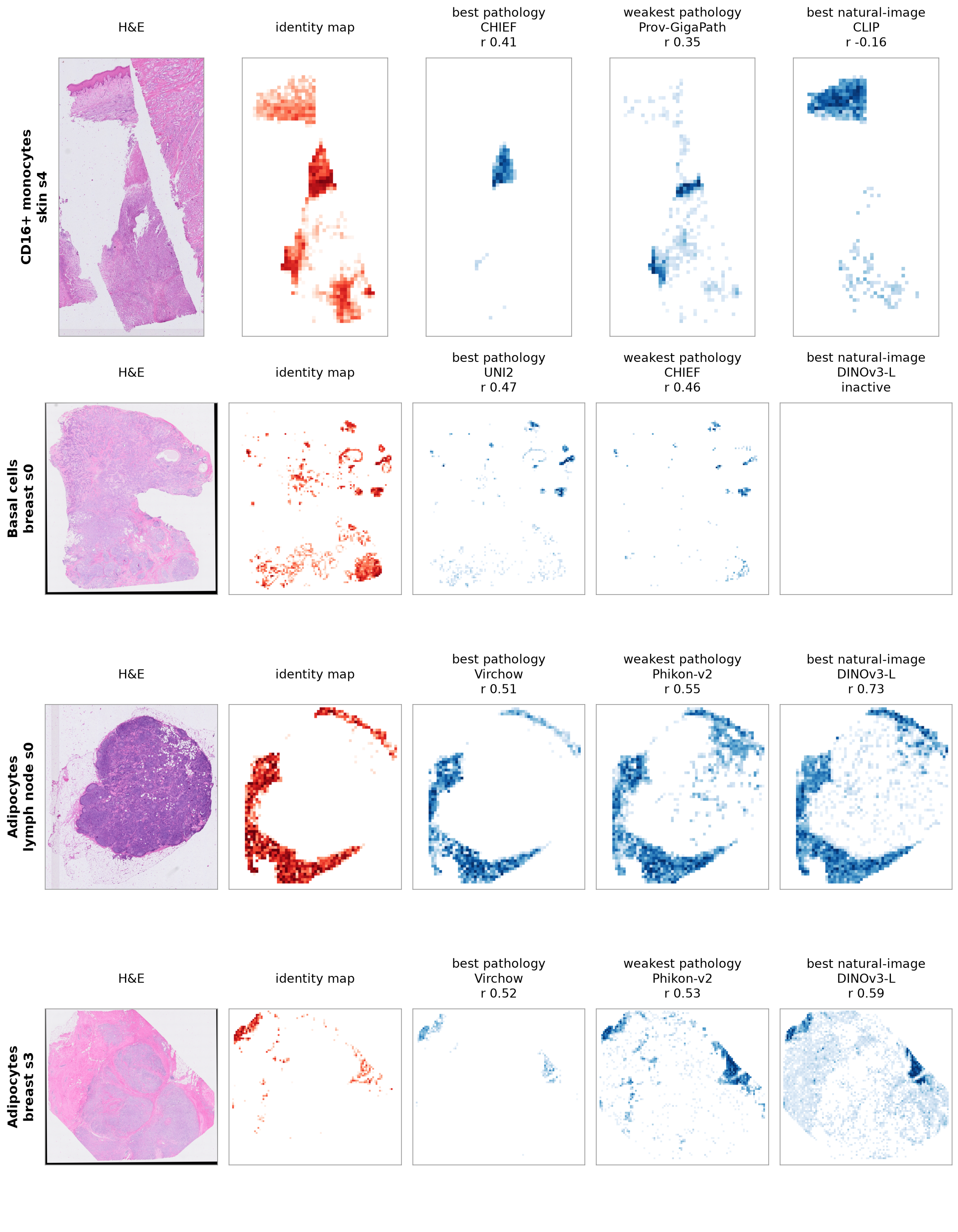}
  \caption{\textbf{More activation maps (2/2).}
  As \cref{fig:maps_main}: H\&E, identity map (red), and the matched groups (blue) of the best and the weakest
  pathology models and of the best natural-image model for the identity. Their
  correlation $r$ on the section is indicated; inactive meaning the group does not fire on the section.}
  \label{fig:supp_maps_more_2}
\end{figure}
\setcounter{figure}{\value{savedfigure}}

\begin{table}[!htbp]
\begin{minipage}[t]{0.30\linewidth}
  \centering
  \caption{\textbf{Main results.}
  BioScore, mean (standard deviation) over seeds, and compact letter display of the paired tests (\cref{fig:win}).
  Every pathology model scores above every natural-image model.}
  \label{tab:main}
  \normalsize
  \setlength{\tabcolsep}{4pt}
  \begin{tabular}{@{}lcc@{}}
    \toprule
    Encoder & BioScore & Tests \\
    \midrule
    Virchow & .371 (.004) & a \\
    H-Optimus-1 & .369 (.005) & a \\
    UNI2 & .361 (.007) & ab \\
    Virchow2 & .356 (.006) & a \\
    UNI & .356 (.003) & abc \\
    Prov-GigaPath & .354 (.007) & ab \\
    Phikon-v2 & .331 (.004) & bc \\
    CHIEF & .299 (.012) & c \\
    \midrule
    DINOv3-L & .232 (.007) & d \\
    DINOv2-L & .223 (.006) & d \\
    CLIP & .214 (.014) & d \\
    \bottomrule
  \end{tabular}
\end{minipage}\hfill
\begin{minipage}[t]{0.68\linewidth}
  \centering
  \caption{\textbf{BioScore per SAE seed.}
  BioScore of each encoder per seed (rank in parentheses), and mean over seeds; two seeds agree at Kendall $\tau=0.83$ on average.}
  \label{tab:supp_seeds}
  \normalsize
  \setlength{\tabcolsep}{2.5pt}
  \begin{tabular}{@{}lcccccc@{}}
    \toprule
    Encoder & 40 & 41 & 42 & 43 & 44 & Mean \\
    \midrule
    Virchow & .366 (2) & .368 (1) & .377 (1) & .373 (1) & .369 (1) & .371 \\
    H-Optimus-1 & .373 (1) & .363 (3) & .373 (2) & .371 (2) & .364 (3) & .369 \\
    UNI2 & .358 (3) & .366 (2) & .351 (6) & .366 (3) & .365 (2) & .361 \\
    Virchow2 & .347 (6) & .357 (6) & .365 (3) & .358 (5) & .355 (4) & .356 \\
    UNI & .355 (4) & .360 (4) & .357 (4) & .352 (6) & .355 (5) & .356 \\
    Prov-GigaPath & .347 (5) & .358 (5) & .353 (5) & .362 (4) & .347 (6) & .354 \\
    Phikon-v2 & .329 (7) & .326 (7) & .335 (7) & .336 (7) & .327 (7) & .331 \\
    CHIEF & .292 (8) & .316 (8) & .288 (8) & .294 (8) & .305 (8) & .299 \\
    \midrule
    DINOv3-L & .225 (10) & .241 (9) & .235 (9) & .236 (9) & .226 (9) & .232 \\
    DINOv2-L & .234 (9) & .222 (10) & .217 (11) & .220 (11) & .224 (10) & .223 \\
    CLIP & .209 (11) & .220 (11) & .224 (10) & .226 (10) & .192 (11) & .214 \\
    \bottomrule
  \end{tabular}
\end{minipage}
\end{table}

\setcounter{savedtable}{\value{table}}\setcounter{table}{22}
\begin{table}[!htbp]
  \centering
  \caption{\textbf{Variation over seeds per identity.}
  Standard deviation over the five SAE seeds of the BioScore of each identity (quality $Q$ of its matched group), per
  encoder, with its mean and maximum over the 11 encoders, the mean Kendall $\tau$ between the rankings of the 11
  encoders on this identity in the 10 pairs of seeds, and the range of the identity's BioScore (mean over seeds) over
  the 11 encoders; grouped as in \cref{fig:supp_radar_all} and sorted by the mean standard deviation within each group.}
  \label{tab:supp_identity_seed_sd}
  \scriptsize
  \setlength{\tabcolsep}{1.4pt}
  \input{tables/supp_identity_seed_sd.tex}
\end{table}
\setcounter{table}{\value{savedtable}}

\FloatBarrier
\section{Magnification}
\label{sec:supp_mag}
\label{sec:magnification}

At 40$\times$, the scoring set holds 443,004 patches of 60\,\textmu m.
We study magnification through four settings, named by the magnification of the SAE training data and of the scoring
set: 20$\times$/20$\times$ (main setting), 20$\times$/40$\times$ (the same SAEs, scored on 40$\times$ patches),
40$\times$/40$\times$, and mixed/40$\times$.
The 40$\times$ pretraining set uses the same 99 slides as the 20$\times$ one (\cref{tab:supp_pretrain}): each
20$\times$ patch contains four 40$\times$ patches, of which we draw one at random, so that both sets hold the same
number of patches at the same positions.
The mixed set keeps, on each slide, a random half of the 20$\times$ patches and replaces the other half by their drawn
40$\times$ patch.

\paragraph{Impact of scoring magnification vs training magnification}
We compare the settings in two steps.
First, the scoring magnification, with the same SAEs (20$\times$/20$\times$ against 20$\times$/40$\times$): scored on
40$\times$ patches (60\,\textmu m), every encoder loses BioScore, but unevenly.
Prov-GigaPath, CHIEF and Virchow lose 0.12--0.13, whereas UNI2 and Virchow2 lose 0.06--0.07 (Virchow2 was pretrained
on a range of magnifications, 5--40$\times$).
Second, the training magnification, with a fixed scoring at 40$\times$ (20$\times$/40$\times$, 40$\times$/40$\times$
and mixed/40$\times$): the three settings give similar rankings (Kendall $\tau$ 0.89--0.93; \cref{tab:supp_tau_settings}),
although training at 40$\times$ raises most pathology models (by up to 0.04).
In all three 40$\times$ settings, the best scores are 0.297--0.312 (\cref{tab:supp_mag}), the pathology models stay well
ahead of the natural-image models (gap 0.119--0.149) and UNI2 ranks first, but their rankings agree only moderately
with the 20$\times$/20$\times$ one (Kendall $\tau$ 0.71--0.78): the scoring magnification matters more than the training
magnification.

\begin{figure}[!htbp]
  \centering
  \includegraphics[width=\linewidth]{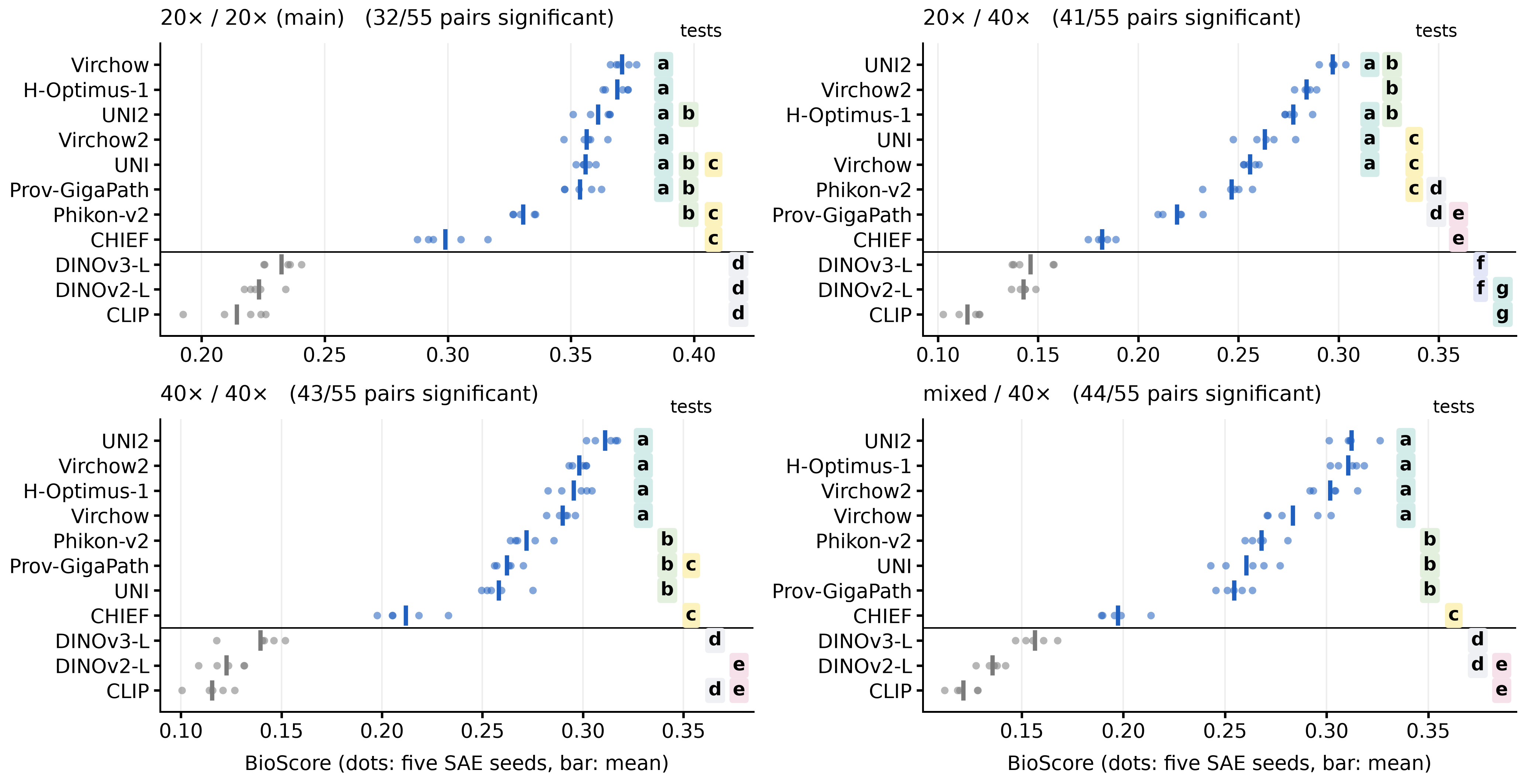}
  \caption{\textbf{BioScore per seed and test tiers in the four settings.}
  As \cref{fig:ranking}, per setting (training / scoring magnification); rows ordered by mean BioScore, letters from
  the paired tests of each setting. Additionally, the number of significantly different encoder pairs is reported.}
  \label{fig:supp_ranking_settings}
\end{figure}

\Cref{fig:supp_ranking_settings} repeats \cref{fig:ranking} in the four settings, with the paired tests recomputed in
each.
The tests separate more pairs when scoring at 40$\times$ (41--44 of 55, against 32 at 20$\times$/20$\times$), and
every pathology model differs significantly from every natural-image model in all four settings.

\begin{table}[!htbp]
\begin{minipage}[t]{0.55\linewidth}
  \centering
  \caption{\textbf{BioScore across magnification settings.}
  Mean (standard deviation) over seeds, by training and scoring magnification; best per column in bold.}
  \label{tab:supp_mag}
  \small
  \setlength{\tabcolsep}{3pt}
  \begin{tabular}{@{}lcccc@{}}
    \toprule
    Training & 20$\times$ & 20$\times$ & 40$\times$ & mixed \\
    Scoring & 20$\times$ & 40$\times$ & 40$\times$ & 40$\times$ \\
    \midrule
    Virchow & \textbf{.371} (.004) & .256 (.003) & .290 (.005) & .283 (.014) \\
    H-Optimus-1 & .369 (.005) & .277 (.006) & .295 (.009) & .311 (.007) \\
    UNI2 & .361 (.007) & \textbf{.297} (.005) & \textbf{.311} (.007) & \textbf{.312} (.009) \\
    Virchow2 & .356 (.006) & .284 (.004) & .298 (.004) & .302 (.010) \\
    UNI & .356 (.003) & .263 (.011) & .258 (.010) & .261 (.014) \\
    Prov-GigaPath & .354 (.007) & .219 (.009) & .262 (.006) & .255 (.007) \\
    Phikon-v2 & .331 (.004) & .247 (.009) & .272 (.009) & .268 (.008) \\
    CHIEF & .299 (.012) & .182 (.005) & .212 (.014) & .197 (.010) \\
    \midrule
    DINOv3-L & .232 (.007) & .146 (.011) & .140 (.013) & .157 (.008) \\
    DINOv2-L & .223 (.006) & .143 (.004) & .123 (.010) & .136 (.005) \\
    CLIP & .214 (.014) & .115 (.008) & .116 (.010) & .121 (.007) \\
    \bottomrule
  \end{tabular}
\end{minipage}\hfill
\begin{minipage}[t]{0.43\linewidth}
  \centering
  \caption{\textbf{Ranking agreement across magnification settings.}
  Kendall $\tau$ between the encoder rankings of the four settings (training / scoring magnification).}
  \label{tab:supp_tau_settings}
  \small
  \setlength{\tabcolsep}{3pt}
  \begin{tabular}{@{}lcccc@{}}
    \toprule
    & 20$\times$/20$\times$ & 20$\times$/40$\times$ & 40$\times$/40$\times$ & mixed/40$\times$ \\
    \midrule
    20$\times$/20$\times$ & -- & 0.75 & 0.71 & 0.78 \\
    20$\times$/40$\times$ & 0.75 & -- & 0.89 & 0.89 \\
    40$\times$/40$\times$ & 0.71 & 0.89 & -- & 0.93 \\
    mixed/40$\times$ & 0.78 & 0.89 & 0.93 & -- \\
    \bottomrule
  \end{tabular}
\end{minipage}
\end{table}

\Cref{tab:supp_tau_settings} compares the rankings of the four settings.

Per identity, the correlation of the matched group is lower at 40$\times$ for 78\% of the (identity, encoder) pairs, and the pathology models keep a
higher mean correlation than the natural-image models on 47 of the 52 identities (45 at 20$\times$).
\Cref{tab:supp_mag_identity} gives the BioScore of each identity at 40$\times$/40$\times$ and its difference with
20$\times$/20$\times$, per encoder.
Averaged over the encoders, the BioScore of 25 identities changes by less than 0.05 between the two settings.
The other 25 lose more than 0.05, most of all structural identities that the encoders recognized well at
20$\times$/20$\times$: adipocytes ($-$0.19, from 0.51), RGS5+ mural cells ($-$0.16, from 0.46), pancreatic islet cells
($-$0.16, from 0.44) and smooth muscle cells ($-$0.15, from 0.52).
Only two identities gain more than 0.05, CD56+ NK cells ($+$0.07) and FOXI1+ ionocytes ($+$0.05).

\setcounter{savedtable}{\value{table}}\setcounter{table}{25}
\begin{table}[!htbp]
  \centering
  \caption{\textbf{Identities at 40$\times$/40$\times$.}
  BioScore of each identity (quality $Q$ of the matched group, mean over five seeds) at 40$\times$/40$\times$, with its
  difference with 20$\times$/20$\times$ in parentheses, per encoder, and the mean difference over the 11 encoders. Rows
  grouped as in \cref{fig:supp_radar_all}.}
  \label{tab:supp_mag_identity}
  \tiny
  \setlength{\tabcolsep}{1.8pt}
  \resizebox{0.96\linewidth}{!}{%
  \input{tables/supp_mag_identity.tex}}
  \vspace{-0.3in}
\end{table}
\setcounter{table}{\value{savedtable}}

\begin{figure}[!htbp]
  \centering
  \includegraphics[width=\linewidth]{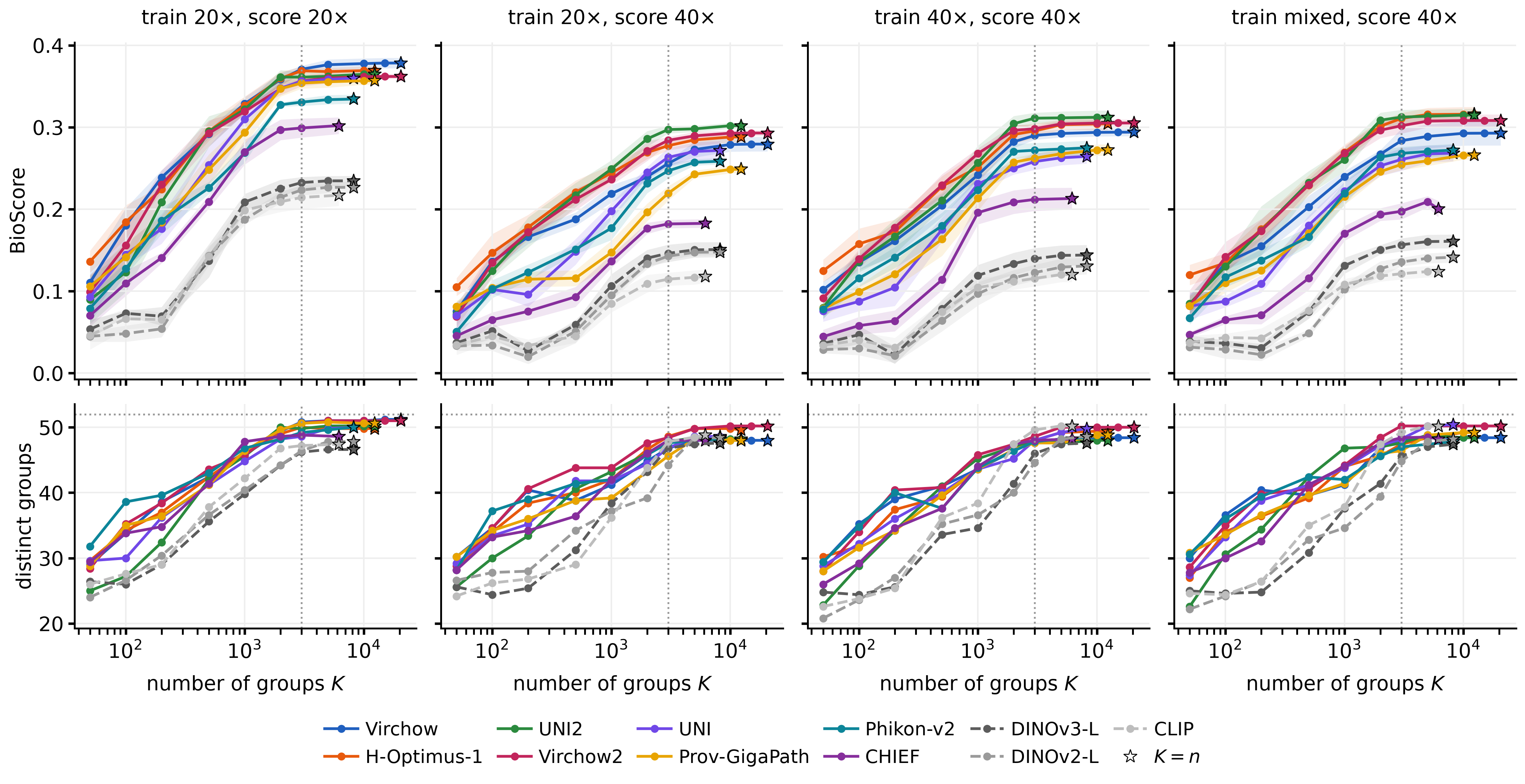}
  \caption{\textbf{Effect of $K$ in every setting.}
  BioScore (top) and distinct matched groups (bottom) against $K$; stars: $K=n$.}
  \label{fig:supp_K}
\end{figure}

\begin{figure}[!htbp]
  \centering
  \includegraphics[width=\linewidth]{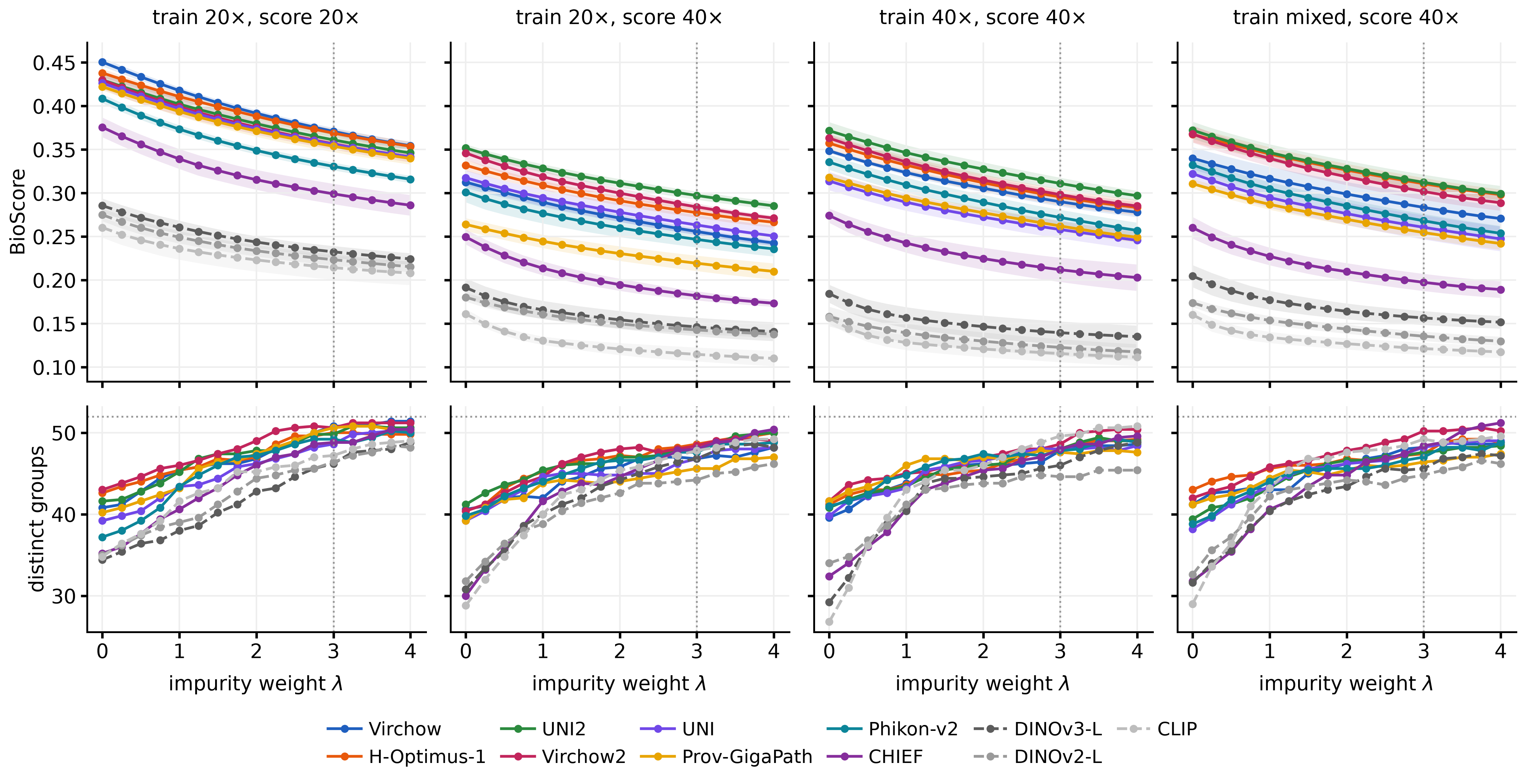}
  \caption{\textbf{Effect of $\lambda$ in every setting.}
  BioScore (top) and distinct matched groups (bottom) against $\lambda$.}
  \label{fig:supp_lambda}
\end{figure}

\Cref{fig:supp_K,fig:supp_lambda} show BioScore and ablation of $K$ and $\lambda$ in
the four settings.
BioScore levels off above $K\approx2{,}000$ in every setting, and the gap between pathology and natural-image models
is present at every $K$ and $\lambda$.

\FloatBarrier
\clearpage  
\section{Full results per identity}
\label{sec:supp_radar}

\begin{figure}[!htbp]
  \centering
  \includegraphics[width=\linewidth,height=0.86\textheight,keepaspectratio]{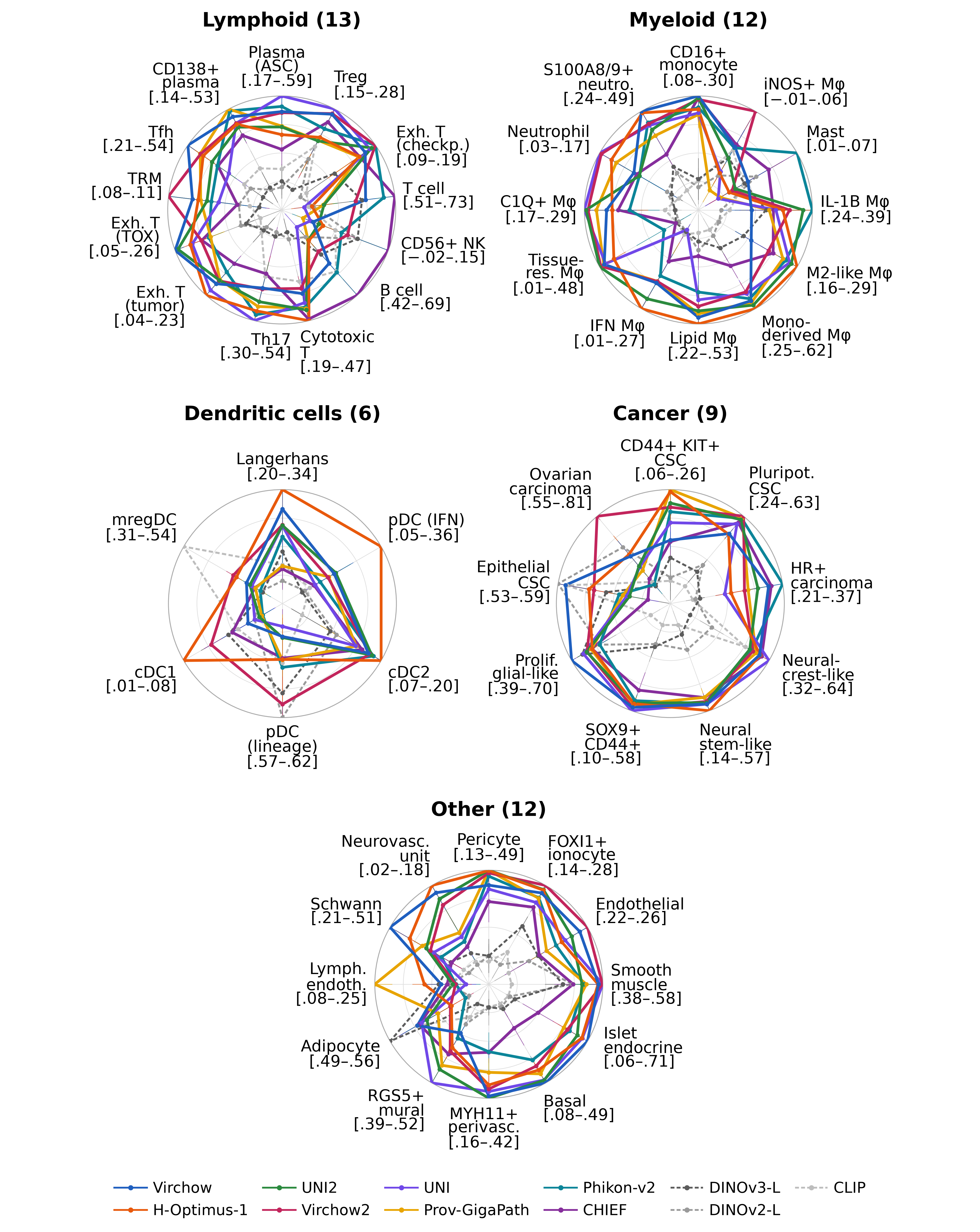}
  \caption{\textbf{All 52 identities.}
  As \cref{fig:radar}, with the immune identities split into lymphoid, myeloid and dendritic cells; values in
  \cref{tab:supp_identity_rho}.}
  \label{fig:supp_radar_all}
\end{figure}

The quality $Q$ selects the matched group and penalizes its impurity; here we first report the correlation $\rho$ of that
group, which shows more directly how closely it follows a given cell identity.
\Cref{fig:supp_radar_all,tab:supp_identity_rho} show the correlation of the matched group for all 52 identities, of which
\cref{fig:radar} shows a selection.
\Cref{fig:supp_radar_q,tab:supp_identity_q} give the same results with the quality $Q$.

\begin{table}[!htbp]
  \centering
  \caption{\textbf{All 52 identities, per encoder.}
  Correlation $\rho$ of the matched group, mean over five seeds; bold: best encoder.}
  \label{tab:supp_identity_rho}
  \scriptsize
  \setlength{\tabcolsep}{3.5pt}
  \resizebox{\linewidth}{!}{%
  \begin{tabular}{@{}lcccccccc@{\hspace{8pt}}ccc@{}}
  \toprule
  Identity & Virchow & H-Opt.-1 & UNI2 & Virchow2 & UNI & GigaPath & Phikon-v2 & CHIEF & DINOv3 & DINOv2 & CLIP \\
  \midrule
  \multicolumn{12}{@{}l}{\textit{Lymphoid} (13)} \\
  Antibody-secreting plasma cell & .52 & .41 & .45 & .51 & \textbf{.59} & .45 & .54 & .34 & .20 & .17 & .25 \\
  B cell & .56 & .47 & .47 & .51 & .42 & .47 & .59 & \textbf{.69} & .52 & .46 & .59 \\
  CD138+ BCMA+ plasma cell & .50 & .46 & .44 & .46 & .46 & \textbf{.53} & .52 & .40 & .15 & .14 & .24 \\
  CD56+ NK cell & .00 & .02 & .01 & .07 & .00 & $-$.02 & .06 & \textbf{.15} & .09 & .09 & .04 \\
  Checkpoint-high exhausted T cell & .18 & .17 & .18 & .19 & .18 & .17 & \textbf{.19} & .17 & .13 & .09 & .10 \\
  Cytotoxic T cell & .39 & \textbf{.47} & .44 & .37 & .42 & .43 & .43 & .47 & .19 & .22 & .35 \\
  Regulatory T cell & .27 & .23 & .23 & .27 & \textbf{.28} & .23 & .25 & .26 & .15 & .20 & .22 \\
  T cell & .66 & .53 & .56 & .64 & .51 & .56 & .71 & \textbf{.73} & .65 & .53 & .52 \\
  T follicular helper cell & \textbf{.54} & .47 & .43 & .48 & .36 & .49 & .41 & .41 & .21 & .26 & .29 \\
  T helper17 cell & .45 & .51 & .49 & .45 & \textbf{.54} & .50 & .52 & .41 & .31 & .30 & .42 \\
  TOX-driven exhausted T cell & \textbf{.26} & .22 & .25 & .19 & .22 & .17 & .18 & .19 & .08 & .09 & .05 \\
  Tissue-resident memory T cell & .10 & .10 & .10 & \textbf{.11} & .09 & .10 & .10 & .09 & .08 & .08 & .08 \\
  Tumour-reactive exhausted T cell & .20 & \textbf{.23} & .19 & .19 & .22 & .19 & .17 & .14 & .05 & .04 & .06 \\
  \midrule
  \multicolumn{12}{@{}l}{\textit{Myeloid} (12)} \\
  CD16+ monocyte & \textbf{.30} & .26 & .29 & .29 & .26 & .25 & .27 & .29 & .10 & .08 & .10 \\
  Inflammatory IL-1B macrophage & .29 & .34 & .37 & .35 & .33 & .32 & \textbf{.39} & .32 & .31 & .24 & .25 \\
  Interferon-activated macrophage & .19 & \textbf{.27} & .24 & .18 & .01 & .18 & .16 & .12 & .02 & .02 & .03 \\
  Lipid-associated macrophage & .50 & \textbf{.53} & .48 & .47 & .44 & .49 & .42 & .30 & .25 & .28 & .22 \\
  M2-like macrophage & .21 & \textbf{.29} & .28 & .24 & .27 & .26 & .27 & .25 & .20 & .17 & .16 \\
  Mast cell & .03 & .02 & .02 & .02 & .01 & .01 & \textbf{.07} & .05 & .03 & .04 & .03 \\
  Mature neutrophil & .11 & .15 & .10 & .16 & \textbf{.17} & .14 & .10 & .10 & .04 & .05 & .03 \\
  Monocyte-derived macrophage & .58 & \textbf{.62} & .60 & .54 & .55 & .59 & .57 & .42 & .33 & .27 & .25 \\
  Perivascular C1Q+ macrophage & .26 & .25 & .28 & .28 & \textbf{.29} & .27 & .23 & .24 & .17 & .17 & .19 \\
  S100A8/A9+ inflammatory neutrophil & \textbf{.49} & .48 & .43 & .45 & .44 & .41 & .45 & .35 & .31 & .29 & .24 \\
  Tissue-resident macrophage & .45 & .40 & \textbf{.48} & .47 & .45 & .46 & .10 & .03 & .02 & .01 & .02 \\
  iNOS+ classically activated macrophage & .03 & .00 & .02 & \textbf{.06} & .00 & $-$.01 & .02 & .03 & .02 & .00 & .02 \\
  \midrule
  \multicolumn{12}{@{}l}{\textit{Dendritic cells} (6)} \\
  Langerhans cell & .31 & \textbf{.34} & .29 & .29 & .29 & .22 & .27 & .22 & .25 & .20 & .23 \\
  Mature regulatory DC (mregDC) & .36 & .39 & .35 & .40 & .34 & .33 & .31 & .33 & .31 & .31 & \textbf{.54} \\
  cDC1 & .02 & \textbf{.08} & .01 & .06 & .02 & .01 & .01 & .04 & .04 & .02 & .03 \\
  cDC2 & .18 & \textbf{.20} & .18 & .18 & .16 & .16 & .19 & .17 & .11 & .12 & .07 \\
  pDC, interferon-producing & .17 & \textbf{.36} & .18 & .15 & .10 & .15 & .12 & .10 & .05 & .06 & .08 \\
  pDC, lineage program & .57 & .59 & .57 & .61 & .57 & .59 & .59 & .59 & .61 & \textbf{.62} & .59 \\
  \midrule
  \multicolumn{12}{@{}l}{\textit{Cancer} (9)} \\
  CD44+ KIT+ cancer stem cell & .15 & .25 & .23 & .22 & .19 & \textbf{.26} & .21 & .14 & .11 & .07 & .06 \\
  Epithelial cancer stem cell & .58 & .57 & .54 & .57 & .55 & .55 & .55 & .53 & .56 & \textbf{.59} & .58 \\
  Hormone-receptor+ carcinoma cell & .34 & .28 & .32 & .30 & .27 & .31 & \textbf{.37} & .35 & .22 & .21 & .21 \\
  Neural stem-like cancer cell & .54 & \textbf{.57} & .53 & .53 & .54 & .51 & .54 & .51 & .19 & .26 & .14 \\
  Neural-crest-like tumour cell & .60 & .60 & .56 & .57 & \textbf{.64} & .58 & .57 & .61 & .32 & .41 & .54 \\
  Ovarian carcinoma cell & .66 & .67 & .62 & \textbf{.81} & .62 & .61 & .55 & .57 & .55 & .70 & .56 \\
  Pluripotency-like cancer stem cell & .53 & .53 & .61 & \textbf{.63} & .58 & .62 & .61 & .59 & .32 & .35 & .24 \\
  Proliferating glial-like tumour cell & \textbf{.70} & .62 & .64 & .64 & .66 & .63 & .59 & .62 & .66 & .59 & .39 \\
  SOX9+ CD44+ stem-like cancer cell & .57 & .55 & .54 & .57 & \textbf{.58} & .55 & .53 & .47 & .23 & .21 & .10 \\
  \midrule
  \multicolumn{12}{@{}l}{\textit{Other} (12)} \\
  Adipocyte & .53 & .50 & .52 & .50 & .53 & .51 & .49 & .53 & \textbf{.56} & .49 & .53 \\
  Basal cell & \textbf{.49} & .42 & .48 & .40 & .48 & .44 & .37 & .20 & .11 & .09 & .08 \\
  Capillary pericyte & .43 & .49 & \textbf{.49} & .48 & .42 & .49 & .47 & .37 & .15 & .15 & .13 \\
  Endothelial cell & .26 & .25 & .26 & \textbf{.26} & .25 & .24 & .25 & .24 & .24 & .23 & .22 \\
  FOXI1+ ionocyte & .26 & .27 & .27 & \textbf{.28} & .25 & .25 & .25 & .24 & .20 & .14 & .16 \\
  Lymphatic endothelial cell & .13 & .16 & .10 & .10 & .08 & \textbf{.25} & .10 & .11 & .13 & .11 & .11 \\
  MYH11+ perivascular cell & .42 & .39 & \textbf{.42} & .40 & .41 & .35 & .29 & .29 & .16 & .16 & .16 \\
  Neurovascular unit (endothelium + pericyte) & .17 & \textbf{.18} & .16 & .14 & .08 & .09 & .07 & .06 & .05 & .02 & .03 \\
  Pancreatic islet endocrine cell & \textbf{.71} & .66 & .62 & .54 & .67 & .51 & .56 & .30 & .12 & .08 & .06 \\
  RGS5+ mural cell & .44 & .47 & .50 & .47 & \textbf{.52} & .49 & .45 & .48 & .39 & .43 & .42 \\
  Schwann cell & \textbf{.51} & .43 & .37 & .35 & .34 & .39 & .31 & .27 & .28 & .21 & .23 \\
  Smooth muscle cell & .57 & .57 & .53 & .58 & \textbf{.58} & .54 & .54 & .51 & .49 & .51 & .38 \\
  \bottomrule
  \end{tabular}}
\end{table}

\setcounter{savedfigure}{\value{figure}}\setcounter{figure}{17}
\begin{figure}[!htbp]
  \centering
  \includegraphics[width=\linewidth,height=0.86\textheight,keepaspectratio]{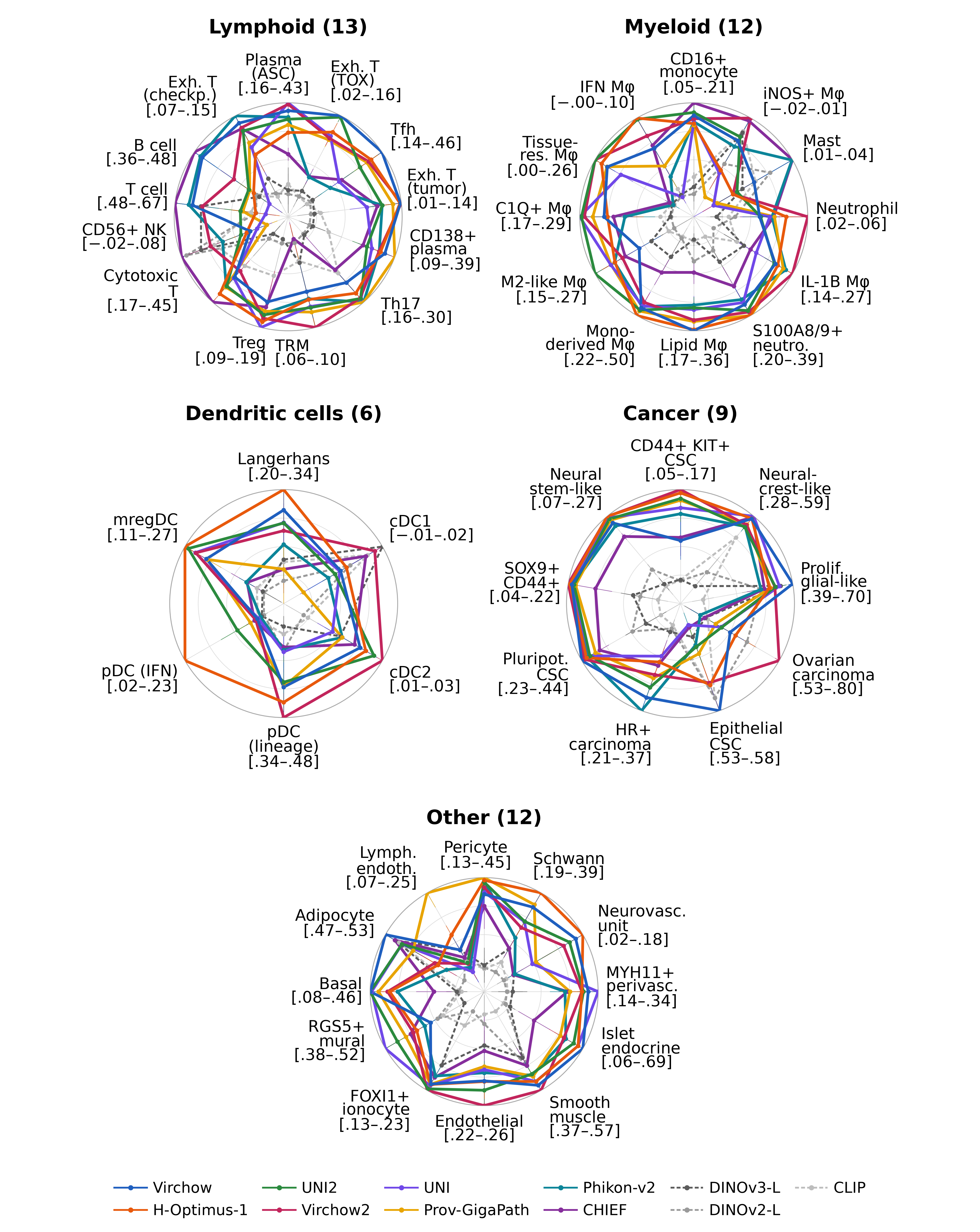}
  \caption{\textbf{All 52 identities, quality $Q$.}
  Quality $Q$ of the matched group per identity; values in
  \cref{tab:supp_identity_q}.}
  \label{fig:supp_radar_q}
\end{figure}
\setcounter{figure}{\value{savedfigure}}

\setcounter{savedtable}{\value{table}}\setcounter{table}{27}
\begin{table}[!htbp]
  \centering
  \caption{\textbf{All 52 identities per encoder, quality $Q$.}
  Quality $Q$ of the matched group per identity, mean over five seeds; bold: best encoder.}
  \label{tab:supp_identity_q}
  \scriptsize
  \setlength{\tabcolsep}{3.5pt}
  \resizebox{\linewidth}{!}{%
  \input{tables/supp_identity_q.tex}}
\end{table}
\setcounter{table}{\value{savedtable}}

\FloatBarrier
\section{Natural-image vs.\ pathology gap breakdown}
\label{sec:supp_lineage}

\Cref{tab:supp_lineage,tab:supp_identity_gap} detail the gaps between pathology and natural-image models per lineage and
per identity, behind \cref{sec:breakdown}.

\begin{table}[!htbp]
\begin{minipage}[t]{0.5\linewidth}
  \centering
  \setcounter{table}{20}
  \caption{\textbf{Gap between the two families per identity.}
  Mean $\rho$ of the matched group over pathology (Path.) and natural-image (Nat.) models, their gap, and the encoder
  with the best $Q$. Rows grouped as in \cref{fig:supp_radar_all}.}
  \label{tab:supp_identity_gap}
  \scriptsize
  \setlength{\tabcolsep}{0.8pt}
\begin{tabular}[t]{@{}llcccl@{}}
    \toprule
    Cell identity & Lineage & Path. & Nat. & Gap & Best ($Q$) \\
    \midrule
    \multicolumn{6}{@{}l}{\textit{Lymphoid} (13)} \\
    CD138+ BCMA+ plasma cell & immune & 0.47 & 0.18 & 0.30 & Prov-GigaPath (0.39) \\
    Antibody-secreting plasma cell & immune & 0.48 & 0.21 & 0.27 & UNI (0.43) \\
    T follicular helper cell & immune & 0.45 & 0.25 & 0.19 & Virchow (0.46) \\
    Cytotoxic T cell & immune & 0.43 & 0.25 & 0.18 & CHIEF (0.45) \\
    Tumor-reactive exhausted T cell & immune & 0.19 & 0.05 & 0.14 & Virchow2 (0.14) \\
    TOX-driven exhausted T cell & immune & 0.21 & 0.07 & 0.14 & Virchow (0.16) \\
    T helper17 cell & immune & 0.48 & 0.35 & 0.14 & Prov-GigaPath (0.30) \\
    Checkpoint-high exhausted T cell & immune & 0.18 & 0.11 & 0.07 & Phikon-v2 (0.15) \\
    Regulatory T cell & immune & 0.25 & 0.19 & 0.06 & UNI (0.19) \\
    T cell & immune & 0.61 & 0.57 & 0.04 & CHIEF (0.67) \\
    Tissue-resident memory T cell & immune & 0.10 & 0.08 & 0.02 & Virchow2 (0.10) \\
    B cell & immune & 0.52 & 0.52 & -0.00 & CHIEF (0.48) \\
    CD56+ NK cell & immune & 0.04 & 0.07 & -0.04 & CHIEF (0.08) \\
    \midrule
    \multicolumn{6}{@{}l}{\textit{Myeloid} (12)} \\
    Tissue-resident macrophage & immune & 0.35 & 0.02 & 0.34 & UNI2 (0.26) \\
    Monocyte-derived macrophage & immune & 0.56 & 0.28 & 0.28 & H-Optimus-1 (0.50) \\
    Lipid-associated macrophage & immune & 0.45 & 0.25 & 0.20 & Virchow (0.36) \\
    CD16+ monocyte & immune & 0.28 & 0.09 & 0.18 & CHIEF (0.21) \\
    S100A8/A9+ inflammatory neutrophil & immune & 0.44 & 0.28 & 0.16 & Prov-GigaPath (0.39) \\
    Interferon-activated macrophage & immune & 0.17 & 0.02 & 0.15 & H-Optimus-1 (0.10) \\
    Perivascular C1Q+ macrophage & immune & 0.26 & 0.18 & 0.09 & UNI (0.29) \\
    Mature neutrophil & immune & 0.13 & 0.04 & 0.09 & Virchow2 (0.06) \\
    M2-like macrophage & immune & 0.26 & 0.18 & 0.08 & UNI2 (0.27) \\
    Inflammatory IL-1B macrophage & immune & 0.34 & 0.27 & 0.07 & Virchow2 (0.27) \\
    iNOS+ classically activated macrophage & immune & 0.02 & 0.01 & 0.01 & Virchow2 (0.01) \\
    Mast cell & immune & 0.03 & 0.03 & -0.00 & CHIEF (0.04) \\
    \midrule
    \multicolumn{6}{@{}l}{\textit{Dendritic cells} (6)} \\
    pDC, interferon-producing & immune & 0.17 & 0.06 & 0.10 & H-Optimus-1 (0.23) \\
    cDC2 & immune & 0.18 & 0.10 & 0.08 & Virchow2 (0.03) \\
    Langerhans cell & immune & 0.28 & 0.22 & 0.05 & H-Optimus-1 (0.34) \\
    cDC1 & immune & 0.03 & 0.03 & 0.00 & DINOv3-L (0.02) \\
    pDC, lineage program & immune & 0.58 & 0.60 & -0.02 & Virchow2 (0.48) \\
    Mature regulatory DC (mregDC) & immune & 0.35 & 0.39 & -0.04 & H-Optimus-1 (0.27) \\
    \bottomrule
  \end{tabular}
\end{minipage}\hfill
\begin{minipage}[t]{0.495\linewidth}
  \centering
  \scriptsize
  \setlength{\tabcolsep}{0.8pt}
  \textit{\cref{tab:supp_identity_gap} (continued)}\\[2pt]
\begin{tabular}[t]{@{}llcccl@{}}
    \toprule
    Cell identity & Lineage & Path. & Nat. & Gap & Best ($Q$) \\
    \midrule
    \multicolumn{6}{@{}l}{\textit{Cancer} (9)} \\
    SOX9+ CD44+ stem-like cancer cell & cancer & 0.54 & 0.18 & 0.36 & Virchow2 (0.22) \\
    Neural stem-like cancer cell & cancer & 0.53 & 0.20 & 0.34 & H-Optimus-1 (0.27) \\
    Pluripotency-like cancer stem cell & cancer & 0.59 & 0.30 & 0.28 & Virchow (0.44) \\
    Neural-crest-like tumour cell & cancer & 0.59 & 0.42 & 0.17 & UNI (0.59) \\
    CD44+ KIT+ cancer stem cell & cancer & 0.21 & 0.08 & 0.13 & Virchow2 (0.17) \\
    Hormone-receptor+ carcinoma cell & cancer & 0.32 & 0.22 & 0.10 & Phikon-v2 (0.37) \\
    Proliferating glial-like tumour cell & cancer & 0.64 & 0.55 & 0.09 & Virchow (0.70) \\
    Ovarian carcinoma cell & cancer & 0.64 & 0.60 & 0.04 & Virchow2 (0.80) \\
    Epithelial cancer stem cell & cancer & 0.55 & 0.58 & -0.02 & Virchow (0.58) \\
    \midrule
    \multicolumn{6}{@{}l}{\textit{Other} (12)} \\
    Pancreatic islet endocrine cell & parenchyma & 0.57 & 0.09 & 0.49 & Virchow (0.69) \\
    Basal cell & epithelial & 0.41 & 0.09 & 0.32 & UNI2 (0.46) \\
    Capillary pericyte & stromal & 0.46 & 0.14 & 0.31 & Prov-GigaPath (0.45) \\
    MYH11+ perivascular cell & stromal & 0.37 & 0.16 & 0.21 & UNI (0.34) \\
    Schwann cell & parenchyma & 0.37 & 0.24 & 0.14 & H-Optimus-1 (0.39) \\
    FOXI1+ ionocyte & epithelial & 0.26 & 0.17 & 0.09 & Virchow2 (0.23) \\
    \begin{tabular}[t]{@{}l@{}}Neurovascular unit\\(endothelium + pericyte)\end{tabular} & other & 0.12 & 0.03 & 0.09 & H-Optimus-1 (0.18) \\
    Smooth muscle cell & stromal & 0.55 & 0.46 & 0.09 & Virchow2 (0.57) \\
    RGS5+ mural cell & stromal & 0.48 & 0.41 & 0.07 & UNI (0.52) \\
    Endothelial cell & endothelial & 0.25 & 0.23 & 0.02 & Virchow2 (0.26) \\
    Lymphatic endothelial cell & endothelial & 0.13 & 0.12 & 0.01 & Prov-GigaPath (0.25) \\
    Adipocyte & parenchyma & 0.51 & 0.52 & -0.01 & Virchow (0.53) \\
    \bottomrule
  \end{tabular}

  \vspace{10pt}
  \setcounter{table}{19}
  \caption{\textbf{Quality per lineage.}
  Mean quality $Q$ per lineage (number of identities in parentheses).}
  \label{tab:supp_lineage}
  \scriptsize
  \setlength{\tabcolsep}{0.8pt}
    \begin{tabular}{@{}lccccccc@{}}
    \toprule
    & Imm. & Canc. & Strom. & Par. & Endo. & Epi. & Oth. \\
    & (31) & (9) & (4) & (3) & (2) & (2) & (1) \\
    \midrule
    Virchow & .23 & .43 & .43 & .53 & .19 & .34 & .17 \\
    H-Optimus-1 & .23 & .42 & .44 & .51 & .20 & .30 & .18 \\
    UNI2 & .23 & .41 & .44 & .49 & .17 & .34 & .16 \\
    Virchow2 & .23 & .44 & .44 & .45 & .18 & .31 & .14 \\
    UNI & .21 & .41 & .45 & .50 & .16 & .34 & .08 \\
    Prov-GigaPath & .22 & .41 & .44 & .46 & .25 & .32 & .09 \\
    Phikon-v2 & .21 & .40 & .42 & .44 & .16 & .28 & .04 \\
    CHIEF & .20 & .38 & .40 & .36 & .17 & .20 & .04 \\
    \midrule
    DINOv3-L & .14 & .30 & .29 & .29 & .18 & .15 & .04 \\
    DINOv2-L & .13 & .33 & .30 & .25 & .16 & .11 & .02 \\
    CLIP & .14 & .29 & .26 & .27 & .16 & .11 & .03 \\
    \bottomrule
  \end{tabular}
\end{minipage}
\end{table}
\setcounter{table}{21}
\FloatBarrier
\section{BioScore on raw embedding}
\label{sec:supp_raw}

To assert the utility of the SAE step, we compute BioScore on the raw embeddings, without SAE or grouping.
With all raw dimensions, BioScore follows the number of units, which grows with the embedding size (Spearman correlation 0.88 between BioScore and embedding size, from 768 to 2560). We therefore give every encoder the same number of units, as the probe of \cref{sec:downstream} does: each embedding is projected on the top 768 principal components (the smallest embedding size) of the 20$\times$ embeddings the SAE is trained on, and each component, scaled to unit variance, is split by sign into two non-negative units, $\max(0, z)$ and $\max(0, -z)$.
These 1,536 units per encoder enter BioScore as the feature groups do (20$\times$/20$\times$, $\lambda=3$).
Without seeds, the paired tests use one value per identity and encoder.

The raw embeddings lead to the same conclusions as the SAE (\cref{fig:supp_raw,tab:supp_raw_identity}): every pathology model scores significantly above every natural-image model, the ranking of the encoders stays close to the one with the SAE (Kendall $\tau=0.75$), and the paired tests separate as many pairs of encoders (31 of 55, against 32).
Most notably, the gap between pathology and natural-image models is smaller on the raw embeddings (0.100, against 0.126 with the SAE), and the ranking transitions more smoothly from the top encoders to the worst ones.
We argue that the SAE variant, on top of a more confident separation between models, brings value with its interpretability: the unit matched to an identity is a group of sparse features whose activation map can be plotted and reviewed (\cref{fig:maps_main}), whereas a principal component mixes the whole embedding.

\setcounter{savedfigure}{\value{figure}}\setcounter{figure}{13}
\begin{figure}[!htbp]
  \centering
  \resizebox{0.6\linewidth}{!}{%
  \includegraphics[width=\linewidth]{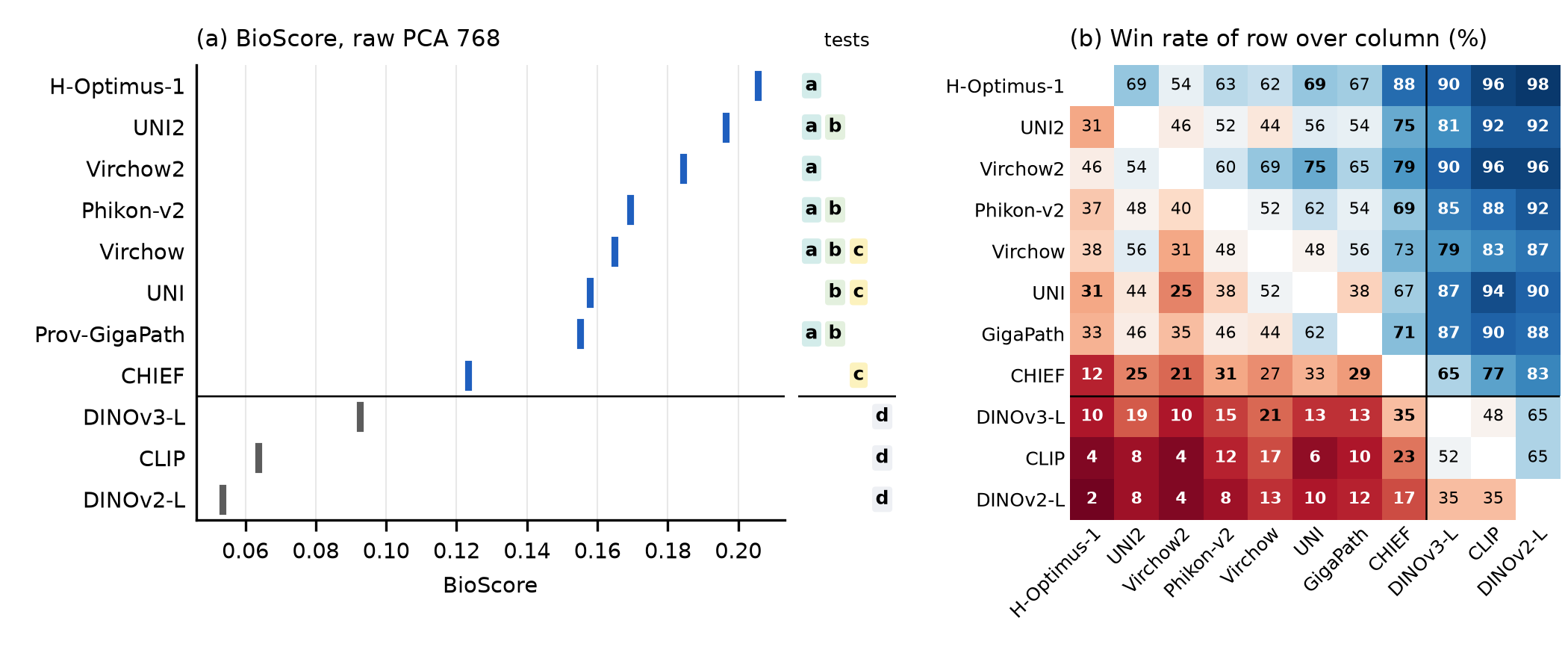}}
  \caption{\textbf{Raw embedding dimensions.}
  (a) BioScore on the raw embeddings (top 768 principal components, split by sign), with the compact letter display of
  the paired tests.
  (b) Win rate of the row encoder over the column encoder on the raw embeddings (share of the 52 identities with the
  higher BioScore); bold: Wilcoxon signed-rank, Holm-corrected $p<0.05$.}
  \label{fig:supp_raw}
\end{figure}
\setcounter{figure}{\value{savedfigure}}

\setcounter{savedtable}{\value{table}}\setcounter{table}{24}
\begin{table}[!htbp]
  \centering
  \caption{\textbf{BioScore per identity on the raw embeddings.}
  Quality $Q$ of the best unit (top 768 principal components, split by sign) for each identity and encoder
  (20$\times$/20$\times$, $\lambda=3$), and BioScore. Rows grouped as in \cref{fig:supp_radar_all}.}
  \label{tab:supp_raw_identity}
  \scriptsize
  \setlength{\tabcolsep}{3pt}
  \resizebox{0.75\linewidth}{!}{%
  \input{tables/supp_raw_identity.tex}}
\end{table}
\setcounter{table}{\value{savedtable}}
\FloatBarrier

%% file: tables/supp_threshold_sets.tex
\begin{tabular}{@{}lcccc@{\hspace{8pt}}ccccccc@{}}
\toprule
& & & & & \multicolumn{7}{c}{Sets per majority lineage} \\
\cmidrule(l){6-12}
Threshold & Sets & Signatures & Singletons & Largest & Immune & Cancer & Stromal & Endothelial & Epithelial & Parenchyma & Other \\
\midrule
0.2 & 24 & 133 & 13 & 59 & 11 & 3 & 3 & 1 & 3 & 3 & 0 \\
0.3 & 30 & 133 & 16 & 54 & 13 & 6 & 3 & 1 & 3 & 3 & 1 \\
0.4 & 35 & 133 & 19 & 44 & 17 & 6 & 3 & 2 & 3 & 3 & 1 \\
0.5 & 44 & 135 & 24 & 24 & 23 & 7 & 5 & 2 & 3 & 3 & 1 \\
0.6 (ours) & 55 & 139 & 33 & 19 & 32 & 8 & 6 & 2 & 3 & 3 & 1 \\
0.7 & 68 & 147 & 46 & 12 & 43 & 10 & 6 & 2 & 3 & 3 & 1 \\
0.8 & 93 & 157 & 71 & 11 & 64 & 10 & 7 & 3 & 4 & 3 & 2 \\
\bottomrule
\end{tabular}

%% file: tables/supp_probe_identity.tex
\begin{tabular}{@{}lcccccccc@{\hspace{6pt}}ccc@{\hspace{6pt}}c@{}}
\toprule
Identity & Virchow & H-Opt.-1 & UNI2 & Virchow2 & UNI & GigaPath & Phikon-v2 & CHIEF & DINOv3 & DINOv2 & CLIP & \begin{tabular}[b]{@{}c@{}}Pathology model\\ranking agreement\end{tabular} \\
\midrule
\multicolumn{13}{@{}l}{\textit{Lymphoid} (10)} \\
Antibody-secreting plasma cell & .70 & .60 & .58 & .59 & .64 & .49 & .52 & .52 & .23 & .27 & .24 & .50 \\
\textbf{B cell} & \textbf{.87} & \textbf{.88} & \textbf{.90} & \textbf{.88} & \textbf{.84} & \textbf{.88} & \textbf{.86} & \textbf{.81} & \textbf{.77} & \textbf{.75} & \textbf{.77} & \textbf{$\boldsymbol{-}$.31} \\
CD138+ BCMA+ plasma cell & .59 & .46 & .49 & .39 & .49 & .54 & .54 & .35 & .03 & .06 & .11 & .62 \\
Checkpoint-high exhausted T cell & .61 & .56 & .60 & .57 & .59 & .57 & .58 & .54 & .46 & .45 & .46 & .26 \\
Cytotoxic T cell & .59 & .61 & .58 & .60 & .60 & .60 & .56 & .54 & .52 & .52 & .51 & $-$.07 \\
Regulatory T cell & .77 & .75 & .76 & .76 & .73 & .75 & .73 & .68 & .60 & .60 & .62 & $-$.12 \\
\textbf{T cell} & \textbf{.90} & \textbf{.89} & \textbf{.90} & \textbf{.90} & \textbf{.87} & \textbf{.88} & \textbf{.88} & \textbf{.83} & \textbf{.76} & \textbf{.74} & \textbf{.76} & \textbf{$\boldsymbol{-}$.12} \\
T follicular helper cell & .66 & .57 & .63 & .66 & .56 & .60 & .64 & .57 & .39 & .38 & .38 & .31 \\
TOX-driven exhausted T cell & .45 & .44 & .51 & .47 & .41 & .44 & .53 & .37 & .26 & .28 & .22 & .02 \\
Tumour-reactive exhausted T cell & .66 & .63 & .63 & .65 & .57 & .63 & .62 & .50 & .41 & .39 & .36 & .93 \\
\midrule
\multicolumn{13}{@{}l}{\textit{Myeloid} (6)} \\
CD16+ monocyte & .29 & .27 & .30 & .29 & .19 & .21 & .22 & .21 & .19 & .18 & .11 & .50 \\
Inflammatory IL-1B macrophage & .40 & .38 & .29 & .31 & .24 & .33 & .28 & .27 & .18 & .13 & .15 & .43 \\
Interferon-activated macrophage & .20 & .14 & .15 & .17 & .12 & .18 & .25 & .15 & .07 & .08 & .03 & $-$.24 \\
M2-like macrophage & .45 & .47 & .39 & .47 & .40 & .41 & .34 & .28 & .34 & .31 & .34 & $-$.24 \\
Mast cell & .17 & .26 & .15 & .17 & .17 & .16 & .11 & .14 & .15 & .16 & .17 & $-$.50 \\
Tissue-resident macrophage & .26 & .21 & .13 & .21 & .19 & .15 & .24 & .18 & .09 & .11 & .10 & $-$.38 \\
\midrule
\multicolumn{13}{@{}l}{\textit{Dendritic cells} (4)} \\
Mature regulatory DC (mregDC) & .73 & .75 & .74 & .74 & .67 & .70 & .70 & .67 & .63 & .65 & .64 & .60 \\
cDC2 & .35 & .33 & .32 & .32 & .26 & .30 & .30 & .31 & .24 & .22 & .13 & .74 \\
pDC, interferon-producing & .53 & .55 & .53 & .54 & .52 & .53 & .53 & .46 & .41 & .42 & .39 & .24 \\
pDC, lineage program & .76 & .68 & .73 & .73 & .70 & .74 & .75 & .77 & .67 & .63 & .66 & $-$.52 \\
\midrule
\multicolumn{13}{@{}l}{\textit{Other} (7)} \\
Adipocyte & .68 & .72 & .72 & .68 & .59 & .67 & .64 & .60 & .61 & .63 & .64 & $-$.07 \\
Basal cell & .71 & .71 & .68 & .71 & .59 & .55 & .53 & .28 & .29 & .28 & .18 & .55 \\
Endothelial cell & .42 & .49 & .46 & .52 & .50 & .43 & .43 & .45 & .37 & .36 & .33 & .45 \\
FOXI1+ ionocyte & .23 & .12 & .10 & .20 & .14 & .11 & .14 & .05 & $-$.09 & $-$.06 & $-$.07 & .05 \\
MYH11+ perivascular cell & .62 & .59 & .65 & .64 & .63 & .62 & .59 & .60 & .37 & .30 & .27 & .62 \\
\textbf{Pancreatic islet endocrine cell} & \textbf{.29} & \textbf{.22} & \textbf{.34} & \textbf{.09} & \textbf{.33} & \textbf{.10} & \textbf{.24} & \textbf{.16} & \textbf{.02} & \textbf{.13} & \textbf{.08} & \textbf{.69} \\
Smooth muscle cell & .61 & .57 & .52 & .57 & .54 & .45 & .41 & .47 & .34 & .28 & .42 & .74 \\
\bottomrule
\end{tabular}

%% file: tables/supp_identity_recipe_sd.tex
\begin{tabular}{@{}lccccccccccc@{\hspace{6pt}}cc@{\hspace{6pt}}cc@{}}
\toprule
Cell identity & Virchow & H-Opt.-1 & UNI2 & Virchow2 & UNI & GigaPath & Phikon-v2 & CHIEF & DINOv3 & DINOv2 & CLIP & Mean & Max & $\tau$ & \begin{tabular}[b]{@{}c@{}}BioScore\\range\end{tabular} \\
\midrule
\multicolumn{16}{@{}l}{\textit{Lymphoid} (13)} \\
T cell & .017 & .034 & .054 & .020 & .068 & .055 & .045 & .014 & .039 & .025 & .013 & .035 & .068 & .61 & 0.473--0.665 \\
Cytotoxic T cell & .007 & .016 & .028 & .014 & .032 & .012 & .026 & .027 & .074 & .038 & .071 & .031 & .074 & .48 & 0.203--0.427 \\
Antibody-secreting plasma cell & .035 & .025 & .027 & .050 & .041 & .007 & .012 & .012 & .046 & .036 & .037 & .030 & .050 & .73 & 0.175--0.421 \\
B cell & .030 & .025 & .046 & .032 & .027 & .055 & .030 & .026 & .031 & .014 & .009 & .030 & .055 & .59 & 0.363--0.477 \\
CD138+ BCMA+ plasma cell & .017 & .022 & .023 & .030 & .029 & .013 & .013 & .025 & .041 & .030 & .061 & .028 & .061 & .75 & 0.107--0.391 \\
T helper17 cell & .019 & .015 & .018 & .009 & .027 & .015 & .014 & .021 & .030 & .062 & .059 & .026 & .062 & .56 & 0.149--0.296 \\
T follicular helper cell & .013 & .030 & .034 & .038 & .029 & .036 & .030 & .018 & .015 & .028 & .016 & .026 & .038 & .80 & 0.140--0.479 \\
Checkpoint-high exhausted T cell & .003 & .013 & .017 & .004 & .025 & .015 & .026 & .015 & .025 & .031 & .039 & .019 & .039 & .56 & 0.076--0.148 \\
CD56+ NK cell & .001 & .010 & .007 & .011 & .004 & .016 & .042 & .004 & .013 & .010 & .036 & .014 & .042 & .38 & 0.040--0.087 \\
Tumour-reactive exhausted T cell & .003 & .009 & .024 & .004 & .013 & .008 & .015 & .027 & .005 & .014 & .009 & .012 & .027 & .68 & 0.026--0.139 \\
Regulatory T cell & .006 & .007 & .006 & .007 & .011 & .010 & .014 & .009 & .012 & .024 & .021 & .012 & .024 & .73 & 0.088--0.194 \\
TOX-driven exhausted T cell & .006 & .005 & .006 & .009 & .008 & .008 & .022 & .011 & .007 & .012 & .004 & .009 & .022 & .82 & 0.027--0.158 \\
Tissue-resident memory T cell & .006 & .006 & .006 & .005 & .012 & .005 & .002 & .010 & .008 & .012 & .006 & .007 & .012 & .69 & 0.064--0.108 \\
\midrule
\multicolumn{16}{@{}l}{\textit{Myeloid} (12)} \\
Inflammatory IL-1B macrophage & .012 & .008 & .012 & .016 & .008 & .018 & .033 & .014 & .028 & .044 & .052 & .022 & .052 & .61 & 0.126--0.275 \\
Interferon-activated macrophage & .023 & .025 & .040 & .028 & .007 & .036 & .029 & .014 & .009 & .009 & .006 & .021 & .040 & .62 & 0.017--0.108 \\
Monocyte-derived macrophage & .010 & .015 & .014 & .016 & .015 & .017 & .010 & .016 & .025 & .037 & .040 & .020 & .040 & .67 & 0.216--0.489 \\
Perivascular C1Q+ macrophage & .014 & .007 & .021 & .013 & .015 & .015 & .019 & .013 & .039 & .030 & .024 & .019 & .039 & .63 & 0.142--0.278 \\
S100A8/A9+ inflammatory neutrophil & .010 & .013 & .017 & .017 & .015 & .014 & .019 & .025 & .037 & .028 & .011 & .019 & .037 & .79 & 0.197--0.413 \\
M2-like macrophage & .020 & .018 & .005 & .009 & .016 & .013 & .029 & .014 & .010 & .017 & .013 & .015 & .029 & .66 & 0.157--0.258 \\
Lipid-associated macrophage & .020 & .011 & .018 & .015 & .015 & .003 & .018 & .007 & .014 & .009 & .029 & .015 & .029 & .72 & 0.176--0.359 \\
CD16+ monocyte & .010 & .005 & .013 & .006 & .011 & .010 & .007 & .023 & .029 & .013 & .025 & .014 & .029 & .50 & 0.071--0.196 \\
Tissue-resident macrophage & .025 & .005 & .006 & .010 & .010 & .006 & .033 & .010 & .010 & .004 & .006 & .011 & .033 & .85 & 0.018--0.255 \\
Mast cell & .008 & .003 & .006 & .013 & .003 & .009 & .014 & .012 & .005 & .007 & .011 & .008 & .014 & .51 & 0.032--0.065 \\
Mature neutrophil & .007 & .007 & .004 & .008 & .002 & .010 & .009 & .006 & .006 & .005 & .006 & .007 & .010 & .55 & 0.026--0.062 \\
iNOS+ classically activated macrophage & .004 & .004 & .006 & .003 & .005 & .005 & .005 & .009 & .010 & .010 & .006 & .006 & .010 & .65 & 0.014--0.036 \\
\midrule
\multicolumn{16}{@{}l}{\textit{Dendritic cells} (6)} \\
Langerhans cell & .014 & .010 & .034 & .022 & .019 & .013 & .031 & .016 & .015 & .020 & .022 & .020 & .034 & .59 & 0.214--0.367 \\
Mature regulatory DC (mregDC) & .008 & .016 & .023 & .023 & .022 & .017 & .028 & .016 & .019 & .013 & .016 & .018 & .028 & .74 & 0.110--0.267 \\
pDC, lineage program & .017 & .027 & .011 & .016 & .022 & .011 & .017 & .007 & .014 & .008 & .003 & .014 & .027 & .79 & 0.339--0.471 \\
pDC, interferon-producing & .014 & .009 & .011 & .007 & .011 & .013 & .007 & .005 & .007 & .005 & .004 & .009 & .014 & .80 & 0.030--0.218 \\
cDC1 & .008 & .006 & .006 & .005 & .007 & .009 & .004 & .005 & .007 & .003 & .005 & .006 & .009 & .41 & 0.025--0.041 \\
cDC2 & .002 & .004 & .003 & .004 & .005 & .007 & .004 & .004 & .005 & .009 & .003 & .005 & .009 & .69 & 0.017--0.035 \\
\midrule
\multicolumn{16}{@{}l}{\textit{Cancer} (9)} \\
Neural-crest-like tumour cell & .013 & .012 & .029 & .015 & .007 & .033 & .027 & .022 & .164 & .088 & .023 & .039 & .164 & .63 & 0.276--0.593 \\
Ovarian carcinoma cell & .026 & .042 & .034 & .031 & .064 & .033 & .039 & .035 & .046 & .018 & .054 & .038 & .064 & .67 & 0.517--0.792 \\
Epithelial cancer stem cell & .005 & .012 & .023 & .019 & .023 & .021 & .025 & .032 & .033 & .114 & .064 & .034 & .114 & .17 & 0.520--0.592 \\
Proliferating glial-like tumour cell & .026 & .038 & .028 & .015 & .023 & .034 & .032 & .040 & .018 & .065 & .036 & .032 & .065 & .62 & 0.415--0.701 \\
Pluripotency-like cancer stem cell & .010 & .029 & .014 & .018 & .006 & .023 & .013 & .029 & .029 & .017 & .017 & .019 & .029 & .76 & 0.204--0.446 \\
Hormone-receptor+ carcinoma cell & .026 & .013 & .032 & .021 & .010 & .018 & .038 & .019 & .009 & .006 & .010 & .018 & .038 & .77 & 0.212--0.369 \\
CD44+ KIT+ cancer stem cell & .009 & .010 & .010 & .017 & .017 & .007 & .017 & .008 & .026 & .021 & .030 & .016 & .030 & .75 & 0.046--0.171 \\
Neural stem-like cancer cell & .004 & .005 & .011 & .010 & .009 & .003 & .007 & .032 & .009 & .019 & .019 & .012 & .032 & .82 & 0.065--0.276 \\
SOX9+ CD44+ stem-like cancer cell & .004 & .003 & .004 & .002 & .003 & .009 & .002 & .011 & .018 & .014 & .014 & .008 & .018 & .73 & 0.041--0.218 \\
\midrule
\multicolumn{16}{@{}l}{\textit{Other} (12)} \\
Pancreatic islet endocrine cell & .016 & .020 & .021 & .022 & .035 & .046 & .041 & .125 & .030 & .033 & .012 & .036 & .125 & .88 & 0.073--0.703 \\
Smooth muscle cell & .019 & .019 & .033 & .032 & .016 & .029 & .012 & .013 & .049 & .074 & .032 & .030 & .074 & .67 & 0.387--0.568 \\
Capillary pericyte & .011 & .018 & .017 & .016 & .017 & .029 & .042 & .064 & .022 & .018 & .046 & .027 & .064 & .66 & 0.105--0.432 \\
RGS5+ mural cell & .016 & .029 & .022 & .022 & .013 & .005 & .025 & .007 & .053 & .067 & .032 & .026 & .067 & .64 & 0.356--0.508 \\
Basal cell & .038 & .047 & .037 & .016 & .019 & .031 & .020 & .036 & .024 & .008 & .002 & .025 & .047 & .74 & 0.079--0.439 \\
Lymphatic endothelial cell & .021 & .037 & .020 & .010 & .028 & .029 & .026 & .014 & .022 & .019 & .029 & .023 & .037 & .69 & 0.110--0.251 \\
Schwann cell & .027 & .034 & .012 & .016 & .023 & .026 & .016 & .029 & .025 & .015 & .027 & .023 & .034 & .80 & 0.206--0.398 \\
MYH11+ perivascular cell & .013 & .015 & .019 & .010 & .016 & .015 & .046 & .039 & .011 & .025 & .023 & .021 & .046 & .65 & 0.124--0.326 \\
Neurovascular unit (endothelium + pericyte) & .040 & .013 & .033 & .047 & .030 & .023 & .020 & .010 & .006 & .005 & .004 & .021 & .047 & .77 & 0.032--0.171 \\
Adipocyte & .024 & .024 & .017 & .016 & .010 & .020 & .030 & .016 & .013 & .009 & .039 & .020 & .039 & .24 & 0.473--0.534 \\
FOXI1+ ionocyte & .009 & .007 & .005 & .003 & .007 & .002 & .011 & .012 & .031 & .024 & .032 & .013 & .032 & .65 & 0.127--0.233 \\
Endothelial cell & .005 & .005 & .005 & .003 & .005 & .007 & .005 & .008 & .008 & .006 & .021 & .007 & .021 & .70 & 0.202--0.265 \\
\midrule
Mean over identities & .014 & .016 & .019 & .016 & .018 & .018 & .021 & .020 & .025 & .024 & .024 & .020 & .043 & .66 & \\
\bottomrule
\end{tabular}

%% file: tables/supp_identity_seed_sd.tex
\begin{tabular}{@{}lccccccccccc@{\hspace{6pt}}cc@{\hspace{6pt}}cc@{}}
\toprule
Cell identity & Virchow & H-Opt.-1 & UNI2 & Virchow2 & UNI & GigaPath & Phikon-v2 & CHIEF & DINOv3 & DINOv2 & CLIP & Mean & Max & $\tau$ & \begin{tabular}[b]{@{}c@{}}BioScore\\range\end{tabular} \\
\midrule
\multicolumn{16}{@{}l}{\textit{Lymphoid} (13)} \\
T helper17 cell & .017 & .040 & .049 & .022 & .067 & .043 & .030 & .066 & .099 & .110 & .011 & .050 & .110 & .18 & 0.160--0.298 \\
T follicular helper cell & .016 & .054 & .079 & .037 & .040 & .052 & .050 & .050 & .037 & .034 & .030 & .044 & .079 & .72 & 0.141--0.458 \\
CD138+ BCMA+ plasma cell & .023 & .031 & .051 & .065 & .026 & .055 & .006 & .017 & .043 & .038 & .047 & .037 & .065 & .60 & 0.088--0.395 \\
T cell & .022 & .050 & .033 & .012 & .021 & .028 & .063 & .015 & .087 & .032 & .016 & .034 & .087 & .51 & 0.479--0.670 \\
Cytotoxic T cell & .021 & .041 & .027 & .042 & .037 & .027 & .034 & .013 & .032 & .071 & .011 & .032 & .071 & .64 & 0.168--0.445 \\
B cell & .058 & .012 & .028 & .021 & .019 & .014 & .027 & .033 & .037 & .016 & .020 & .026 & .058 & .53 & 0.355--0.477 \\
Regulatory T cell & .005 & .019 & .017 & .050 & .026 & .003 & .014 & .024 & .029 & .041 & .029 & .023 & .050 & .37 & 0.094--0.188 \\
Antibody-secreting plasma cell & .005 & .021 & .019 & .017 & .004 & .023 & .023 & .018 & .067 & .030 & .030 & .023 & .067 & .82 & 0.159--0.429 \\
Checkpoint-high exhausted T cell & .012 & .019 & .019 & .021 & .032 & .014 & .014 & .013 & .027 & .035 & .024 & .021 & .035 & .50 & 0.066--0.154 \\
CD56+ NK cell & .016 & .015 & .017 & .029 & .018 & .012 & .052 & .012 & .023 & .018 & .017 & .021 & .052 & .55 & $-$0.019--0.082 \\
Tumour-reactive exhausted T cell & .011 & .023 & .023 & .023 & .023 & .022 & .015 & .015 & .007 & .018 & .011 & .018 & .023 & .58 & 0.015--0.138 \\
Tissue-resident memory T cell & .012 & .033 & .024 & .017 & .014 & .008 & .009 & .007 & .014 & .021 & .010 & .015 & .033 & .33 & 0.060--0.099 \\
TOX-driven exhausted T cell & .005 & .016 & .008 & .013 & .006 & .012 & .007 & .013 & .018 & .019 & .009 & .012 & .019 & .79 & 0.019--0.164 \\
\midrule
\multicolumn{16}{@{}l}{\textit{Myeloid} (12)} \\
S100A8/A9+ inflammatory neutrophil & .039 & .022 & .017 & .050 & .044 & .036 & .018 & .046 & .034 & .028 & .038 & .034 & .050 & .45 & 0.202--0.393 \\
Perivascular C1Q+ macrophage & .036 & .045 & .030 & .016 & .014 & .021 & .037 & .031 & .055 & .065 & .020 & .034 & .065 & .40 & 0.172--0.286 \\
Monocyte-derived macrophage & .030 & .023 & .014 & .033 & .051 & .030 & .016 & .045 & .065 & .007 & .030 & .031 & .065 & .59 & 0.216--0.505 \\
Interferon-activated macrophage & .033 & .038 & .043 & .048 & .007 & .042 & .014 & .047 & .006 & .006 & .005 & .026 & .048 & .53 & $-$0.000--0.096 \\
Inflammatory IL-1B macrophage & .017 & .021 & .017 & .014 & .031 & .027 & .026 & .015 & .033 & .049 & .033 & .026 & .049 & .52 & 0.138--0.270 \\
Lipid-associated macrophage & .022 & .022 & .022 & .022 & .028 & .023 & .017 & .023 & .012 & .032 & .030 & .023 & .032 & .75 & 0.174--0.363 \\
M2-like macrophage & .032 & .019 & .018 & .015 & .022 & .024 & .013 & .028 & .017 & .026 & .027 & .022 & .032 & .45 & 0.152--0.269 \\
Tissue-resident macrophage & .025 & .038 & .012 & .025 & .039 & .016 & .020 & .014 & .012 & .006 & .007 & .019 & .039 & .64 & 0.002--0.259 \\
CD16+ monocyte & .011 & .003 & .027 & .014 & .014 & .003 & .002 & .016 & .024 & .026 & .035 & .016 & .035 & .55 & 0.054--0.210 \\
Mast cell & .014 & .007 & .010 & .012 & .008 & .017 & .021 & .019 & .011 & .021 & .010 & .014 & .021 & .12 & 0.010--0.041 \\
Mature neutrophil & .008 & .003 & .004 & .022 & .006 & .013 & .009 & .015 & .003 & .019 & .010 & .010 & .022 & .30 & 0.019--0.059 \\
iNOS+ classically activated macrophage & .008 & .007 & .010 & .010 & .006 & .011 & .010 & .006 & .011 & .007 & .011 & .009 & .011 & .33 & $-$0.015--0.013 \\
\midrule
\multicolumn{16}{@{}l}{\textit{Dendritic cells} (6)} \\
Langerhans cell & .052 & .059 & .016 & .023 & .029 & .052 & .050 & .036 & .039 & .011 & .034 & .037 & .059 & .36 & 0.195--0.336 \\
Mature regulatory DC (mregDC) & .007 & .030 & .023 & .024 & .021 & .027 & .014 & .031 & .037 & .024 & .023 & .024 & .037 & .65 & 0.114--0.269 \\
pDC, lineage program & .038 & .014 & .040 & .019 & .031 & .027 & .027 & .016 & .016 & .006 & .020 & .023 & .040 & .53 & 0.342--0.484 \\
pDC, interferon-producing & .021 & .008 & .041 & .024 & .013 & .033 & .011 & .018 & .009 & .009 & .012 & .018 & .041 & .40 & 0.025--0.226 \\
cDC1 & .011 & .023 & .004 & .018 & .014 & .017 & .001 & .007 & .013 & .006 & .009 & .011 & .023 & .17 & $-$0.007--0.021 \\
cDC2 & .005 & .003 & .001 & .002 & .005 & .004 & .005 & .005 & .004 & .012 & .007 & .005 & .012 & .53 & 0.008--0.033 \\
\midrule
\multicolumn{16}{@{}l}{\textit{Cancer} (9)} \\
Ovarian carcinoma cell & .049 & .069 & .038 & .014 & .020 & .028 & .031 & .051 & .072 & .032 & .073 & .043 & .073 & .57 & 0.528--0.803 \\
Proliferating glial-like tumour cell & .015 & .025 & .039 & .026 & .016 & .024 & .031 & .060 & .017 & .068 & .037 & .033 & .068 & .45 & 0.388--0.705 \\
Neural-crest-like tumour cell & .009 & .015 & .012 & .012 & .019 & .015 & .021 & .011 & .122 & .066 & .014 & .029 & .122 & .70 & 0.276--0.592 \\
CD44+ KIT+ cancer stem cell & .026 & .016 & .032 & .005 & .023 & .004 & .028 & .025 & .042 & .043 & .037 & .026 & .043 & .61 & 0.049--0.175 \\
Hormone-receptor+ carcinoma cell & .025 & .026 & .018 & .029 & .015 & .029 & .064 & .046 & .008 & .016 & .004 & .026 & .064 & .61 & 0.208--0.366 \\
Pluripotency-like cancer stem cell & .013 & .005 & .015 & .018 & .009 & .002 & .009 & .025 & .056 & .016 & .052 & .020 & .056 & .72 & 0.226--0.445 \\
Epithelial cancer stem cell & .016 & .013 & .012 & .024 & .014 & .014 & .019 & .018 & .049 & .018 & .015 & .019 & .049 & .29 & 0.531--0.581 \\
SOX9+ CD44+ stem-like cancer cell & .003 & .016 & .004 & .007 & .003 & .006 & .005 & .033 & .024 & .031 & .020 & .014 & .033 & .67 & 0.040--0.222 \\
Neural stem-like cancer cell & .007 & .011 & .011 & .028 & .007 & .019 & .003 & .008 & .017 & .014 & .024 & .014 & .028 & .71 & 0.067--0.269 \\
\midrule
\multicolumn{16}{@{}l}{\textit{Other} (12)} \\
Schwann cell & .035 & .112 & .033 & .057 & .035 & .055 & .035 & .021 & .021 & .062 & .045 & .046 & .112 & .51 & 0.189--0.388 \\
Pancreatic islet endocrine cell & .016 & .015 & .028 & .116 & .023 & .046 & .042 & .124 & .040 & .028 & .022 & .045 & .124 & .84 & 0.063--0.695 \\
Capillary pericyte & .031 & .012 & .029 & .033 & .025 & .011 & .029 & .080 & .061 & .030 & .065 & .037 & .080 & .56 & 0.125--0.451 \\
Smooth muscle cell & .020 & .013 & .017 & .021 & .022 & .017 & .033 & .043 & .040 & .018 & .106 & .032 & .106 & .42 & 0.375--0.574 \\
Neurovascular unit (endothelium + pericyte) & .034 & .033 & .065 & .045 & .039 & .032 & .015 & .010 & .019 & .014 & .006 & .028 & .065 & .56 & 0.020--0.184 \\
MYH11+ perivascular cell & .015 & .023 & .013 & .009 & .022 & .020 & .032 & .019 & .030 & .033 & .086 & .028 & .086 & .62 & 0.142--0.338 \\
Basal cell & .022 & .032 & .020 & .030 & .015 & .022 & .013 & .072 & .042 & .017 & .006 & .026 & .072 & .82 & 0.079--0.456 \\
Lymphatic endothelial cell & .029 & .021 & .015 & .036 & .024 & .033 & .033 & .031 & .011 & .033 & .022 & .026 & .036 & .42 & 0.071--0.255 \\
RGS5+ mural cell & .017 & .020 & .020 & .018 & .022 & .033 & .022 & .019 & .043 & .014 & .047 & .025 & .047 & .58 & 0.375--0.523 \\
Adipocyte & .016 & .008 & .013 & .016 & .015 & .019 & .038 & .023 & .034 & .022 & .047 & .023 & .047 & .23 & 0.473--0.530 \\
FOXI1+ ionocyte & .011 & .009 & .007 & .005 & .005 & .007 & .013 & .012 & .013 & .025 & .066 & .016 & .066 & .50 & 0.129--0.232 \\
Endothelial cell & .004 & .008 & .014 & .007 & .008 & .006 & .013 & .019 & .007 & .014 & .023 & .011 & .023 & .43 & 0.216--0.261 \\
\midrule
Mean over identities & .020 & .024 & .023 & .025 & .021 & .023 & .023 & .028 & .032 & .028 & .028 & .025 & .054 & .52 & \\
\bottomrule
\end{tabular}

%% file: tables/supp_mag_identity.tex
\begin{tabular}{@{}lcccccccc@{\hspace{6pt}}ccc@{\hspace{6pt}}c@{}}
\toprule
Identity & Virchow & H-Opt.-1 & UNI2 & Virchow2 & UNI & GigaPath & Phikon-v2 & CHIEF & DINOv3 & DINOv2 & CLIP & Mean diff. \\
\midrule
\multicolumn{13}{@{}l}{\textit{Lymphoid} (13)} \\
Antibody-secreting plasma cell & .45 (+.04) & .41 (+.07) & .44 (+.06) & .44 (+.02) & .47 (+.04) & .40 (+.03) & .45 (+.06) & .25 ($-$.02) & .06 ($-$.11) & .06 ($-$.10) & .02 ($-$.17) & $-$.01 \\
B cell & .44 ($-$.02) & .40 (+.02) & .38 (.00) & .46 (+.05) & .42 (+.07) & .43 (+.04) & .44 ($-$.03) & .46 ($-$.02) & .36 ($-$.01) & .39 (+.02) & .37 ($-$.01) & +.01 \\
CD138+ BCMA+ plasma cell & .44 (+.08) & .44 (+.10) & .43 (+.13) & .51 (+.17) & .41 (+.09) & .47 (+.07) & .41 (+.07) & .28 (.00) & .08 ($-$.02) & .05 ($-$.04) & .04 ($-$.12) & +.05 \\
CD56+ NK cell & .03 (+.02) & .06 (+.05) & .01 (.00) & .10 (+.05) & .07 (+.07) & .01 (+.02) & .13 (+.10) & .23 (+.15) & .17 (+.11) & .12 (+.04) & .14 (+.09) & +.07 \\
Checkpoint-high exhausted T cell & .15 (.00) & .14 (+.03) & .12 ($-$.01) & .14 (.00) & .13 (+.01) & .11 ($-$.01) & .16 (+.01) & .13 ($-$.01) & .08 (.00) & $-$.01 ($-$.08) & .04 ($-$.02) & $-$.01 \\
Cytotoxic T cell & .40 (+.06) & .42 (+.01) & .40 (+.02) & .41 (+.09) & .41 (+.07) & .43 (+.05) & .40 (+.02) & .41 ($-$.03) & .23 (+.06) & .15 ($-$.03) & .39 (+.10) & +.04 \\
Regulatory T cell & .17 (+.01) & .15 ($-$.04) & .14 ($-$.03) & .16 ($-$.02) & .11 ($-$.08) & .13 ($-$.04) & .10 ($-$.08) & .10 ($-$.07) & .07 ($-$.03) & .08 ($-$.02) & .06 ($-$.07) & $-$.04 \\
T cell & .61 ($-$.03) & .59 (+.09) & .54 (+.01) & .64 (+.03) & .57 (+.09) & .61 (+.08) & .58 ($-$.06) & .65 ($-$.02) & .54 ($-$.08) & .53 (+.03) & .52 (+.01) & +.02 \\
T follicular helper cell & .41 ($-$.05) & .37 ($-$.04) & .27 ($-$.09) & .34 ($-$.06) & .19 ($-$.09) & .20 ($-$.20) & .18 ($-$.06) & .17 ($-$.11) & .09 ($-$.07) & .11 ($-$.05) & .14 (.00) & $-$.08 \\
T helper17 cell & .23 ($-$.03) & .24 ($-$.04) & .22 ($-$.07) & .22 ($-$.07) & .21 ($-$.08) & .22 ($-$.08) & .20 ($-$.10) & .18 ($-$.06) & .11 ($-$.05) & .08 ($-$.08) & .11 ($-$.13) & $-$.07 \\
TOX-driven exhausted T cell & .13 ($-$.03) & .09 ($-$.04) & .08 ($-$.08) & .09 ($-$.03) & .09 ($-$.04) & .06 ($-$.06) & .04 ($-$.02) & .04 ($-$.01) & $-$.01 ($-$.03) & $-$.01 ($-$.03) & .00 ($-$.02) & $-$.04 \\
Tissue-resident memory T cell & .04 ($-$.04) & .03 ($-$.06) & .04 ($-$.05) & .03 ($-$.07) & .03 ($-$.06) & .03 ($-$.06) & .04 ($-$.05) & .04 ($-$.02) & .03 ($-$.04) & .03 ($-$.04) & .04 ($-$.03) & $-$.05 \\
Tumour-reactive exhausted T cell & .10 ($-$.04) & .06 ($-$.08) & .07 ($-$.04) & .05 ($-$.09) & .03 ($-$.06) & .05 ($-$.08) & .07 ($-$.04) & .07 ($-$.03) & .00 ($-$.02) & .00 ($-$.02) & .00 ($-$.03) & $-$.05 \\
\midrule
\multicolumn{13}{@{}l}{\textit{Myeloid} (12)} \\
CD16+ monocyte & .14 ($-$.05) & .13 ($-$.04) & .14 ($-$.05) & .17 ($-$.02) & .07 ($-$.11) & .10 ($-$.07) & .12 ($-$.06) & .12 ($-$.09) & .02 ($-$.05) & .02 ($-$.04) & .01 ($-$.07) & $-$.06 \\
Inflammatory IL-1B macrophage & .22 ($-$.02) & .21 ($-$.03) & .21 ($-$.03) & .25 ($-$.02) & .17 ($-$.04) & .22 ($-$.04) & .24 ($-$.02) & .20 (.00) & .12 ($-$.07) & .07 ($-$.07) & .03 ($-$.15) & $-$.04 \\
Interferon-activated macrophage & .02 ($-$.04) & .01 ($-$.09) & .04 ($-$.06) & .02 ($-$.06) & $-$.01 ($-$.01) & $-$.01 ($-$.04) & .01 ($-$.02) & .02 ($-$.05) & .00 (.00) & $-$.01 ($-$.01) & .00 ($-$.01) & $-$.03 \\
Lipid-associated macrophage & .32 ($-$.05) & .34 ($-$.02) & .33 (+.02) & .35 (+.01) & .29 ($-$.03) & .29 ($-$.05) & .29 ($-$.02) & .22 ($-$.02) & .21 (+.04) & .21 (.00) & .11 ($-$.06) & $-$.02 \\
M2-like macrophage & .07 ($-$.13) & .11 ($-$.13) & .11 ($-$.16) & .10 ($-$.13) & .07 ($-$.17) & .08 ($-$.17) & .06 ($-$.18) & .06 ($-$.16) & .03 ($-$.15) & .01 ($-$.14) & .03 ($-$.13) & $-$.15 \\
Mast cell & .02 (.00) & .00 ($-$.02) & .00 ($-$.02) & .01 ($-$.01) & .00 ($-$.01) & .00 ($-$.02) & .02 ($-$.02) & .01 ($-$.03) & .00 ($-$.02) & .00 ($-$.03) & .01 ($-$.02) & $-$.02 \\
Mature neutrophil & .11 (+.07) & .12 (+.07) & .11 (+.07) & .14 (+.09) & .08 (+.04) & .10 (+.05) & .13 (+.08) & .08 (+.04) & .01 ($-$.01) & .01 ($-$.02) & .01 ($-$.02) & +.04 \\
Monocyte-derived macrophage & .48 (.00) & .45 ($-$.05) & .49 (.00) & .51 (+.06) & .46 ($-$.01) & .44 ($-$.05) & .47 (.00) & .31 ($-$.04) & .26 ($-$.03) & .24 (+.02) & .15 ($-$.07) & $-$.02 \\
Perivascular C1Q+ macrophage & .08 ($-$.18) & .10 ($-$.15) & .10 ($-$.18) & .13 ($-$.15) & .12 ($-$.16) & .11 ($-$.16) & .06 ($-$.17) & .09 ($-$.15) & .03 ($-$.15) & .03 ($-$.14) & .03 ($-$.15) & $-$.16 \\
S100A8/A9+ inflammatory neutrophil & .26 ($-$.11) & .31 ($-$.08) & .26 ($-$.12) & .26 ($-$.12) & .22 ($-$.14) & .23 ($-$.16) & .23 ($-$.12) & .24 ($-$.08) & .16 ($-$.10) & .15 ($-$.08) & .11 ($-$.09) & $-$.11 \\
Tissue-resident macrophage & .22 (.00) & .22 ($-$.02) & .32 (+.06) & .33 (+.07) & .14 ($-$.03) & .08 ($-$.14) & .00 ($-$.01) & .06 (+.05) & .00 ($-$.01) & $-$.01 ($-$.01) & .00 ($-$.01) & .00 \\
iNOS+ classically activated macrophage & .00 ($-$.01) & .00 (.00) & $-$.01 ($-$.02) & $-$.01 ($-$.02) & $-$.01 (.00) & $-$.01 (.00) & $-$.01 ($-$.01) & .01 (.00) & .00 ($-$.01) & .00 (.00) & .00 ($-$.01) & $-$.01 \\
\midrule
\multicolumn{13}{@{}l}{\textit{Dendritic cells} (6)} \\
Langerhans cell & .23 ($-$.08) & .24 ($-$.10) & .25 ($-$.04) & .28 (+.01) & .23 ($-$.05) & .12 ($-$.09) & .20 ($-$.05) & .13 ($-$.09) & .13 ($-$.10) & .11 ($-$.09) & .12 ($-$.11) & $-$.07 \\
Mature regulatory DC (mregDC) & .17 ($-$.06) & .19 ($-$.08) & .18 ($-$.08) & .12 ($-$.13) & .10 ($-$.15) & .07 ($-$.15) & .08 ($-$.07) & .05 ($-$.09) & .02 ($-$.10) & .03 ($-$.08) & .05 ($-$.08) & $-$.10 \\
cDC1 & .00 ($-$.01) & .00 ($-$.01) & $-$.01 ($-$.01) & .01 ($-$.01) & .00 ($-$.01) & $-$.01 (.00) & .00 (.00) & .01 (.00) & .00 ($-$.02) & .00 ($-$.01) & .00 ($-$.01) & $-$.01 \\
cDC2 & .03 (.00) & .02 ($-$.01) & .03 (.00) & .03 (.00) & .01 (.00) & .02 (.00) & .02 (.00) & .02 (.00) & .00 ($-$.02) & $-$.01 ($-$.01) & .00 ($-$.01) & $-$.01 \\
pDC, interferon-producing & .00 ($-$.04) & .02 ($-$.20) & $-$.01 ($-$.10) & .00 ($-$.05) & .00 ($-$.04) & $-$.01 ($-$.07) & .00 ($-$.03) & .01 ($-$.04) & .00 ($-$.03) & .00 ($-$.03) & .00 ($-$.04) & $-$.06 \\
pDC, lineage program & .36 ($-$.07) & .35 ($-$.11) & .34 ($-$.08) & .35 ($-$.13) & .29 ($-$.10) & .31 ($-$.13) & .31 ($-$.07) & .30 ($-$.07) & .27 ($-$.07) & .26 ($-$.12) & .28 ($-$.08) & $-$.09 \\
\midrule
\multicolumn{13}{@{}l}{\textit{Cancer} (9)} \\
CD44+ KIT+ cancer stem cell & .01 ($-$.09) & .02 ($-$.15) & .04 ($-$.13) & .03 ($-$.15) & .03 ($-$.12) & .01 ($-$.15) & .02 ($-$.12) & .02 ($-$.09) & .00 ($-$.05) & .00 ($-$.05) & .01 ($-$.04) & $-$.10 \\
Epithelial cancer stem cell & .52 ($-$.06) & .52 ($-$.04) & .48 ($-$.06) & .54 ($-$.03) & .59 (+.06) & .53 ($-$.02) & .51 ($-$.03) & .53 (.00) & .45 ($-$.09) & .28 ($-$.29) & .18 ($-$.39) & $-$.09 \\
Hormone-receptor+ carcinoma cell & .26 ($-$.08) & .27 ($-$.01) & .29 ($-$.04) & .27 ($-$.03) & .29 (+.03) & .30 (.00) & .27 ($-$.09) & .27 ($-$.02) & .23 (+.01) & .22 (.00) & .28 (+.07) & $-$.02 \\
Neural stem-like cancer cell & .12 ($-$.13) & .13 ($-$.13) & .14 ($-$.12) & .13 ($-$.14) & .13 ($-$.13) & .14 ($-$.11) & .12 ($-$.12) & .05 ($-$.16) & .03 ($-$.04) & .03 ($-$.09) & .01 ($-$.06) & $-$.11 \\
Neural-crest-like tumour cell & .51 ($-$.07) & .55 ($-$.03) & .51 ($-$.04) & .53 ($-$.03) & .47 ($-$.12) & .51 ($-$.04) & .51 ($-$.04) & .45 ($-$.13) & .07 ($-$.20) & .21 ($-$.13) & .15 ($-$.35) & $-$.11 \\
Ovarian carcinoma cell & .71 (+.08) & .67 (+.02) & .63 (+.03) & .70 ($-$.11) & .67 (+.07) & .64 (+.06) & .62 (+.09) & .67 (+.12) & .66 (+.11) & .63 ($-$.06) & .57 (+.01) & +.04 \\
Pluripotency-like cancer stem cell & .35 ($-$.09) & .39 ($-$.04) & .37 ($-$.05) & .38 ($-$.06) & .34 ($-$.08) & .36 ($-$.05) & .37 ($-$.06) & .29 ($-$.11) & .19 ($-$.08) & .19 ($-$.12) & .09 ($-$.13) & $-$.08 \\
Proliferating glial-like tumour cell & .54 ($-$.17) & .61 ($-$.01) & .61 ($-$.03) & .53 ($-$.11) & .57 ($-$.09) & .55 ($-$.07) & .56 ($-$.04) & .43 ($-$.18) & .55 ($-$.11) & .45 ($-$.15) & .38 ($-$.01) & $-$.09 \\
SOX9+ CD44+ stem-like cancer cell & .13 ($-$.08) & .12 ($-$.10) & .13 ($-$.08) & .13 ($-$.09) & .12 ($-$.10) & .12 ($-$.09) & .10 ($-$.11) & .03 ($-$.14) & .01 ($-$.08) & .00 ($-$.07) & .01 ($-$.03) & $-$.09 \\
\midrule
\multicolumn{13}{@{}l}{\textit{Other} (12)} \\
Adipocyte & .30 ($-$.23) & .30 ($-$.19) & .31 ($-$.21) & .33 ($-$.17) & .32 ($-$.20) & .31 ($-$.20) & .46 ($-$.02) & .35 ($-$.18) & .28 ($-$.24) & .23 ($-$.24) & .30 ($-$.22) & $-$.19 \\
Basal cell & .39 ($-$.07) & .46 (+.08) & .45 (.00) & .31 ($-$.07) & .28 ($-$.17) & .22 ($-$.20) & .28 ($-$.06) & .14 ($-$.05) & .09 (.00) & .09 (.00) & .07 ($-$.01) & $-$.05 \\
Capillary pericyte & .28 ($-$.11) & .32 ($-$.12) & .32 ($-$.12) & .32 ($-$.10) & .23 ($-$.17) & .31 ($-$.14) & .26 ($-$.18) & .21 ($-$.13) & .02 ($-$.12) & .04 ($-$.09) & .02 ($-$.11) & $-$.13 \\
Endothelial cell & .14 ($-$.11) & .13 ($-$.12) & .14 ($-$.11) & .14 ($-$.12) & .10 ($-$.14) & .10 ($-$.14) & .11 ($-$.14) & .10 ($-$.13) & .06 ($-$.17) & .04 ($-$.18) & .03 ($-$.19) & $-$.14 \\
FOXI1+ ionocyte & .38 (+.16) & .36 (+.14) & .34 (+.11) & .36 (+.13) & .33 (+.10) & .31 (+.09) & .34 (+.12) & .24 (+.02) & .06 ($-$.14) & .07 ($-$.06) & .07 ($-$.08) & +.05 \\
Lymphatic endothelial cell & .09 ($-$.03) & .14 ($-$.01) & .16 (+.07) & .08 ($-$.01) & .04 ($-$.03) & .02 ($-$.23) & .04 ($-$.04) & .04 ($-$.07) & .05 ($-$.07) & .05 ($-$.06) & .07 ($-$.03) & $-$.05 \\
MYH11+ perivascular cell & .31 ($-$.01) & .32 (+.01) & .30 ($-$.01) & .28 ($-$.02) & .33 (.00) & .28 (.00) & .28 (+.01) & .16 ($-$.12) & .08 ($-$.07) & .06 ($-$.08) & .07 ($-$.07) & $-$.03 \\
Neurovascular unit (endothelium + pericyte) & .13 ($-$.03) & .13 ($-$.06) & .14 ($-$.02) & .17 (+.02) & .06 ($-$.01) & .08 ($-$.01) & .07 (+.02) & .01 ($-$.03) & .01 ($-$.04) & .01 ($-$.01) & .01 ($-$.02) & $-$.02 \\
Pancreatic islet endocrine cell & .42 ($-$.27) & .38 ($-$.28) & .60 ($-$.03) & .46 ($-$.08) & .22 ($-$.44) & .40 ($-$.12) & .36 ($-$.20) & .16 ($-$.15) & .03 ($-$.08) & .01 ($-$.07) & .02 ($-$.04) & $-$.16 \\
RGS5+ mural cell & .25 ($-$.19) & .34 ($-$.12) & .37 ($-$.14) & .33 ($-$.15) & .34 ($-$.18) & .33 ($-$.15) & .33 ($-$.12) & .32 ($-$.16) & .20 ($-$.17) & .18 ($-$.25) & .24 ($-$.18) & $-$.16 \\
Schwann cell & .32 ($-$.03) & .24 ($-$.15) & .27 ($-$.04) & .25 ($-$.05) & .23 ($-$.08) & .20 ($-$.16) & .26 ($-$.02) & .18 ($-$.06) & .17 ($-$.07) & .18 ($-$.01) & .13 ($-$.08) & $-$.07 \\
Smooth muscle cell & .46 ($-$.10) & .48 ($-$.08) & .47 ($-$.06) & .51 ($-$.07) & .46 ($-$.09) & .43 ($-$.11) & .40 ($-$.14) & .36 ($-$.15) & .25 ($-$.24) & .18 ($-$.33) & .08 ($-$.29) & $-$.15 \\
\midrule
Mean & .25 ($-$.05) & .25 ($-$.05) & .25 ($-$.04) & .26 ($-$.04) & .22 ($-$.06) & .22 ($-$.06) & .22 ($-$.04) & .19 ($-$.06) & .13 ($-$.06) & .11 ($-$.07) & .11 ($-$.07) & $-$.05 \\
\bottomrule
\end{tabular}

%% file: tables/supp_identity_q.tex
\begin{tabular}{@{}lcccccccc@{\hspace{8pt}}ccc@{}}
\toprule
Identity & Virchow & H-Opt.-1 & UNI2 & Virchow2 & UNI & GigaPath & Phikon-v2 & CHIEF & DINOv3 & DINOv2 & CLIP \\
\midrule
\multicolumn{12}{@{}l}{\textit{Lymphoid} (13)} \\
Antibody-secreting plasma cell & .41 & .34 & .38 & .42 & \textbf{.43} & .36 & .39 & .28 & .17 & .16 & .19 \\
B cell & .46 & .38 & .39 & .41 & .36 & .39 & .47 & \textbf{.48} & .37 & .38 & .37 \\
CD138+ BCMA+ plasma cell & .36 & .34 & .29 & .34 & .31 & \textbf{.39} & .34 & .28 & .10 & .09 & .16 \\
CD56+ NK cell & .00 & .01 & .01 & .05 & $-$.01 & $-$.02 & .03 & \textbf{.08} & .06 & .08 & .04 \\
Checkpoint-high exhausted T cell & .15 & .11 & .14 & .14 & .12 & .12 & \textbf{.15} & .14 & .09 & .07 & .07 \\
Cytotoxic T cell & .34 & .41 & .38 & .32 & .35 & .38 & .38 & \textbf{.45} & .17 & .19 & .30 \\
Regulatory T cell & .16 & .18 & .18 & .18 & \textbf{.19} & .17 & .17 & .17 & .09 & .10 & .13 \\
T cell & .63 & .50 & .53 & .61 & .48 & .53 & .64 & \textbf{.67} & .62 & .50 & .51 \\
T follicular helper cell & \textbf{.46} & .41 & .36 & .40 & .28 & .39 & .24 & .29 & .17 & .16 & .14 \\
T helper17 cell & .26 & .28 & .29 & .29 & .29 & \textbf{.30} & .30 & .23 & .16 & .16 & .24 \\
TOX-driven exhausted T cell & \textbf{.16} & .14 & .16 & .12 & .13 & .12 & .05 & .05 & .03 & .02 & .02 \\
Tissue-resident memory T cell & .08 & .09 & .09 & \textbf{.10} & .09 & .09 & .09 & .06 & .07 & .06 & .07 \\
Tumour-reactive exhausted T cell & .14 & .14 & .11 & \textbf{.14} & .09 & .13 & .11 & .11 & .02 & .01 & .03 \\
\midrule
\multicolumn{12}{@{}l}{\textit{Myeloid} (12)} \\
CD16+ monocyte & .19 & .17 & .19 & .18 & .18 & .17 & .18 & \textbf{.21} & .07 & .05 & .08 \\
Inflammatory IL-1B macrophage & .24 & .25 & .24 & \textbf{.27} & .21 & .25 & .26 & .20 & .19 & .14 & .18 \\
Interferon-activated macrophage & .06 & \textbf{.10} & .09 & .07 & .00 & .04 & .02 & .06 & .00 & .00 & .01 \\
Lipid-associated macrophage & \textbf{.36} & .36 & .32 & .34 & .32 & .34 & .31 & .24 & .17 & .22 & .18 \\
M2-like macrophage & .20 & .24 & \textbf{.27} & .23 & .25 & .25 & .24 & .22 & .19 & .15 & .16 \\
Mast cell & .03 & .02 & .02 & .02 & .01 & .01 & .04 & \textbf{.04} & .02 & .03 & .03 \\
Mature neutrophil & .04 & .05 & .04 & \textbf{.06} & .04 & .04 & .04 & .04 & .02 & .03 & .02 \\
Monocyte-derived macrophage & .48 & \textbf{.50} & .48 & .46 & .47 & .49 & .47 & .35 & .29 & .22 & .22 \\
Perivascular C1Q+ macrophage & .26 & .25 & .28 & .28 & \textbf{.29} & .27 & .23 & .24 & .17 & .17 & .19 \\
S100A8/A9+ inflammatory neutrophil & .37 & .39 & .38 & .38 & .36 & \textbf{.39} & .35 & .32 & .26 & .24 & .20 \\
Tissue-resident macrophage & .22 & .24 & \textbf{.26} & .26 & .17 & .22 & .00 & .02 & .01 & .00 & .01 \\
iNOS+ classically activated macrophage & .01 & $-$.01 & .01 & \textbf{.01} & $-$.01 & $-$.02 & .00 & .01 & .01 & .00 & .01 \\
\midrule
\multicolumn{12}{@{}l}{\textit{Dendritic cells} (6)} \\
Langerhans cell & .30 & \textbf{.34} & .28 & .27 & .29 & .21 & .25 & .21 & .23 & .20 & .22 \\
Mature regulatory DC (mregDC) & .23 & \textbf{.27} & .26 & .25 & .25 & .22 & .15 & .15 & .12 & .11 & .13 \\
cDC1 & .01 & .01 & .00 & .02 & .01 & $-$.01 & .00 & .02 & \textbf{.02} & .01 & .02 \\
cDC2 & .03 & .03 & .03 & \textbf{.03} & .02 & .02 & .02 & .02 & .02 & .01 & .01 \\
pDC, interferon-producing & .04 & \textbf{.23} & .09 & .05 & .04 & .06 & .03 & .05 & .03 & .02 & .04 \\
pDC, lineage program & .44 & .46 & .43 & \textbf{.48} & .38 & .44 & .38 & .37 & .34 & .38 & .35 \\
\midrule
\multicolumn{12}{@{}l}{\textit{Cancer} (9)} \\
CD44+ KIT+ cancer stem cell & .10 & .17 & .16 & \textbf{.17} & .15 & .16 & .14 & .11 & .05 & .05 & .05 \\
Epithelial cancer stem cell & \textbf{.58} & .57 & .54 & .56 & .53 & .55 & .54 & .53 & .54 & .57 & .57 \\
Hormone-receptor+ carcinoma cell & .34 & .28 & .32 & .30 & .27 & .31 & \textbf{.37} & .28 & .22 & .21 & .21 \\
Neural stem-like cancer cell & .25 & \textbf{.27} & .26 & .27 & .26 & .26 & .24 & .21 & .07 & .11 & .07 \\
Neural-crest-like tumour cell & .58 & .58 & .54 & .55 & \textbf{.59} & .55 & .55 & .59 & .28 & .34 & .50 \\
Ovarian carcinoma cell & .63 & .65 & .60 & \textbf{.80} & .60 & .58 & .53 & .54 & .55 & .69 & .55 \\
Pluripotency-like cancer stem cell & \textbf{.44} & .43 & .42 & .44 & .42 & .41 & .43 & .40 & .27 & .30 & .23 \\
Proliferating glial-like tumour cell & \textbf{.70} & .62 & .64 & .64 & .66 & .63 & .59 & .60 & .66 & .59 & .39 \\
SOX9+ CD44+ stem-like cancer cell & .21 & .22 & .21 & \textbf{.22} & .22 & .21 & .21 & .17 & .09 & .08 & .04 \\
\midrule
\multicolumn{12}{@{}l}{\textit{Other} (12)} \\
Adipocyte & \textbf{.53} & .49 & .52 & .49 & .52 & .51 & .49 & .52 & .52 & .47 & .52 \\
Basal cell & .45 & .37 & \textbf{.46} & .38 & .45 & .42 & .34 & .19 & .10 & .09 & .08 \\
Capillary pericyte & .39 & .44 & .43 & .42 & .40 & \textbf{.45} & .44 & .35 & .14 & .13 & .13 \\
Endothelial cell & .25 & .25 & .25 & \textbf{.26} & .24 & .24 & .24 & .23 & .23 & .22 & .22 \\
FOXI1+ ionocyte & .22 & .22 & .23 & \textbf{.23} & .23 & .22 & .21 & .22 & .20 & .13 & .15 \\
Lymphatic endothelial cell & .12 & .16 & .09 & .09 & .07 & \textbf{.25} & .08 & .10 & .12 & .10 & .09 \\
MYH11+ perivascular cell & .32 & .30 & .31 & .30 & \textbf{.34} & .28 & .27 & .27 & .15 & .14 & .14 \\
Neurovascular unit (endothelium + pericyte) & .17 & \textbf{.18} & .16 & .14 & .08 & .09 & .04 & .04 & .04 & .02 & .03 \\
Pancreatic islet endocrine cell & \textbf{.69} & .66 & .62 & .54 & .66 & .51 & .56 & .30 & .11 & .08 & .06 \\
RGS5+ mural cell & .44 & .46 & .50 & .47 & \textbf{.52} & .49 & .45 & .48 & .38 & .43 & .42 \\
Schwann cell & .35 & \textbf{.39} & .32 & .30 & .32 & .36 & .27 & .25 & .25 & .19 & .21 \\
Smooth muscle cell & .56 & .55 & .53 & \textbf{.57} & .55 & .54 & .54 & .51 & .49 & .51 & .37 \\
\bottomrule
\end{tabular}

%% file: tables/supp_raw_identity.tex
\begin{tabular}{@{}lccccccccccc@{}}
\toprule
Cell identity & Virchow & H-Opt.-1 & UNI2 & Virchow2 & UNI & GigaPath & Phikon-v2 & CHIEF & DINOv3 & DINOv2 & CLIP \\
\midrule
\multicolumn{12}{@{}l}{\textit{Lymphoid} (13)} \\
Antibody-secreting plasma cell & .331 & .320 & .264 & .361 & .216 & .214 & .311 & .252 & .109 & .111 & .152 \\
B cell & .447 & .367 & .364 & .365 & .240 & .342 & .386 & .347 & .162 & .245 & .258 \\
CD138+ BCMA+ plasma cell & .255 & .178 & .198 & .216 & .176 & .240 & .138 & .149 & .074 & .014 & .141 \\
CD56+ NK cell & .082 & .117 & .169 & .146 & .122 & .111 & .082 & .063 & .031 & .070 & .150 \\
Checkpoint-high exhausted T cell & $-$.014 & .047 & $-$.011 & .070 & .004 & $-$.012 & .078 & .008 & $-$.016 & .010 & $-$.002 \\
Cytotoxic T cell & $-$.034 & $-$.003 & .011 & .010 & .019 & $-$.018 & .105 & $-$.036 & .086 & .009 & $-$.042 \\
Regulatory T cell & .142 & .106 & .069 & .166 & .031 & .084 & .248 & .051 & .064 & .063 & .029 \\
T cell & .554 & .311 & .640 & .355 & .323 & .323 & .412 & .537 & .164 & .228 & .238 \\
T follicular helper cell & .116 & .076 & .080 & .174 & .087 & .113 & .124 & .084 & .012 & .039 & .012 \\
T helper17 cell & .162 & .198 & .089 & .079 & .046 & .077 & .057 & .073 & .007 & .032 & .033 \\
TOX-driven exhausted T cell & .012 & $-$.003 & $-$.003 & .022 & $-$.004 & $-$.022 & $-$.019 & $-$.046 & .006 & $-$.016 & $-$.034 \\
Tissue-resident memory T cell & .033 & .043 & .062 & .061 & .083 & .094 & .066 & .010 & .007 & $-$.027 & .021 \\
Tumour-reactive exhausted T cell & .103 & .020 & .053 & $-$.017 & $-$.019 & .009 & $-$.030 & $-$.055 & $-$.045 & $-$.034 & $-$.076 \\
\midrule
\multicolumn{12}{@{}l}{\textit{Myeloid} (12)} \\
CD16+ monocyte & $-$.044 & .032 & $-$.018 & .059 & .021 & .163 & .016 & .022 & $-$.054 & $-$.023 & $-$.038 \\
Inflammatory IL-1B macrophage & .004 & $-$.011 & .012 & $-$.011 & $-$.025 & .004 & .020 & $-$.037 & $-$.071 & $-$.057 & $-$.039 \\
Interferon-activated macrophage & $-$.063 & $-$.005 & $-$.035 & $-$.001 & .003 & $-$.009 & .038 & $-$.056 & $-$.069 & $-$.069 & $-$.042 \\
Lipid-associated macrophage & .020 & .139 & .153 & .139 & .105 & .154 & .171 & .105 & .089 & .104 & .032 \\
M2-like macrophage & .018 & .021 & $-$.002 & $-$.006 & .036 & $-$.017 & .053 & .021 & .042 & .007 & .018 \\
Mast cell & .047 & .057 & .022 & .065 & .102 & .142 & .074 & .040 & .042 & .003 & .022 \\
Mature neutrophil & $-$.046 & $-$.002 & $-$.008 & .028 & .006 & .013 & .008 & $-$.053 & $-$.027 & $-$.042 & $-$.043 \\
Monocyte-derived macrophage & .187 & .310 & .322 & .275 & .253 & .273 & .272 & .095 & .197 & .127 & .088 \\
Perivascular C1Q+ macrophage & .134 & .153 & .108 & .142 & .122 & .148 & .150 & .178 & .056 & .033 & .042 \\
S100A8/A9+ inflammatory neutrophil & .123 & .245 & $-$.019 & .235 & .087 & .118 & .193 & .235 & $-$.030 & .041 & $-$.023 \\
Tissue-resident macrophage & $-$.003 & $-$.013 & $-$.019 & .009 & .028 & $-$.021 & $-$.064 & $-$.024 & $-$.035 & $-$.050 & $-$.005 \\
iNOS+ classically activated macrophage & $-$.068 & $-$.022 & $-$.065 & $-$.030 & $-$.032 & $-$.026 & $-$.035 & .021 & $-$.078 & $-$.042 & $-$.074 \\
\midrule
\multicolumn{12}{@{}l}{\textit{Dendritic cells} (6)} \\
Langerhans cell & .186 & .298 & .236 & .200 & .140 & .094 & .166 & .135 & .091 & .121 & .109 \\
Mature regulatory DC (mregDC) & .210 & .143 & .203 & .136 & .155 & .187 & .185 & .114 & .044 & .067 & .093 \\
cDC1 & $-$.035 & $-$.019 & $-$.041 & $-$.051 & $-$.024 & $-$.022 & $-$.059 & $-$.044 & $-$.015 & $-$.061 & $-$.053 \\
cDC2 & $-$.035 & $-$.018 & $-$.055 & $-$.022 & $-$.025 & $-$.021 & $-$.063 & $-$.059 & $-$.046 & $-$.061 & $-$.042 \\
pDC, interferon-producing & .008 & .044 & .003 & .036 & .023 & $-$.000 & .007 & .016 & .017 & $-$.022 & $-$.011 \\
pDC, lineage program & .367 & .392 & .326 & .318 & .314 & .321 & .388 & .350 & .185 & .164 & .198 \\
\midrule
\multicolumn{12}{@{}l}{\textit{Cancer} (9)} \\
CD44+ KIT+ cancer stem cell & .028 & .082 & .026 & .076 & .038 & .026 & .034 & .075 & $-$.037 & $-$.045 & $-$.022 \\
Epithelial cancer stem cell & .138 & .347 & .302 & .246 & .267 & .240 & .235 & .213 & .089 & .087 & .084 \\
Hormone-receptor+ carcinoma cell & .254 & .253 & .246 & .146 & .193 & .204 & .310 & .172 & .024 & .071 & .021 \\
Neural stem-like cancer cell & .081 & .084 & $-$.008 & .086 & .015 & .001 & $-$.051 & $-$.055 & $-$.030 & $-$.053 & $-$.037 \\
Neural-crest-like tumour cell & $-$.027 & .323 & .256 & .083 & .108 & .143 & .116 & .127 & $-$.003 & $-$.066 & .034 \\
Ovarian carcinoma cell & .487 & .481 & .411 & .642 & .582 & .541 & .400 & .360 & .338 & .110 & .213 \\
Pluripotency-like cancer stem cell & .153 & .163 & .087 & .114 & .040 & .087 & .045 & .011 & .011 & .007 & $-$.028 \\
Proliferating glial-like tumour cell & .648 & .521 & .648 & .393 & .809 & .669 & .453 & .533 & .318 & .104 & .134 \\
SOX9+ CD44+ stem-like cancer cell & .107 & .060 & .055 & .124 & .021 & .063 & .040 & $-$.020 & $-$.017 & $-$.022 & $-$.004 \\
\midrule
\multicolumn{12}{@{}l}{\textit{Other} (12)} \\
Adipocyte & .281 & .464 & .429 & .298 & .291 & .318 & .385 & .416 & .530 & .413 & .267 \\
Basal cell & .293 & .246 & .309 & .224 & .219 & .081 & .115 & .029 & .008 & $-$.013 & .010 \\
Capillary pericyte & .077 & .118 & .039 & .155 & .025 & .040 & .096 & .054 & .001 & .030 & $-$.010 \\
Endothelial cell & .070 & .174 & .084 & .100 & .201 & .059 & .069 & .095 & .108 & $-$.011 & .015 \\
FOXI1+ ionocyte & .093 & .081 & .077 & .114 & .095 & .085 & .077 & .051 & .029 & $-$.001 & $-$.001 \\
Lymphatic endothelial cell & .085 & .162 & .076 & .126 & .086 & .086 & .103 & .066 & .081 & .064 & .067 \\
MYH11+ perivascular cell & .346 & .417 & .411 & .434 & .258 & .260 & .294 & .206 & .213 & .056 & .111 \\
Neurovascular unit (endothelium + pericyte) & .101 & .043 & .101 & .091 & .028 & .101 & .064 & .042 & $-$.005 & $-$.010 & .003 \\
Pancreatic islet endocrine cell & .132 & .206 & .195 & .160 & .159 & .140 & .237 & .117 & .002 & .055 & .065 \\
RGS5+ mural cell & .148 & .074 & .242 & .103 & .072 & .075 & .138 & .036 & .033 & .001 & .057 \\
Schwann cell & .252 & .245 & .322 & .297 & .281 & .285 & .304 & .173 & .136 & .083 & .122 \\
Smooth muscle cell & .082 & .236 & .160 & .162 & .064 & .135 & .131 & .051 & .098 & .017 & .116 \\
\midrule
BioScore & .165 & .206 & .197 & .184 & .158 & .155 & .169 & .123 & .092 & .054 & .064 \\
\bottomrule
\end{tabular}